\documentclass[10pt,journal,compsoc]{IEEEtran}
\ifCLASSOPTIONcompsoc
  \usepackage[nocompress]{cite}
\else
  \usepackage{cite}
\fi

\ifCLASSINFOpdf
\else
\fi
\usepackage{amsmath,mathrsfs,amssymb}
\usepackage{hyperref}
\usepackage[table]{xcolor}

\usepackage{booktabs}       % professional-quality tables
\usepackage{longtable}
\usepackage{threeparttable}
\usepackage{multirow}
\usepackage{multicol}
\usepackage{graphicx}
\usepackage{bm}

\usepackage{amsthm}
\usepackage{dsfont}
\usepackage{lipsum} 

\usepackage{makecell,rotating,multirow,diagbox}
\usepackage{bm}
\usepackage{makecell}
\usepackage{cuted}

\usepackage{amsmath, amssymb}
\usepackage{algorithm}
\usepackage{algpseudocode}
\usepackage{soul}

\usepackage{adjustbox}

\usepackage{caption}
\usepackage{subfig}

\usepackage{amssymb}% http://ctan.org/pkg/amssymb
\usepackage{pifont}% http://ctan.org/pkg/pifont

\newtheorem{theorem}{Theorem}

\def\a{\mathbf{a}}

\def\e{\mathbf{e}}

\def\y{\mathbf{y}}
\def\W{\mathbf{W}}
\def\X{\mathbf{X}}

\def\P{\mathbf{P}}
\def\Q{\mathbf{Q}}

\def\x{\bm{x}}

\def\y{\mathbf{y}}
\def\W{\mathbf{W}}

\def\w{\textbf{w}}

\def\x{\boldsymbol{x}}

\def\a{\mathbf{a}}

\def\e{\mathbf{e}}

\def\y{\mathbf{y}}
\def\W{\mathbf{W}}
\def\X{\mathbf{X}}

\def\w{\textbf{w}}

\def\x{\bm{x}}

\def\y{\mathbf{y}}
\def\W{\mathbf{W}}

\def\b{\mathbf{b}}
\def\Y{\mathbf{Y}}
\def\w{\textbf{w}}

\begin{document}
\title{No One-Size-Fits-All Neurons: Task-based Neurons for Artificial Neural Networks}

\author{Feng-Lei Fan$^\dag$, Meng Wang$^\dag$, Hang-Cheng Dong, Jianwei Ma$^*$, Tieyong Zeng$^*$

% <-this % stops a space
\IEEEcompsocitemizethanks{
% \IEEEcompsocthanksitem *Tieyong Zeng, Yiming Cui, and Jing-Xiao Liao are co-corresponding authors.
% % note need leading \protect in front of \\ to get a newline within \thanks as
% % \\ is fragile and will error, could use \hfil\break instead.
% E-mail: see http://www.michaelshell.org/contact.html

\IEEEcompsocthanksitem Feng-Lei Fan and Meng Wang are co-first authors. Jianwei Ma and Tieyong Zeng are co-corresponding authors. The research is supported by the National Natural Science Foundation of China (nos. 42230806and U23B6010), the China National Petroleum Corporation-Peking University Strategic Cooperation Project of Fundamental Research, the Lagrange Mathematics and Computing Research Center, Huawei Technologies France, and the Startup Fund of the City University of Hong Kong.

\IEEEcompsocthanksitem Feng-Lei Fan is with Department of Data Science, City University of Hong Kong, Kowloon, Hong Kong. 

\IEEEcompsocthanksitem 
Meng Wang is with School of Mathematics, Harbin Institute of Technology, Harbin, Heilongjiang Province 150001, China.

\IEEEcompsocthanksitem 
Hang-Cheng Dong is with School of Instrumentation, Harbin Institute of Technology, Harbin, Heilongjiang Province 150001, China.

\IEEEcompsocthanksitem 
Jianwei Ma is with School of Mathematics, Harbin Institute of Technology, Harbin, China, and School of Earth and Space Sciences, Peking University, Beijing, China.

\IEEEcompsocthanksitem Tieyong Zeng is with Institute for Advanced Study, Beijing Normal-Hong Kong Baptist University, Zhuhai, Guangdong, China. 

}
% <-this % stops an unwanted space
}

\markboth{Journal of \LaTeX\ Class Files,~Vol.~14, No.~8, August~2023}%
{Shell \MakeLowercase{\textit{et al.}}: Bare Demo of IEEEtran.cls for Computer Society Journals}



\IEEEtitleabstractindextext{%
\begin{abstract}

In the past decade, many successful networks are on novel architectures, which almost exclusively use the same type of neurons. 
Recently, more and more deep learning studies have been inspired by the idea of NeuroAI and the neuronal diversity observed in human brains, leading to the proposal of novel artificial neuron designs. Designing well-performing neurons represents a new dimension relative to designing well-performing neural architectures. Biologically, the brain does not rely on a single type of neuron that universally functions in all aspects. Instead, in our brain, neurons are often task-based. In this study, we address the following question: since the human brain is a task-based neuron user, can the artificial network design go from the task-based architecture design to the task-based neuron design? Since methodologically there are no one-size-fits-all neurons, given the same structure, task-based neurons can enhance the feature representation ability relative to the existing universal neurons due to the intrinsic inductive bias for the task. Specifically, we propose a two-step framework for prototyping task-based neurons.
First, symbolic regression is used to identify optimal formulas that fit input data by utilizing base functions such as polynomials. We introduce VSR that stacks all variables in a vector and regularizes each input variable to perform the same computation, which can increase the regression speed, facilitate efficacy in high dimensions, and enable parallel computation. Second, we parameterize the acquired elementary formula to make parameters learnable, which serves as the aggregation function of the neuron. The activation functions such as ReLU and the sigmoidal functions remain the same because they have proven to be good. As the initial step, we evaluate the proposed framework using polynomials as base functions. Empirically, systematic experimental results on synthetic data, classic benchmarks, and real-world applications show that the proposed task-based neuron design is not only feasible but also delivers competitive performance over other state-of-the-art models.  
We have shared our code in \url{https://github.com/NewT123-WM/Task_based_neurons}.

\end{abstract}

}

\maketitle

\IEEEdisplaynontitleabstractindextext

% \fi
\IEEEpeerreviewmaketitle

\IEEEraisesectionheading{\section{Introduction}\label{sec:introduction}}
\vspace{-0.2cm}


\IEEEPARstart{I}{n} the past decade, a majority of deep learning research is on designing outstanding architectures, such as shortcuts \cite{he2016deep} and neural architecture search (NAS, \cite{hassantabar2022curious}). Almost exclusively, these works employ neurons of the same type that use an inner product and a nonlinear activation. We refer to such a neuron as a linear neuron, and a network made of these neurons as a linear network (LN) hereafter. Recently, the field “NeuroAI" emerged \cite{zador2022toward} to advocate that a large amount of neuroscience knowledge can help catalyze the next generation of AI. This idea is well-motivated, as the brain remains the most intelligent system to date, and an artificial network can be regarded as a miniature of the brain. Following the advocacy of “NeuroAI", it is noted that our brain is made up of many functionally and morphologically different neurons, while the existing mainstream artificial networks are homogeneous at the neuronal level. Thus, why not introduce neuronal diversity and examine associated merits?

Our overarching opinion is that the neuron type and architecture are two complementary dimensions of an artificial network. Designing well-performing neurons represents a new dimension relative to designing well-performing architectures. Therefore, the neuronal type should be given full attention to harness the full potential of connectionism. In recent years, a plethora of studies have introduced new neurons into deep learning 
% \cite{chrysos2021deep, jiang2020nonlinear, mantini2021cqnn, goyal2020improved, liao2022attention} 
such as polynomial neurons \cite{chrysos2021deep} and quadratic neurons \cite{jiang2020nonlinear, mantini2021cqnn, goyal2020improved, liao2022attention}. Despite focusing only on a specific type of neuron, this thread of studies reasonably verifies the feasibility and potential of developing deep learning with new neurons. However, the performance of these neurons is not universally satisfactory, and the improvement is minor on some tasks. Biological neuronal diversity, both in terms of morphology and functionality, arises from the brain's needs to perform complex tasks \cite{peng2021morphological}. The brain does not rely on a single type of neuron to universally function in all aspects. Instead, neurons are often task-based in our brain. Hence, in the realm of deep learning, we think that promoting neuronal diversity should also not be limited to specific neuron types like linear or quadratic neurons. Instead, it should take into account the specific task at hand.

\textit{Can we design different neurons for different tasks (task-based neurons)?} Computationally, the philosophy of task-based architectures and task-based neurons is quite distinct. 
The former is “one-for-all”, aligning with the scaling law \cite{bahri2024explaining}, which implicitly assumes that stacking a universal and basic type of neurons into different structures
can solve a wide class of complicated nonlinear problems. This philosophy is
well underpinned by the universal approximation theorem \cite{hornik1990universal}. The latter is “one-for-one", which assumes that there are no one-size-fits-all neuron types, and it is better to solve a specific problem by prototyping customized neurons. Because task-based neurons are imparted with the implicit bias for the task, the network of task-based neurons can integrate the task-driven forces of all these neurons, which given the same structure should exhibit stronger performance than the network of generic neurons. The key difference between task-based and generic neurons is that the mathematical expression in the task-based neurons is adaptive according to the preference of the task, while in the generic neurons, the mathematical expression is preset.

% More favorably, task-based neurons and task-based architectures are complementary, and their synergy can lead to the creation of even more powerful models compared to their individual usage. 

% \begin{figure}[htb] 
% \vspace{-0.3cm}
% \centering
% \includegraphics[width=0.85\linewidth]{Figure/Figure_task_based_neurons.pdf}
% \vspace{-0.3cm}
% \caption{Two steps to establish a task-based neuron. a) The symbolic regression constructs an elementary neuronal model, which stacks all variables in a vector and regularizes each input variable to perform the same computation. b) The acquired elementary formula is parameterized to be learnable, serving as the aggregation function of a neuron. } 
% \label{fig:task_based_neuron_design}
% \end{figure}
% \vspace{-0.2cm}

\begin{figure*}[h!]
	\centering
	\includegraphics[width=0.9\textwidth]{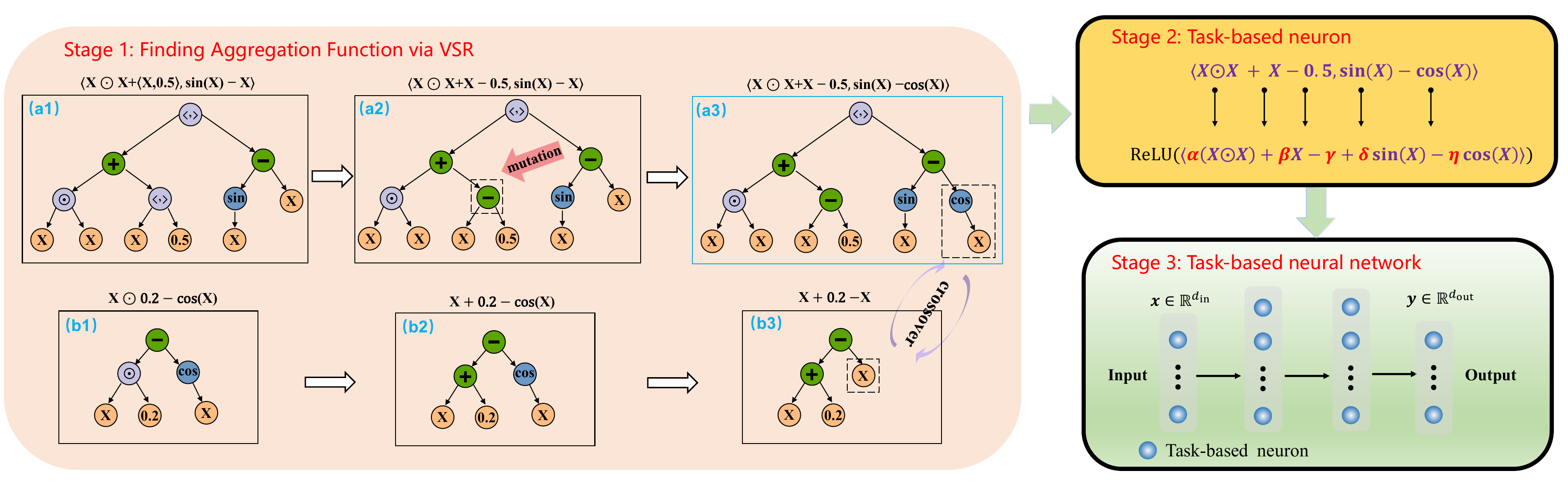}
	\caption{A schematic diagram of constructing a task‑based network. First, VSR combines all input variables into a vector and defines vector‑computational operators. Second, after VSR fits a formula, the coefficients are parameterized and an activation function is attached to obtain a neuron, which is task-oriented. Third, we construct a network with task-based neurons, which we refer to as a task-based network. }
\label{fig:sr_para}
\vspace{-0.5cm}
\end{figure*}

Along this direction, three main challenges face us in prototyping task-based neurons: 1) How to efficiently design task-based neurons? 2) How to make the resultant neurons transferable to a network? 3) How to transfer the superiority at the neuronal level to the network level? Here, we propose a two-step framework to address these challenges: First, we introduce the vectorized symbolic regression (VSR) to construct an elementary neuronal model, as depicted in Figure \ref{fig:sr_para}. Symbolic regression (SR) draws inspiration from scientific discoveries in physics \cite{schmidt2009distilling}, aiming to identify optimal formulas that fit input data by utilizing base functions such as logarithmic, trigonometric, and exponential functions. VSR stacks all variables in a vector and regularizes each input variable to perform the same computation. Given the complexity and unclear nonlinearity of the tasks, formulas learned from VSR can capture the underlying patterns in the data, and these patterns are different in different contexts. Thus, fixed formulas used in pre-designed neurons are disadvantageous. Second, we parameterize the acquired elementary formula to make parameters learnable, which serves as the aggregation function of the neuron. The role of VSR is to identify the basic patterns behind data, the parameterization allows the task-based neurons to adapt and interact with each other within a network. The activation functions such as the sigmoidal function remain the same when connected to a network, as these activation functions are widely tested as well-performed. 

VSR in our framework greatly expedites the search process by avoiding learning highly complex and disordered formulas, particularly for high-dimensional inputs (challenge 1). It also facilitates parallel computing and ensures the feasibility of building a deep network with the designed neurons. Moreover, according to the type of data, the learned formulas will be converted into a neuron in multi-layer perceptrons or a convolution kernel in convoluitional networks (challenge 2). 
The formulas searched by VSR capture basic patterns in the data and are not sufficiently complex to solve highly intricate problems due to the ablated search space. By connecting task-based neurons into a lightweight network, we leverage the power of connectionism, enabling further amplification of the advantages without concerns about overfitting (challenge 3). We refer to a network made of task-based neurons as a task-based network (TN) hereafter. 

As the initial step, we evaluate the feasibility and superiority of the proposed framework on tabular data and images. Motivated by the success of quadratic and polynomial neurons, we mainly use polynomials as the base functions for symbolic regression. System experiments show that task-based neurons and associated networks can outperform networks of preset neurons and other state-of-the-art models. To summarize, our contributions are threefold:

\begin{itemize}
    \item Towards NeuroAI, we propose a framework to design task-based neurons, which is a new dimension compared to task-based architectures and can greatly expand the armory of deep learning models.

    \item We propose the VSR to solve the computational challenges in prototyping new neurons. Methodologically, our work is the first to introduce symbolic regression into the design of neurons in deep learning.

    % \item Methodologically, we merge two paradigms by synergizing task-based neurons and task-based architectures to build a novel neural network model.

    \item With systematic experiments over synthetic data, public data, and real-world applications, we confirm the effectiveness of the task-based neurons.
    
\end{itemize}

\vspace{-0.6cm}

\section{Related Work}

\textbf{Neuronal diversity}.
There has been a growing interest in recent years to prototype new neurons and introduce neuronal diversity into artificial networks \cite{fan2023towards}. It is important to clarify that modifying the activation function should not be considered as creating new neurons. This is because the decision boundary of a neuron is solely determined by the aggregation function, as long as the activation function is monotonic. Currently, excluding spiking neurons \cite{ostojic2014two} featuring the spatiotemporal processing ability, exploring neuronal diversity primarily revolves around polynomial or quadratic neurons \cite{fan2023towards}, which replace the inner product with a polynomial or a quadratic function, expanding the range of computations by individual neurons. 

\begin{table}[htbp]
\centering
\caption{A summary of the recently-proposed quadratic neurons. $\sigma(\cdot)$ is the nonlinear activation function. $\odot$ denotes Hadamard product. $\W \in \mathbb{R}^{n\times n}$, $\w_i \in \mathbb{R}^{n\times 1}$, and the bias terms in these neurons are omitted for simplicity.}
\vspace{-0.3cm}
\scalebox{0.95}{
\begin{tabular}{l|l}
\hline
Authors           & Formulations          \\ \hline
Zoumpourlis \textit{et al.}(2017) \cite{zoumpourlis2017non} &  $\y=\sigma(\x^{\top}\W\x+\w^\top\x)$              \\ \hline
Jiang \textit{et al.}(2019) \cite{jiang2020nonlinear}  & 
\multirow{2}{*}{$\y=\sigma(\x^{\top}\W\x$)} 
\\ \cline{1-1}
Mantini \& Shah(2021) \cite{mantini2021cqnn}     &                        \\ \hline
Goyal \textit{et al.}(2020) \cite{goyal2020improved}    & $\y=\sigma(\w^\top(\x\odot\x))$               \\ \hline
Bu \& Karpatne(2021) \cite{bu2021quadratic}       & $\y=\sigma((\w_1^\top\x)(\w_2^\top\x))$             \\ \hline
Xu \textit{et al.}(2022) \cite{xu2022quadralib} & $\y=\sigma((\w_1^\top\x) (\w_2^\top\x)+\w_3^\top\x)$ \\
\hline
Fan \textit{et al.}(2018)   \cite{fan2018new}      & $\y=\sigma((\w_1^\top\x)(\w_2^\top\x)+\w_3^\top(\x\odot\x))$        \\ \hline
\end{tabular}}
\label{tab:neurons}
\vspace{-0.1cm}
\end{table}

In recent years, polynomial neurons were revisited in the hope of enhancing the expressive ability of a single neuron. The key issue in designing polynomial neurons is to reduce the complexity such that they can be deployed into a deep network. \cite{liu2021dendrite, chrysos2021deep} decreased the complexity of polynomial neurons via tensor decomposition and factor sharing, while a majority of studies directly used quadratic neurons to save parameters to express high-order terms. Table \ref{tab:neurons} summarizes the recently-proposed quadratic neurons. As seen, given an $n$-dimensional input, the complexity of neurons in \cite{zoumpourlis2017non, jiang2020nonlinear,mantini2021cqnn} is of $\mathcal{O}(n^2)$, which is still not bearable for deep networks, while neurons from \cite{goyal2020improved,bu2021quadratic,xu2022quadralib,fan2018new} enjoy the linear parametric complexity. Notably, neurons in \cite{goyal2020improved,bu2021quadratic,xu2022quadralib} are the special cases of \cite{fan2018new}. Polynomial and quadratic neurons have demonstrated competitive performance in many tasks \cite{liao2022attention}.

\noindent
\textbf{Neural Architecture Search (NAS).}
Neural architecture search (NAS, \cite{hassantabar2022curious}) has attracted substantial attention due to its success in automatically discovering novel and efficient network architectures. NAS typically operates at the architecture level, searching for optimal configurations of layer types, depths, widths, and connectivity patterns. These techniques often rely on weight-sharing supernets or performance predictors to reduce search cost.

Both our work and NAS involve the evolutionary method because the objective being optimized does not have gradients. However, the existing NAS approaches mainly focus on designing macro-structures, while largely overlooking neuronal types. Our proposed methodology for task-based neurons shift the search granularity from network-level architecture to neuron-level expression design. While universal approximation theorems suggest that linear networks can approximate these functions given sufficiently many neurons, this is primarily a theoretical guarantee. In practice, representing such functions using basic NAS-style operations remains challenging. Neuronal diversity is orthogonal to traditional NAS and offers a new perspective for AutoML. As such, task-based neurons can be integrated into various architectures discovered by NAS methods. The synergy between two lines of research suggests stronger potential, which will be our future research.

\noindent
\textbf{Symbolic Regression in Deep Learning}
Recent advances in symbolic regression include learning underlying PDEs from data \cite{KIYANI2023116258}, differentiable symbolic regression \cite{zeng2023differentiable}, and so on. Some works use neural networks to improve symbolic regression. For example, \cite{kim2020integration} used a neural network architecture to span the hypothesis space of symbolic regression such that the formulas can be learned in an end-to-end manner. \cite{zhang2023deep} extended symbolic regression to solve parametric systems whose coefficients may vary but the intrinsic structure of the underlying equation keeps intact. \cite{liu2023snr} proposed a Symbolic Network-based Rectifiable Learning Framework (SNR) that can correct errors generated in the learning-with-experience model. 

However, the role of symbolic regression in deep learning is much less explored.  Note that the recent Kolmogorov-Arnold Network (KAN)  \cite{liu2024kan} fits the coefficients of basis functions to learn the activation function. Though it is close to the symbolic learning, it remains in the paradigm of coefficient fitting. To the best of our knowledge, our work is the first time that symbolic regression has been used for task-based neuronal designs.

% \subsection{Tabular Data} 

% Recent advances in tabular data analysis have focused on improving the performance and efficiency of modeling techniques, especially in machine learning. Some notable advances include
% gradient boosting algorithms, such as XGBoost \cite{chen2016xgboost}, LightGBM \cite{ke2017lightgbm}, and CatBoost \cite{hancock2020catboost}, network-based method like DANet \cite{chen2022danets}, transformer-based models like TabTransformer \cite{huang2020tabtransformer} and TabBERT \cite{yin2020tabert}, and NAS-based method like TabNAS \cite{yang2022tabnas}. In this work, we show that the proposed task-based neuron design can empower a network to establish new state-of-the-art results over tabular data.
\vspace{-0.4cm}

\section{Method}

In this section, we first provide a detailed description of how to create task-based neurons using the proposed VSR. Then, we will explain how the proposed framework addresses the aforementioned three challenges.

\vspace{-0.3cm}
\subsection{VSR }

\textbf{1. Overall Framework}. Unlike traditional regression algorithms that fit numerical coefficients, symbolic regression automatically discovers analytical mathematical expressions from data. To construct task‑based neurons, we propose VSR, which regularizes every variable to learn the same formula, allowing us to organize all variables into a vector. The formulas are then based on vector computation, as illustrated in Figure~\ref{fig:sr_para}. Unlike traditional symbolic regression, which tends to identify heterogeneous formulas, VSR enforces homogeneity for the scalability of neurons. Given a dataset $ D = \left\{ (\X_i, y_i) \right\}_{i=1}^N $, where $ \X_i \in \mathbb{R}^d $ represents the $i$-th input sample, and \( y_i \) represents the corresponding observed output, the goal of symbolic regression is to search for a function \( f: \mathbb{R}^d \to \mathbb{R} \) in a predefined space of mathematical expressions that best fits the data:
\begin{equation}
    \min_{f \in \mathcal{F}} \sum_{i=1}^N L(f(\X_i), y_i),
    \label{loss}
\end{equation}
where \( L(\cdot, \cdot) \) is the error function that is also called the fitness function in genetic programming \cite{kinnear1994advances}, e.g., mean squared error, and \( \mathcal{F} \) represents the space of functions composed of basic operators. There are three kinds of operator sets: i) $\Phi_1 = \{\sin, \cos, \log, \exp, \dots \}$, which is a collection of the basic functions. The input argument is a vector. For example, $\sin(\X)=[\sin x_1, \sin x_2, \dots, \sin x_d]$; ii) $\Phi_2 = \{\odot, \langle,\rangle\}$ that denote two basic operations between two vectors: $\P\odot\Q=[p_1q_1,p_2q_2,\dots,p_dq_d]$ is the Hadamard prodoct, while $\langle \P,\Q\rangle=p_1q_1+p_2q_2+\dots+p_dq_d$ is the inner product; iii) $\Phi_3 = \{+,-,\times,\div\}$ that operate on two scalars, two vectors, or one scalar and one vector. For $+,-,\times,\div$ between one scalar and cone vector, the scalar conducts the same computation for each element of the vector. For example, $a-\X=[a-x_1, a-x_2, \cdots, a-x_d]$. The core of symbolic regression is to search for the optimal analytical formula in $\mathcal{F}$ composed of operators from $\Phi_1, \Phi_2, \Phi_3$. 

Note that in VSR, both the scalar $a$ and constant vector $\a=[a,a,\cdots,a]$ are needed. The former computes the Hadamard product with a vector, while the latter computes the inner product with a vector.

\textbf{2. Expression Representation}. The mathematical expressions need to be encoded appropriately. Like traditional SR, a mathematical expression in VSR is also represented as a tree. An expression tree is a hierarchical structure: (1) Internal nodes: representing operators from $\Phi_1, \Phi_2, \Phi_3$. (2) Leaf nodes: representing the input vector $\X$, constants, or constant vectors. Such a tree structure allows the mathematical expression to be expressed through a series of simple computational steps, facilitating the subsequent search algorithms to provide a unified expression format. The formula is obtained by reading the tree from the bottom to the up. For instance, the tree (b1) in Figure \ref{fig:sr_para} denotes the formula $\X \odot \bm{0.2}-\cos(\X)$. In addition, it should be noted that VSR does not lose interactions among variables---it captures them through combining two inner products. For example, $\langle \W_1, \x \rangle\cdot \langle\W_2,\x \rangle = (w_{11}x_1 + w_{12}x_2)(w_{21}x_1 + w_{22}x_2)$ enables interaction terms like $x_1 \cdot x_2$.

\textbf{3. Search Mechanism of VSR}. Due to the complexity of the mathematical expression, which is highly non-linear and difficult to directly optimize, symbolic regression typically employs genetic algorithms to explore the space. The most common method is Genetic Programming. In the framework, each expression tree is treated as an individual, and evolution proceeds through the following steps:

\textit{i) Initialization}: Randomly generate a population of expression trees as the initial population.

\textit{ii) Mutation}: An operation that randomly selects a position in an individual and generates a new individual through single‑point mutation. Due to its randomness, mutation can reintroduce functions and variables that were previously eliminated, potentially leading to the discovery of novel and effective expressions. By injecting variability into the population, mutation plays a crucial role in exploring the solution space and preventing premature convergence to suboptimal solutions (see Figure~\ref{fig:sr_para}, where the tree (a2) is obtained from the tree (a1) by mutating a node). 

Specially, during evolution, the constant mutation is performed: with probability $p$, a constant leaf is perturbed by adding Gaussian noise or replaced by a new random constant. No gradient‑based tuning occurs.

\textit{iiii) Crossover}: An operation that generates new individuals by the subtree crossover among selected individuals, thereby exploring the symbolic expression space. The specific method randomly selects subtrees of the winner candidates and exchanges them (see Figure~\ref{fig:sr_para}, where tree (a3) exchanges the branch with the tree (b3)). This operation promotes diversity in the population and can lead to the discovery of effective mathematical formulas.

\textit{iv) Sanity Check and Fitness Evaluation}:  The loss function requires a scalar prediction per sample. To guarantee this property, if the output is a vector, we automatically reduce it to a scalar by summing over all coordinates. This reduction is applied deterministically and does not discard any individual. The resulting scalar is then used to compute the fitness (e.g., MSE). Then, we compare the fitness of each expression tree based on Eq. \eqref{loss}. This applies Darwin's principle of natural selection (survival of the fittest) to determine the winning individual according to the fitness function.

\textbf{4. Prototyping Neurons and Convolutions}. We expect that VSR can identify hidden patterns behind data collected from different tasks. Leveraging these patterns to prototype new neurons would be useful. These patterns are basic and not necessarily specific functions. For instance, we refer to a cell as circular that is characterized by an elliptical equation, but we don't need to specify the radius of the circle. To take advantage of these patterns, we reparameterize the learned formula by making the fixed constants trainable to better fit the task at hand. Such neurons should perform better than generic neurons. Now, let us prototype task-driven neurons for fully-connected networks and task-driven convolutions for convolutional networks. For simplicity, we call both task-based neurons. 

% \begin{figure*}[!htbp]
% 	\centering	
    
% 	% \begin{adjustbox}{width=1.1\textwidth,center}		
% 	% 	\subfloat[\label{fig:fcn_case}]{\includegraphics[width=0.5\textwidth]{Figure/fcn_case.pdf}}
% 	% 	\subfloat[\label{fig:conv_case}]{\includegraphics[width=0.2\textwidth]{Figure/conv_case.pdf}}
% 	% \end{adjustbox}		

%     \begin{adjustbox}{width=1\textwidth,center}
%         \begin{tabular}{cc}
%             \subfloat[\label{fig:fcn_case}]{\includegraphics[width=0.5\textwidth]{Figure/fcn_case.pdf}} &
%             \subfloat[\label{fig:conv_case}]{\includegraphics[width=0.5\textwidth]{Figure/conv_case.pdf}} 
%         \end{tabular}
%     \end{adjustbox}
    
% 	\caption{	
%  Illustration of task-driven neurons in different architectures. (a) Fully connected case: the input vector $\x \in \mathbb{R}^{d_{\text{in}}}$ is expanded into polynomial terms (e.g., $\x$, $\x\odot\x$), each multiplied by a learnable weight matrix $\W_k$, and the results are summed to form the output $\y \in \mathbb{R}^{d_{\text{out}}}$. (b) Convolutional case: the input feature map $\x \in \mathbb{R}^{C_{\text{in}}\times H \times D}$ is expanded into higher-order terms (e.g., $\x$, $\x\odot^3\x$), each convolved with a learnable kernel $\W_k$, and the results are summated to produce the output $\Y \in \mathbb{R}^{C_{\text{out}}\times H' \times W'}$.
% 	}
% 	\label{fig:diff_arc}
% \end{figure*}

\underline{(a) Fully-Connected Networks.}
Let $\underbrace{\x\odot\x\odot\cdots\odot\x}_{k}=\x\odot^k\x$ denote element-wise exponentiation to the power of $k$. In a fully connected layer, the input $\x \in \mathbb{R}^{d_{\text{in}}}$ is expanded into a polynomial according to the symbolic expression. Suppose that we have the following learned formula
\[
f(\x) = \bm{0.068}(\x\odot^2\x)^\top+\bm{0.15}\x^\top+0.76.
\]
Then, we parameterize the above formula to get a neuron. In the matrix form, for each monomial $\x\odot^k\x$, we associate a learnable weight matrix $\W_k \in \mathbb{R}^{d_{\text{out}} \times d_{\text{in}}}$. 
The forward computation is 
\[
\y = \W_2 (\x\odot^2\x)^\top + \W_1 \x^\top + \b, \quad \y \in \mathbb{R}^{d_{\text{out}}},
\] 
where $\b \in \mathbb{R}^{d_{\text{out}}}$ is the bias vector. The illustration is shown in Figure \ref{fig:sr_para}.

\underline{(b) Convolutional Case.} In convolutional layers, the same practice is applied to feature maps. 
Given a symbolic expression such as
\[
f(\x) = \bm{1.5}(\x\odot^3\x)^\top - \bm{1.5}\x + 0.5,
\]
where $\x \in \mathbb{R}^{C_{\text{in}} \times H \times D}$ is the input feature map, the module expands $\x$ into a polynomial ($\x$ and $\x\odot^3\x$). 
For each monomial $\x\odot^k\x$, we assign an independent, learnable convolution kernel $\W_k \in \mathbb{R}^{C_{\text{out}} \times C_{\text{in}} \times h \times d}$.
The forward computation becomes
\[
\Y = \texttt{Conv}(\W_3, \x\odot^3\x) + \texttt{Conv}(\W_1, \x), \quad \Y \in \mathbb{R}^{C_{\text{out}} \times H' \times D'},
\]
where $\texttt{Conv}$ denotes a standard convolution operation. 

Moreover, in both fully-connected and convolutional networks, we also use the activation function like ReLU. Thus, taking $\mathtt{ReLU}$ as an example, we have  
\[
\mathtt{ReLU}(\y) = \mathtt{ReLU}(\W_2 (\x\odot^2\x)^\top + \W_1 \x^\top + \b),
\]
and 
\[
\mathtt{ReLU}(\Y) = \mathtt{ReLU}(\texttt{Conv}(\W_3, \x\odot^3\x) + \texttt{Conv}(\W_1, \x)).
\]

\vspace{-0.2cm}

In prototyping task-based neurons, we consider three important questions: 1) How to efficiently design task-based neurons? 2) How to make the resultant neurons transferable to a network? 3) How to transfer the superiority at the neuronal level to the network level? Now, we explain how the above proposed method addresses them one by one:

\textbf{Question 1}: ``How to efficiently design task-based neurons?'' — The traditional SR is not efficient, particularly for high-dimensional inputs: First, the regression process of traditional symbolic regression becomes slow and computationally expensive for high-dimensional inputs. The search space becomes vast for high-dimensional inputs, as it requires checking an arbitrary form of interactions among two or more input variables. Recently, it was proved \cite{virgolin2022symbolic} that SR is NP-hard, which means that no known solution for SR problems can be determined in polynomial time  across all instances. Recent state-of-the-art work only scaled SR for problems of around 50 dimensions \cite{ruan2025discovering}. Second, the formulas learned by the traditional symbolic regression are heterogeneous, which do not support parallel computing and GPU acceleration. Thus, such formulas cannot serve as the aggregation function of a neuron because the resultant neuron cannot be easily integrated into deep networks.

\underline{Low Complexity and Parallel Computing}: Due to the homogeneity, the proposed VSR leads to mathematical formulas with much fewer parameters. Given $d$-dimensional inputs, the number of parameters is $\mathcal{O}(d)$, which is at the same level as the linear neuron. Moreover, because each variable conducts the same operation, formulas obtained from the proposed VSR can be organized into the vector or matrix computation, which can facilitate parallel computation aided by GPUs to support more efficient training of deep networks composed of task-based neurons.

\underline{Regression Performance}: Moreover, VSR is not merely a parallelizable approximation of SR; it can, in fact, surpass SR in fitting high-dimensional data. VSR enforces a homogeneous formula structure across input variables, thereby enabling a more effective and scalable exploration of the hypothesis space. Experiments in Supplementary Materials ( Section 1.2 in SMs) confirm the superiority of VSR over SR in high dimensions.

\textbf{Question 2}: ``How to make the resultant neurons transferable to a network?'' — Our framework ensures seamless network integration through three complementary mechanisms: i) \textit{Computational Homogeneity}: If a task-based neuron uses a heterogeneous formula, it is almost impossible to train a large-scale network based on them. VSR-derived formulas exhibit structural uniformity across input dimensions, enabling efficient vectorized implementation. This property permits GPU-accelerated matrix operations essential for neural networks. ii) \textit{Differentiable Parameterization}: We don't stick to the formula learned by VSR. We think the pattern indicated by the formula is instrumental. By replacing symbolic coefficients with learnable parameters, we ensure the end-to-end differentiability and a direct integration into a network. iii) \textit{Domain-Specific Adaptation}: Based on the type of data, the context for constructing the formula is different. For tabular data, we adopt the standard VSR, and the resultant formula will be prototyped into a neuron in MLPs. While for images, we do VSR after using an autoencoder to do dimension reduction for an image. Then, the learned formula serves as convolution kernels to ensure transferability to CNNs (see Section 3 of SMs).

\textbf{Question 3}: ``How to transfer the superiority at the neuronal level to the network level?'' -- Let $\e_1^*$ and $\e_l^*$ be the optimal errors of fully-connected networks with one hidden layer $\mathcal{N}_1$ and $l$ hidden layers $\mathcal{N}_l$. We think that neuronal type and architecture are two orthogonal contributors to a network's power. It is seen in Theorem \ref{superiority} that $\|\e_l^*\|\leq \|\e_1^*\|$ for $l\geq 2$, when $\mathcal{N}_l$ consists of $\mathcal{N}_1$. Thus, there are two ways of improving a network's power: reducing $\|\e_1^*\|$ with neuron design and reducing $\|\e_l\|$ with architecture design. Empirically, their efficacy in reducing the error and which one is dominant are up to different problems. Since $\|\e_1^*\|$ is the upper bound of $\|\e_l^*\|$, the superiority at the neuronal level (a lower $\|\e_1^*\|$) can lead to the superiority at the network level (a lower $\|\e_l^*\|$). This is approximation power analysis, and in practice, we avoid the overfitting by intentionally designing task-based networks that are shallower and have fewer parameters. Thus, the task-specific inductive bias embedded in individual neurons can effectively enhance network-level performance. We now informally state Theorem \ref{superiority}. 

\begin{theorem}[Informal]
For any $l>1$, we have
$\left\|\mathbf{e}_{l}^{*}\right\| \leq\left\|\mathbf{e}_{1}^{*}\right\|$ when $\mathcal{N}_l$ consists of $\mathcal{N}_1$ structurally.
\label{superiority}
\end{theorem}

Please refer to Section 5 of SMs for detailed proof. 

\noindent \textbf{Remark 1}. This theorem establishes that the optimal approximation error $\left\|\mathbf{e}_{l}^{*}\right\|$ is no greater than $\left\|\mathbf{e}_{1}^{*}\right\|$. It means that composing task-based neurons further reduces the initial error $\left\|\mathbf{e}_{1}^{*}\right\|$, while the task-based neuron by VSR provides a lower $\left\|\mathbf{e}_{1}^{*}\right\|$ than the conventional neuron. This process is analogous to the progressive refinement of a Taylor series by adding more and more higher-order terms. This theorem reflects the fact that deeper networks have at least as much representational capacity as shallower ones. Depth is still essential. However, to make a model achieve a similar approximation ability, designing more powerful neurons (like our task‑based neurons) is likely to save the depth—a pattern we indeed observe empirically.

% Furthermore, since task-based neurons and task-based architectures address the two most important and complementary facets of a network, it is natural to synergize task-based neurons with task-based architectures discovered through Neural Architecture Search (NAS, \cite{hassantabar2022curious, yang2022tabnas}). The methodology of combining task-based neurons with task-based architectures should outperform approaches that involve task-based neurons with a vanilla architecture or linear neurons with a task-based architecture. By combining task-based neurons with task-based architectures, we can further enhance the network's performance by leveraging the benefits of both components.

\vspace{-0.3cm}

\section{Analysis Experiments}
In this section, we present a series of experiments designed to analyze the feasibility, necessity, and superiority of the proposed task-based neurons. For all experiments, we prescribe that the function space of symbolic regression is polynomial, and symbolic discovery is performed strictly on training data to prevent any potential data leakage.
We first validate via the synthetic data the feasibility of the proposed framework that VSR can capture correct hidden formulas from heavily-noised data (Section 1.1 in SMs). Then, we find that VSR is inferior to SR in fitting low-dimensional data, while VSR much outperforms SR in high-dimensional cases. But for low-dimensional data, the resultant task-based networks can outperform the symbolic regression (Section 1.2 in SMs). Furthermore, we compare task-based neurons with neurons using random polynomials to confirm that the polynomials learned from the symbolic regression are reasonable (Section 1.3 of SMs). Next, we compare task-based neurons with conventional neurons, under different architectures (Section 1.4 in SMs) and four different activation functions (Section 1.5 in SMs), and different quadratic neurons (Section 1.6 in SMs) to show the superiority of task-based neurons.  Also, we extend the search space from polynomial bases to trigonometric functions (Section 1.7 in SMs). In addition, we present the computational properties of VSR, such as the convergence behavior and search time (Section 1.8 in SMs) as well as hyper-parameter sensitivity of VSR (Section 2 in SMs).

\begin{table}[t!]
\caption{Comparison of different tabular models under minimal-capacity (“single-layer”) configurations.}
\centering
\scalebox{0.8}{
\begin{tabular}{l|c|c|c}
\toprule
\textbf{Model} & \textbf{Particle Collision} & \textbf{Asteroid Prediction} & \textbf{\# Parameters}  \\
\midrule
TabNet & $0.0055 \pm 0.0006 $  & $\mathbf{0.0854 \pm 0.1221}$ & $\approx$ 2K  \\  
TabTransformer & $0.0278 \pm 0.0187$ & $0.3863 \pm 0.2851$ & $\approx$ 14K  \\  
FT-Transformer & $0.0064 \pm 0.0059$ & $ 0.2136 \pm 0.2033$ & $\approx$ 3K  \\ 
DANets & $0.0080 \pm 0.0014 $ & $0.1412 \pm 0.1464$ & $\approx$ 19K  \\  
\cellcolor{green!15}Task-based Network & \cellcolor{green!15}$\mathbf{0.0029 \pm 0.0006}$ & \cellcolor{green!15}$0.1275 \pm 0.1745 $ & \cellcolor{green!15}$\approx$2K \\  
\bottomrule
\end{tabular}}
\label{tab:single_layer}
\vspace{-0.3cm}
\end{table}

\subsection{Tabular Datasets}
Here, we compare the task-based networks with other state-of-the-art models over two real-world tasks: \textbf{High-energy Particle Collision Prediction}  and \textbf{Asteroid Prediction}. To highlight the superiority of the network using task-based neurons, we select advanced machine learning models for comparison, namely XGBoost\cite{chen2016xgboost}, LightGBM\cite{ke2017lightgbm}, CatBoost\cite{hancock2020catboost}, TabNet\cite{arik2021tabnet}, TabTransformer\cite{huang2020tabtransformer}, FT-Transformer\cite{gorishniy2021revisiting} and DANETs\cite{chen2022danets}. All these models are either classic models or recent models that were published in prestigious venues of machine learning. We also describe the configurations of each baseline model and the details of two datasets in Section 4 of SMs.

% \begin{table}[htb]
% \caption{The test results (MSE errors) of different models on particle collision and asteroid prediction datasets.}
% \vspace{-0.3cm}
%     \centering
%     \scalebox{1.0}{\begin{tabular}{l|c|c}
%     \toprule  

%  Method & particle collision  & asteroid prediction\\ \hline
%  XGBoost & $0.0094\pm0.0006$ & $0.0646\pm0.1031$
%  \\
%  LightGBM & $0.0056\pm0.0004$ & $0.1391\pm0.1676$ 
%  \\
%  CatBoost & $0.0028\pm0.0002$ & $0.0817\pm0.0846$
%  \\
%  TabNet & $0.0040\pm0.0006$  & $0.0627\pm0.0939$ 
%  \\ 
%  TabTransformer & $0.0038\pm0.0008$ & $0.4219\pm0.2776$ 
%  \\
%  FT-Transformer & $0.0050\pm0.0020$ & $0.2136\pm0.2189$ 
%  \\
%  DANETs & $0.0076\pm0.0009$ & $0.1709\pm0.1859$
%  \\
%  Task-based Network  & $\mathbf{0.0016\pm0.0005}$ & $\mathbf{0.0513\pm0.0551}$
%  \\

%   \bottomrule
%     \end{tabular}}
%     \label{tab:particle_collision}
%     \vspace{-0.3cm}
% \end{table}

\begin{table}[htb]
\caption{The test results (MSE errors) of different models on particle collision and asteroid prediction datasets.}
\vspace{-0.3cm}
    \centering
    \scalebox{1.0}{\begin{tabular}{l|c|c}
    \toprule  
    Method & Particle Collision  & Asteroid Prediction\\
    \midrule
    XGBoost & $0.0094\pm0.0006$ & $0.0646\pm0.1031$ \\
    LightGBM & $0.0056\pm0.0004$ & $0.1391\pm0.1676$ \\
    CatBoost & $0.0028\pm0.0002$ & $0.0817\pm0.0846$ \\    
    TabNet & $0.0040\pm0.0006$ & $0.0463 \pm 0.0523$ \\
    TabTransformer & $0.0052\pm0.0010$ & $ 0.3539 \pm 0.2568 $ \\
    FT-Transformer & $0.0052\pm0.0007$ & $0.1814 \pm 0.1765 $ \\
    DANets & $0.0064\pm0.0006$ & $ 0.1317 \pm 0.1368 $ \\
    \textbf{Task-based Network} & $\mathbf{0.0016 \pm 0.0005}$ & $\mathbf{0.0327 \pm 0.0164 }$ \\
  \bottomrule
    \end{tabular}}
    \label{tab:particle_collision}
    \vspace{-0.3cm}
\end{table}

For both datasets, the training set, validation set, and test set are randomly divided at a ratio of 8:1:1. We choose the MSE as the evaluation metric. To ensure a reliable comparison and mitigate the influence of randomness, we conduct 10 times tests for each model, and the final results are presented in the form of $\mathrm{mean} \pm (\mathrm{std})$. The detailed test results are shown in Table \ref{tab:particle_collision}.

It can be seen that while TabTransformer has good performance on the particle collision dataset but unsatisfactory performance on the asteroid dataset, CatBoost and TabNet consistently perform good on both datasets. The highlight of Table \ref{tab:particle_collision} is that the task-based network is the best performer on both datasets. It leads TabTransformer, FT-transformer, and DANETs by a large margin. 

Furthermore, regarding why simplified architectures can still achieve good results, we conduct an additional experiment in which all compared models are minimized to their “single-layer” variants to compare the intrinsic ability of their foundational blocks. As shown in Table~\ref{tab:single_layer}, the minimal task-based network has been well-performing, \textit{i.e.}, it is the best and second-best performer on the particle collision and asteroid prediction datasets, respectively. This observation suggests that task-based neurons are expressive. Therefore, the complicated structure is longer needed to seek more learning power.

\vspace{-0.4cm}
\section{Comparative Experiments}

\begin{table*}[ht!]
\centering

\caption{Performance comparison of KAN and its variants with TN. Best results are highlighted in \textbf{bold}.}
\label{tab:kan_vs_tn}

\resizebox{1.0\textwidth}{!}{
\begin{tabular}{lccccccccccccccc}
\toprule
\multirow{2}{*}{Datasets} & \multicolumn{3}{c}{KAN}  & \multicolumn{3}{c}{ChebyKAN}  & \multicolumn{3}{c}{FourierKAN}  & \multicolumn{3}{c}{FastKAN}  & \multicolumn{3}{c}{TN} \\
\cmidrule(lr){2-4} \cmidrule(lr){5-7} \cmidrule(lr){8-10} \cmidrule(lr){11-13} \cmidrule(lr){14-16}
  & Perf. & FLOPs & Param. & Perf. & FLOPs & Param. & Perf. & FLOPs & Param. & Perf. & FLOPs & Param. & Perf. & FLOPs & Param.\\
\midrule
California Housing & 0.2093 & 2891 & 621 & \underline{0.2016} & 860 & 345 & 0.2165 & 1635 & 700 & 0.2183 & 1837 & 665 & \textbf{0.0553} & 322 & 340\\
House Sales & \underline{0.1207} & 5556 & 1323 & 0.1375 & 1770 & 735 & 0.2282 & 3390 & 1480 & 0.1466 & 3696 & 1393 & \textbf{0.0139} & 1368 & 1391\\
Airfoil Self Noise & 0.1731 & 2276 & 459 & \underline{0.1461} & 650 & 255 & 1.2447 & 1230 & 520 & 0.3747 & 1408 & 497 & \textbf{0.0170} & 224 & 241\\
Diamonds & \underline{0.0181} & 3096 & 675 & 0.0198 & 930 & 375 & 0.0216 & 1770 & 760 & 0.0375 & 1980 & 721 & \textbf{0.0069} & 670 & 685\\
Abalone & \underline{0.6677} & 2891 & 621 & 0.6697 & 860 & 345 & 1.0214 & 1635 & 700 & 0.6916 & 1837 & 665 & \textbf{0.0246} & 564 & 584\\
Bike Sharing & 0.0640 & 3711 & 837 & 0.0642 & 1140 & 465 & \underline{0.0637} & 2175 & 940 & 0.0863 & 2409 & 889 & \textbf{0.0076} & 888 & 919\\
Space GA & 0.3318 & 2481 & 513 & \underline{0.3139} & 720 & 285 & 1.1100 & 1365 & 580 & 0.4465 & 1551 & 553 & \textbf{0.0055} & 256 & 267\\
Airlines Delay & \underline{0.9288} & 3096 & 675 & 0.9308 & 930 & 375 & 0.9297 & 1770 & 760 & 0.9335 & 1980 & 721 & \textbf{0.1690} & 636 & 667\\ 
\hline  \hline
Credit & \underline{0.7367} & 5405 & 1296 & 0.7350 & 1730 & 720 & 0.5700 & 3315 & 1451 & 0.7117 & 3607 & 1365 & \textbf{0.7498} & 1208 & 1248\\
Heloc & \underline{0.7230} & 5815 & 1404 & 0.7208 & 1870 & 780 & 0.6987 & 3585 & 1571 & 0.7125 & 3893 & 1477 & \textbf{0.7262} & 1308 & 1332\\
Electricity & \textbf{0.8145} & 2945 & 648 & \underline{0.8139} & 890 & 360 & 0.8092 & 1695 & 731 & 0.7979 & 1891 & 693 & 0.8119 &  608 &	636 \\
Phoneme & \underline{0.8501} & 2330 & 486 & 0.8492 & 680 & 270 & 0.6815 & 1290 & 551 & 0.8289 & 1462 & 525 & \textbf{0.8651} & 464 & 482\\
Magic Telescope & \underline{0.8734} & 3355 & 756 & \textbf{0.8753} & 1030 & 420 & 0.8573 & 1965 & 851 & 0.8582 & 2177 & 805 & 0.8531 &  388 & 400 \\
Vehicle & \underline{0.7686} & 5103 & 1242 & 0.7549 & 1650 & 690 & 0.2725 & 3165 & 1393 & 0.6294 & 3429 & 1309 & \textbf{0.8000} & 1152 & 1173\\
Orange Vs Grapefruit & 0.9555 & 2330 & 486 & \textbf{0.9835} & 680 & 270 & 0.9517 & 1290 & 551 & 0.9628 & 1462 & 525 & \underline{0.9813} &  260	& 279 \\
Eye Movements & \underline{0.5826} & 6894 & 1701 & 0.5777 & 2250 & 945 & 0.4919 & 4320 & 1902 & 0.5414 & 4662 & 1785 & \textbf{0.6080} &  1636 & 1675 \\
\bottomrule
\end{tabular}
\label{kan_comparison}
}
\vspace{-0.5cm}
\end{table*}

\subsection{Comparison with KANs} 

% \begin{table}[!htbp]
% \centering
% \footnotesize % 设置字体大小为小号
% \caption{Comparison of KAN and TN performance.}
% \label{tab:kan_vs_tn_sr}
% \begin{tabular}{lcccccc}
% \toprule
%  \multirow{2}{*}{Datasets} & \multicolumn{3}{c}{KAN} & \multicolumn{3}{c}{TN} \\
% \cmidrule(lr){2-4} \cmidrule(lr){5-7}
%  & Perf. & FLOPs & Param. & Perf. & FLOPs & Param. \\
% \midrule

% California Housing & 0.1017 & 19544 & 7056 & 0.1135 & 3104 & 1601 \\
% House Sales & 0.0073 & 10551 & 3384 & 0.0075 & 1488 & 769 \\
% Airfoil & 0.0220 & 3017 & 684 & 0.0712 & 152 & 89 \\
% Diamonds & 0.0040 & 8241 & 2520 & 0.0043 & 1104 & 577 \\
% Abalone & 0.0232 & 3740 & 900 & 0.0243 & 392 & 209 \\
% Bike Sharing & 0.0086 & 6460 & 1872 & 0.0081 & 1296 & 673 \\
% Space GA & 0.0056 & 3258 & 756 & 0.0063 & 328 & 177 \\
% Airlines Delay & 0.1563 & 11997 & 3816 & 0.1602 & 1680 & 881 \\ 
% \midrule \midrule
% Credit & 0.7431 & 3474 & 864 & 0.7441 & 1184 & 618 \\
% Heloc & 0.7027 & 7186 & 1944 & 0.7117 & 848 & 438 \\
% Electricity & 0.7962 & 2992 & 720 & 0.7932 & 1056 & 554 \\
% Phoneme & 0.8292 & 6845 & 2016 & 0.7942 & 864 & 458 \\
% MagicTelescope & 0.8479 & 8770 & 2736 & 0.8389 & 1184 & 618 \\
% Vehicle & 0.7576 & 12138 & 4032 & 0.7676 & 1728 & 892 \\
% Oranges-vs-Grape & 0.9302 & 6845 & 2016 & 0.9235 & 864 & 458 \\
% Eye Movements & 0.5819 & 6704 & 1800 & 0.5940 & 784 & 406 \\
% \bottomrule
% \end{tabular}
% \end{table}

We compare TN with KAN and its variants, including ChebyKAN \cite{ss2024chebyshev}, FourierKAN \cite{fourierKAN}, and FastKAN \cite{li2024kolmogorov} under the following settings: For regression tasks, the dataset is partitioned into training and test sets in an 8:2 ratio. The activation function is Sigmoid with MSE as the loss function and RMSProp as the optimizer. For classification tasks, the same train-test split is applied, using the Sigmoid activation function. The loss function and optimizer are set to cross-entropy and Adam, respectively. All experiments are repeated 5 times and reported by the mean. We also report the number of parameters and FLOPs to reflect the computational cost. Notably, the reported FLOPs correspond to per-sample computation (i.e., batch size = 1), which provides a more accurate and fair measurement of the intrinsic computational complexity of each model, independent of batch-level implementation details. 

Since these datasets have already been evaluated in our previous experiments, where TN was compared with LN, we do not repeat all implementation details of task-based networks here. The task-based neurons used in this experiment are constructed from the symbolic formulas identified in Table~7 of the supplementary material. 

Table~\ref{tab:kan_vs_tn} show that TN achieves superior or highly competitive performance compared to KAN-based methods across both regression and classification tasks. For regression tasks (first eight datasets), TN significantly outperforms all three KAN variants in terms of MSE. The improvement is substantial across all datasets, with TN often reducing the error by a large margin (e.g., California Housing, Airfoil, Abalone, and Space GA). This indicates that the symbolic neuron design provides a much more effective function approximation mechanism than basis-function-based KAN variants. For classification tasks (last eight datasets), TN also demonstrates strong performance and achieves the best accuracy on multiple datasets, including Credit, Heloc, Phoneme, Vehicle, and Eye Movements. Even in cases where TN is not the top performer (e.g., Electricity or Magic Telescope), its performance remains highly competitive while maintaining a lower computational cost. Moreover, in most cases, TN requires fewer FLOPs and comparable or fewer parameters than KAN and its variants. This suggests that the improvement is not due to increased model capacity, but rather stems from a more expressive and task-adaptive neuron design.

\vspace{-0.3cm}

\subsection{Image Datasets}

\begin{figure}[!htb]
\vspace{-0.2cm}
	\centering
	\includegraphics[width=\linewidth]{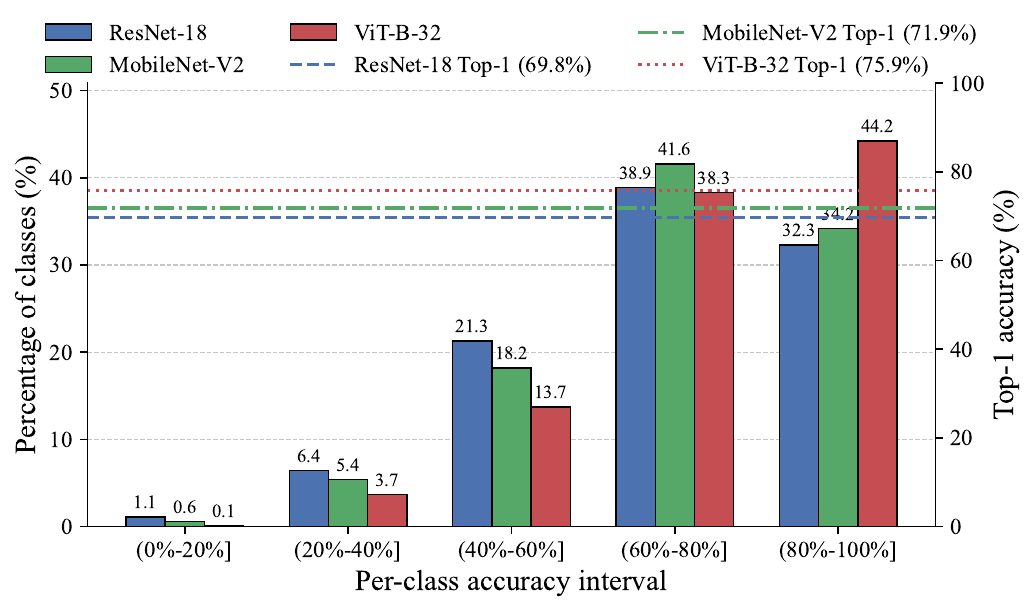}
	\caption{Distribution of prediction accuracy of different models.}
\label{fig:accuracy_distribution}
\vspace{-0.4cm}
\end{figure}

Generalization has long been regarded as the ultimate goal of machine learning, yet it is crucial to recognize that generalization itself is bounded both theoretically and practically. From a theoretical standpoint, a model with a finite number of parameters cannot possibly generalize across infinite tasks. Infinite tasks stem from the complexity of the real world, fine-grained nature of tasks, and so on.  Each model operates within a limited representational capacity, and its ability to generalize is therefore constrained by the scope of its hypothesis space. Let us take the classification on the ImageNet challenge as the example, which has 1,000 classes. We analyze class-wise accuracies of three representative backbone architectures—MobileNet-V2 (the lightweight network), ResNet-18 (the classic convolutional network), and ViT-B-32 (the classic attention network). As shown in Figure \ref{fig:accuracy_distribution}, three networks achieve good overall performance, however, ResNet-18 only achieves above 80\% accuracy on 32.3\% of classes, while the remaining 67.7\% of classes fall into lower accuracy ranges. Similarly, even ViT-B-32, a modern transformer-based architecture, fails to surpass 60\% accuracy on over 17.5\% of classes. This public example clearly demonstrates that it is hard for a generic model to achieve uniformly high performance across all categories. Moreover, regardless of retraining, corner cases always exist when the number of classes grows.

In this vein, task-driven neurons will be highly beneficial in terms of specifically addressing those cases. Moreover, it is superior to using another generic model on those cases because task-based neurons have the prior knowledge embedded into its aggregation function. For ResNet-18, MobileNet-V2, and ViT-B-32, we identify the 20 worst-performing classes based on the ImageNet validation set, respectively, and we specifically train the task-based neurons on these challenging cases. Because the original three networks are overly large for this 20-class small dataset, we prototype backbone structure as shown in Figure 9 of Section 6 in SMs. We replace the convolutional filters in the backbone network with task-based neurons.

The accuracy is reported in Table~\ref{results: imagenet}. Across all three backbones, task-based networks, though with fewer parameters, consistently outperform the corresponding counterpart using conventional neurons. These results strongly support the applicability of our neuron design strategy beyond the tabular domain. This also demonstrates that task-based neurons offer superior functional expressiveness and better task alignment compared to one-size-fits-all neuron designs.
\vspace{-0.4cm}
\begin{table}[htbp]
\centering
\caption{Top-$1$ validation accuracy on 20 worst-performing class of the ImageNet datasets by MobileNet-V2, ResNet-18, and ViT-B-32.}
\scalebox{0.8}{\begin{tabular}{lcccc}
\toprule
Method & \#Parameters& MobileNet-V2 & ResNet-18 & ViT-B-32 \\
\midrule
Conventional neurons & 0.38M & 71.9\% & 67.6\% & 68.6\% \\
Task-based neurons &0.37M & \textbf{74.8\%} & \textbf{74.7\%} & \textbf{71.3\%} \\
\bottomrule
\end{tabular}}
\label{results: imagenet}
\vspace{-0.3cm}
\end{table}

% \begin{table}[htbp]
% \centering
% \caption{Trainable parameter counts for different neuron settings.}
% \begin{tabular}{lccc}
% \toprule
% Method & MobileNetV2 & ResNet18 & ViT-B/32 \\
% \midrule
% Generic Neurons & 381,136 & 381,396 & 381,136 \\
% Task-based Neurons & \textbf{426,260} & \textbf{370,964} & \textbf{393,620} \\
% \bottomrule
% \end{tabular}
% \label{results: parameters}
% \end{table}

% \begin{figure*}[!htbp]
% \centering
%     \begin{adjustbox}{width=1\textwidth,center}
%         \begin{tabular}{ccc}
%             \subfloat[\label{fig:resnet_curve}]{\includegraphics[width=0.33\textwidth]{Figure/resnet_acc_vs_num_classes.pdf}} &
%             \subfloat[\label{fig:mobilenet_curve}]{\includegraphics[width=0.33\textwidth]{Figure/mobilenet_acc_vs_num_classes.pdf}} \\
%             \subfloat[\label{fig:vit_curve}]{\includegraphics[width=0.33\textwidth]{Figure/vit_acc_vs_num_classes.pdf}} 
            
%         \end{tabular}
%     \end{adjustbox}
%     \caption{
%        Validation accuracy of conventional and task-based networks on worst-class subsets identified by three different backbone models.
%     }
%     \label{fig:classes_curve}
% \end{figure*}

\begin{figure}[!htbp]
\centering
\includegraphics[width=1\linewidth]{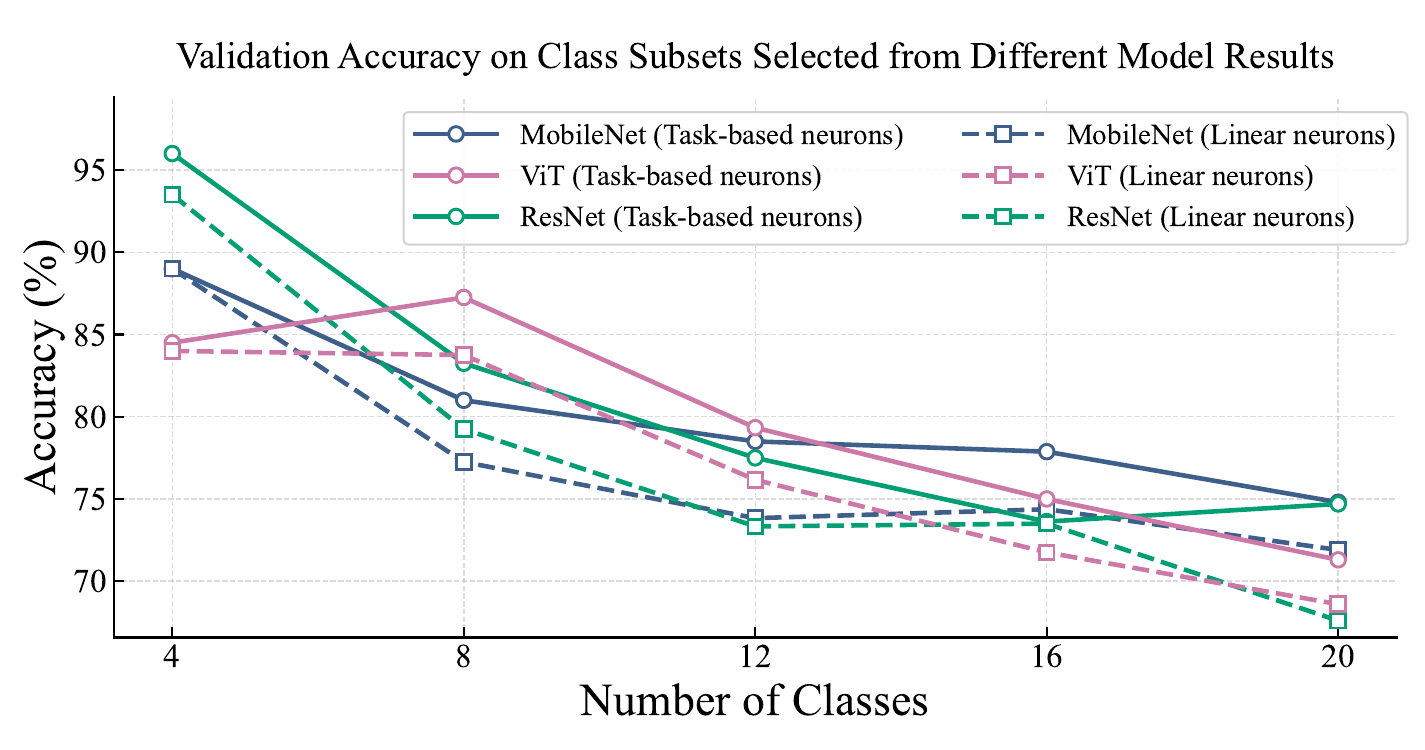}
\caption{Validation accuracy of conventional and task-based networks on worst-class subsets identified by three different backbone models.}
\label{fig:classes_curve}
\vspace{-0.3cm}
\end{figure}

To further examine how the complexity of classification tasks influences the effectiveness of task-based neurons, we design a controlled experiment based on the same 20-class subsets. Specifically, we progressively increase the number of classes: 4-class, 8-class, 12-class, 16-class, and 20-class classification. Each task is formed by selecting $k$ least accurate classes (with $k \in {4, 8, 12, 16, 20}$) from the 20-class subset. For every task, we train both the conventional and task-based networks under identical configurations. Figure~\ref{fig:classes_curve} summarizes the accuracy obtained on these progressively larger classification subsets (4, 8, 12, 16, and 20 classes).  
\vspace{-0.3cm}

\underline{MobileNet-V2-based subsets}:  
  Two models achieve comparable accuracy on the 4-class subset, and the task-based-neuron model consistently yields higher accuracy from 8 classes onward. The difference remains moderate across all settings, suggesting that the task-based formulation preserves accuracy better as the number of categories increases. \underline{ResNet-18-based subsets}:
  Both models exhibit a monotonic decline in accuracy as the number of classes increases, due to the rising task complexity.  
  The task-based model maintains higher accuracy across most configurations, with the gap narrowing near the 16-class setting and reappearing at 20 classes. \underline{ViT-B-32-based subsets}:
  Both models exhibit a consistent decline in accuracy as the number of evaluated classes increases.  Across all tested subsets, the task-based model achieves higher accuracy than its counterparts, and both follow a similar overall trend.

\vspace{-0.3cm}
\section{Conclusion and Future Work}

In this paper, towards NeuroAI, we have proposed the roadmap for task-based neurons via symbolic regression, which is a new frontier of neural network research compared to the architecture design. Systematic experiments over 10 synthetic datasets, 25 public datasets, and 2 real-world applications, have confirmed the potential of the task-based neuronal designs. In the future, on the one hand, the process of the VSR should be assiduously optimized. We can further investigate how to select suitable base functions for different scenarios to replace the simple symbolic regression algorithm in this paper. Also, from the perspective of algorithmic efficiency, the proposed VSR has not been facilitated by parallel acceleration and GPU acceleration, so the regression speed has room for improvement. 

%\section*{acknowledgement}
\vspace{-0.45cm}
\renewcommand\refname{Reference}

\ifCLASSOPTIONcaptionsoff
  \newpage
\fi



%\bibliographystyle{IEEEtran}

\bibliographystyle{ieeetr}
\bibliography{reference.bib}


% \end{IEEEbiography}


% % insert where needed to balance the two columns on the last page with
% %\newpage



%\vfill


\end{document}

\title{Supplementary Materials of “No One-Size-Fits-All Neurons: Task-based Neurons for Artificial Neural Networks"}

\author{Feng-Lei Fan$^\dag$, Meng Wang$^\dag$, Hang-Cheng Dong, Jianwei Ma$^*$, Tieyong Zeng$^*$

% <-this % stops a space
\IEEEcompsocitemizethanks{
% \IEEEcompsocthanksitem *Tieyong Zeng, Yiming Cui, and Jing-Xiao Liao are co-corresponding authors.
% % note need leading \protect in front of \\ to get a newline within \thanks as
% % \\ is fragile and will error, could use \hfil\break instead.
% E-mail: see http://www.michaelshell.org/contact.html

\IEEEcompsocthanksitem Feng-Lei Fan and Meng Wang are co-first authors. Jianwei Ma and Tieyong Zeng are co-corresponding authors. The research is supported by the National Natural Science Foundation of China (nos. 42230806and U23B6010), the China National Petroleum Corporation-Peking University Strategic Cooperation Project of Fundamental Research, the Lagrange Mathematics and Computing Research Center, Huawei Technologies France, and the Startup Fund of the City University of Hong Kong.

\IEEEcompsocthanksitem Feng-Lei Fan is with Department of Data Science, City University of Hong Kong, Kowloon, Hong Kong. 

\IEEEcompsocthanksitem 
Meng Wang is with School of Mathematics, Harbin Institute of Technology, Harbin, Heilongjiang Province 150001, China.

\IEEEcompsocthanksitem 
Hang-Cheng Dong is with School of Instrumentation, Harbin Institute of Technology, Harbin, Heilongjiang Province 150001, China.

\IEEEcompsocthanksitem 
Jianwei Ma is with School of Mathematics, Harbin Institute of Technology, Harbin, China, and School of Earth and Space Sciences, Peking University, Beijing, China.

\IEEEcompsocthanksitem Tieyong Zeng is with Institute for Advanced Study, Beijing Normal-Hong Kong Baptist University, Zhuhai, Guangdong, China.

}
% <-this % stops an unwanted space
}

\markboth{Journal of \LaTeX\ Class Files,~Vol.~14, No.~8, August~2023}%
{Shell \MakeLowercase{\textit{et al.}}: Bare Demo of IEEEtran.cls for Computer Society Journals}



\IEEEtitleabstractindextext{%

\begin{IEEEkeywords}
Machine learning, NeuroAI, neuronal diversity, symbolic regression, task-based neurons 
\end{IEEEkeywords}}

\maketitle

\IEEEdisplaynontitleabstractindextext

% \fi
\IEEEpeerreviewmaketitle


\begin{table*}
\caption{Formulas learned by the vectorized symbolic regression with respect to different noise levels.}
    \centering
    \scalebox{0.8}{\begin{tabular}{l|l|l}
        \hline
\makecell[c]{\multirow{2}{*}{True Function}} & 
\multicolumn{2}{c}{Predicted Function} \\ \cline{2-3}
& \makecell[c]{$\epsilon=0\% $} & \makecell[c]{$\epsilon=5\% $} \\
   
        \hline 

$p_1(\bm{x})=\bm{6} (\x \odot \x)^\top$  & $\bm{6} (\x \odot \x)^\top$ & $\bm{6} (\x \odot \x)^\top$   \\ 

$p_2(\bm{x})=\bm{3}(\x \odot^4 \x)^\top$ & $\bm{3}(\x \odot^4 \x)^\top$  & $\bm{3}(\x \odot^4 \x)^\top - \bm{14}(\x\odot\x)^\top-\bm{88} \x^\top +106$\\ 
 
$ p_3(\bm{x})=\bm{4}(\x\odot\x)^\top + \bm{5} \x^\top $  & $\bm{4}(\x\odot\x)^\top + \bm{5} \x^\top$  & $\bm{4}(\x\odot\x)^\top + \bm{5} \x^\top-42$ \\ 

$ p_4(\bm{x})=\bm{2}(\x\odot^4\x)^\top + \bm{6}\x^\top $  & $\bm{2}(\x\odot^4\x)^\top + \bm{6}\x^\top$ & $\bm{2}(\x\odot^4\x)^\top- \bm{4}(\x\odot\x)^\top  - \bm{387}\x^\top -14344$ \\ 

$ p_5(\bm{x})=\bm{2}(\x\odot^4\x)^\top + \bm{3} (\x\odot\x)^\top + \bm{6}\x^\top$ & $\bm{2}(\x\odot^4\x)^\top + \bm{3} (\x\odot\x)^\top + \bm{6}\x^\top$ & $\bm{2}(\x \odot^4 \x)^\top + \bm{12}(\x \odot \x)^\top - \bm{547}\x^\top + 3162$  \\ 

$ p_6(\bm{x})=\bm{3}(\x\odot^4\x)^\top + \bm{4}(\x\odot^3\x)^\top  + \bm{2}(\x\odot\x)^\top + \bm{3}\x^\top$ & $\bm{3}(\x\odot^4\x)^\top + \bm{4}(\x\odot^3\x)^\top  + \bm{2}(\x\odot\x)^\top + \bm{3}\x^\top$ & $\bm{3}(\x\odot^4\x)^\top + \bm{2}(\x\odot^3\x)^\top - \bm{3}(\x\odot\x)^\top + \bm{3243}\x^\top - 10331$\\ 

$p_7(\bm{x})=\bm{4}(\x\odot^4 \x)^\top +\bm{2}(\x\odot\x)^\top$ & $\bm{4}(\x\odot^4 \x)^\top +\bm{2}(\x\odot\x)^\top$ & $\bm{4}(\x\odot^4\x)^\top + \bm{4}(\x\odot^3\x)^\top + \bm{14}(\x\odot\x)^\top -\bm{6287}\x^\top -729$ \\ 

$p_8(\bm{x})=\bm{2}(\x\odot^4 \x)^\top + \bm{3}(\x\odot^3\x)^\top$ & $\bm{2}(\x\odot^4 \x)^\top + \bm{3}(\x\odot^3\x)^\top$ & $\bm{2}(\x\odot^4\x)^\top + \bm{3}(\x\odot^3\x)^\top+\bm{2}(\x\odot\x)^\top -\bm{264}\x^\top+78$\\ 

$p_9(\bm{x})=\bm{2}(\x\odot^3 \x)^\top +\bm{3}\x^\top$ & $\bm{2}(\x\odot^3 \x)^\top +\bm{3}\x^\top$ & $\bm{2}(\x\odot^3\x)^\top +\bm{12}\x^\top+282$ \\ 

$p_{10}(\bm{x})=\bm{3}(\x\odot^5\x)^\top +\bm{4}(\x\odot^4\x)^\top +\bm{5}\x^\top$ & $\bm{3}(\x\odot^5\x)^\top +\bm{4}(\x\odot^4\x)^\top +\bm{5}\x^\top$  & $\bm{3}(\x\odot^5\x)^\top + \bm{4}(\x\odot^4\x)^\top+\bm{6}(\x\odot^3\x)^\top+\bm{13}(\x\odot\x)^\top+\bm{22}\x^\top$\\  \hline \hline
 
\makecell[c]{\multirow{1}{*}{True Function}}& \makecell[c]{\multirow{1}{*}{$\epsilon=15\%$}} & \makecell[c]{\multirow{1}{*}{$\epsilon=30\% $}} \\ \hline

$p_1(\bm{x})=\bm{6} (\x \odot \x)^\top$ & $ \bm{6} (\x \odot \x)^\top +\bm{5} \x^\top + 11$ &  $\bm{6} (\x \odot \x)^\top+\bm{10} \x^\top + 13$   \\ 

$p_2(\bm{x})=\bm{3}(\x \odot^4 \x)^\top$ & $\bm{3}(\x \odot^4 \x)^\top - \bm{42}(\x \odot \x)^\top - \bm{441}\x^\top +1431$  & $\bm{3}(\x \odot^4 \x)^\top - \bm{3}(\x 
 \odot^3 \x)^\top - \bm{98}(\x \odot \x)^\top + \bm{3488} \x^\top -131$ \\ 
 
$ p_3(\bm{x})=\bm{4}(\x\odot\x)^\top + \bm{5} \x^\top $  & $\bm{4}(\x\odot\x)^\top + \bm{6} \x^\top-174$ & $\bm{4}(\x\odot\x)^\top+\bm{6}\x^\top-383$ \\ 

$ p_4(\bm{x})=\bm{2}(\x\odot^4\x)^\top + \bm{6}\x^\top $ & $\bm{2}(\x\odot^4\x)^\top - \bm{9}(\x\odot\x)^\top - \bm{1688}\x^\top - 70667$ & $\bm{2}(\x\odot^4\x)^\top - \bm{35}(\x \odot \x)^\top - \bm{2859}\x^\top -36$ \\ 

$ p_5(\bm{x})=\bm{2}(\x\odot^4\x)^\top + \bm{3} (\x\odot\x)^\top + \bm{6}\x^\top$ &\makecell[l]{$\bm{2}(\x \odot^4 \x)^\top  + (\x \odot^3 \x)^\top + \bm{21}(\x \odot \x)^\top $ \\ $ -\bm{2021}\x^\top +7564$} & $\bm{2}(\x \odot^4 \x)^\top + \bm{6}(\x \odot^3 \x)^\top + \bm{46}(\x \odot \x)^\top - \bm{12055}\x^\top - 559$ \\ 

$ p_6(\bm{x})=\bm{3}(\x\odot^4\x)^\top + \bm{4}(\x\odot^3\x)^\top  + \bm{2}(\x\odot\x)^\top + \bm{3}\x^\top$  & $\bm{3}(\x\odot^4\x)^\top - \bm{11}(\x\odot\x)^\top +\bm{10980}\x^\top +155$ & $\bm{3}(\x\odot^4\x)^\top + \bm{3}(\x\odot^3\x)^\top  -\bm{74}(\x\odot\x)^\top +\bm{3405}\x^\top -25168$ \\ 

$p_7(\bm{x})=\bm{4}(\x\odot^4 \x)^\top +\bm{2}(\x\odot\x)^\top$ & \makecell[l]{$\bm{4}(\x\odot^4\x)^\top  + \bm{4}(\x\odot^3\x)^\top +\bm{36}(\x\odot\x)^\top $ \\ $-\bm{4378}\x^\top-319$} & $\bm{4}(\x\odot^4\x)^\top+\bm{12}(\x\odot^3\x)^\top+\bm{76}(\x\odot\x)^\top -\bm{17838}\x^\top+126$ \\ 

$p_8(\bm{x})=\bm{2}(\x\odot^4 \x)^\top + \bm{3}(\x\odot^3\x)^\top$ & \makecell[l]{$\bm{2}(\x\odot^4\x)^\top + \bm{3}(\x\odot^3\x)^\top  + \bm{7}(\x\odot\x)^\top $ \\ $- \bm{668}\x^\top + 12332$} & $\bm{2}(\x\odot^4\x)^\top + \bm{4}(\x\odot^3\x)^\top + \bm{20}(\x\odot\x)^\top - \bm{1255}\x^\top - 8$ \\  

$p_9(\bm{x})=\bm{2}(\x\odot^3 \x)^\top +\bm{3}\x^\top$ & $\bm{2}(\x\odot^3\x)^\top -(\x\odot\x)^\top +\bm{35}\x^\top +2713$ & $\bm{2}(\x\odot^3\x)^\top +\bm{65}\x^\top+2376$ \\ 

$p_{10}(\bm{x})=\bm{3}(\x\odot^5\x)^\top +\bm{4}(\x\odot^4\x)^\top +\bm{5}\x^\top$ & \makecell[l]{$\bm{3}(\x\odot^5\x)^\top + \bm{4}(\x\odot^4\x)^\top+\bm{4}(\x\odot^3\x)^\top $ \\ $-\bm{240}(\x\odot\x)^\top-\bm{196}\x^\top-2940$} & \makecell[l]{$\bm{3}(\x\odot^5\x)^\top + \bm{6}(\x\odot^4\x)^\top +\bm{21}(\x\odot^3\x)^\top$ \\ $-\bm{3523}(\x\odot^2\x)^\top+\bm{678}\x^\top+2325$} \\
\hline
    \end{tabular}  
    }
    \label{tab:1}
\end{table*}

\section{Analysis Experiments}
In this section, we supplement two experimental results: one is to validate the feasibility of the proposed methodology of the task-based neurons and the other is to compare the performance of task-based neurons with quadratic neurons. 

\subsection{Feasibility of Task-based Neurons}

Here, we conduct experiments to validate the feasibility of the proposed task-based neurons, with an emphasis on whether or not the vectorized symbolic regression can identify the appropriate formulas from noisy data and the impact of noise. We first make synthetic data which are clean signals perturbed by noise. Then, we apply the vectorized symbolic regression to the synthetic data to learn the mathematical formula. Next, we use the learned formula as the basic neuron to build a task-based network. Through this experiment, we can verify whether the task-based network could achieve lower regression errors while utilizing fewer parameters compared to a linear network. 

\begin{table}[htb]
\caption{The error rates of the symbolic regression for clean and noisy data.}
    \centering
    \scalebox{0.8}{\begin{tabular}{c|c|c|c|c}
    \toprule  
    % \hline
    True Function & $\epsilon=0\% $ & $\epsilon=5\% $ & $\epsilon=15\% $ & $\epsilon=30\%$
    \\ \hline
$p_1(\bm{x})$ & $0.00\%$ & $0.30\%$ & $0.75\%$ & $1.51\%$ \\ 
       $p_2(\bm{x})$ &  $0.00\%$ & $0.34\%$ & $1.00\%$ & $2.37\%$ \\ 
       $p_3(\bm{x})$ &  $0.00\%$ & $0.14\%$ & $0.59\%$ & $1.28\%$ \\ 
       $p_4(\bm{x})$ &  $0.00\%$ & $0.23\%$ & $0.79\%$ & $1.44\%$ \\ 
       $p_5(\bm{x})$ &  $0.00\%$ & $0.37\%$ & $0.90\%$ & $2.35\%$ \\ 
       $p_6(\bm{x})$ &  $0.00\%$ & $0.35\%$ & $1.44\%$ & $1.96\%$ \\ 
       $p_7(\bm{x})$ &  $0.00\%$ & $0.51\%$ & $0.71\%$ & $1.80\%$ \\ 
       $p_7(\bm{x})$ & $ 0.00\%$ & $0.11\%$ & $0.37\%$ & $0.73\%$ \\ 
      $p_9(\bm{x})$ &   $0.00\%$ & $0.68\%$ & $5.60\%$ & $4.81\%$ \\ 
       $p_{10}(\bm{x})$ &  $0.00\%$ & $0.15\%$ & $1.26\%$ & $4.26\%$ \\ \hline
  
    \end{tabular}}
    \label{tab:reg_error}
\end{table}

\textbf{Synthetic data generation}. 10 multivariate polynomials are constructed, as shown in Table \ref{tab:1}. For simplicity and conciseness, we prescribe 
\begin{equation}
    \sum_{i=1}^n rx_i^s = \bm{r}\cdot {\underbrace{(\x\odot\cdot \odot \x)}_{\text{s ~times}}}^\top = \bm{r}\cdot (\x\odot^s\x)^\top,
\end{equation}
where $\bm{r}=[r,r,\cdots,r] \in \mathbb{R}^d$, and $\odot$ is a Hadarmad product. Moreover, we add three levels of noise into polynomials to test the noise-resistant learning ability of the vectorized symbolic regression. After obtaining clean data, low ($\epsilon=5\%$), medium ($\epsilon=15\%$), and high ($\epsilon=30\%$) levels of noise are added by using the following formula

\begin{equation}
\bar{\bm{y}} = \bm{y} + \epsilon E(\bm{y})\cdot\bm{n}, 
\end{equation}
where $\bm{y}$ is the output of some polynomial, $\bar{\bm{y}}$ is the noisy output, $\bm{n}$ is a vector whose elements obey $\mathcal{G}(0,1)$, $\epsilon$ stands for noise level, and $E(\bm{y})$ represents the power of the data. $E(\bm{y})$ is calculated as $E(\bm{y}) =  \sqrt{\frac{1}{N} \sum_{i=1}^{N} \bm{y}_i^2}$, where $N$ is the number of data points. Here, we uniformly sample 600 data points in the interval $[-50,50]$.

\begin{figure}[htb]
\vspace{-0.3cm}
  \flushleft{ 
\includegraphics[width=\linewidth]{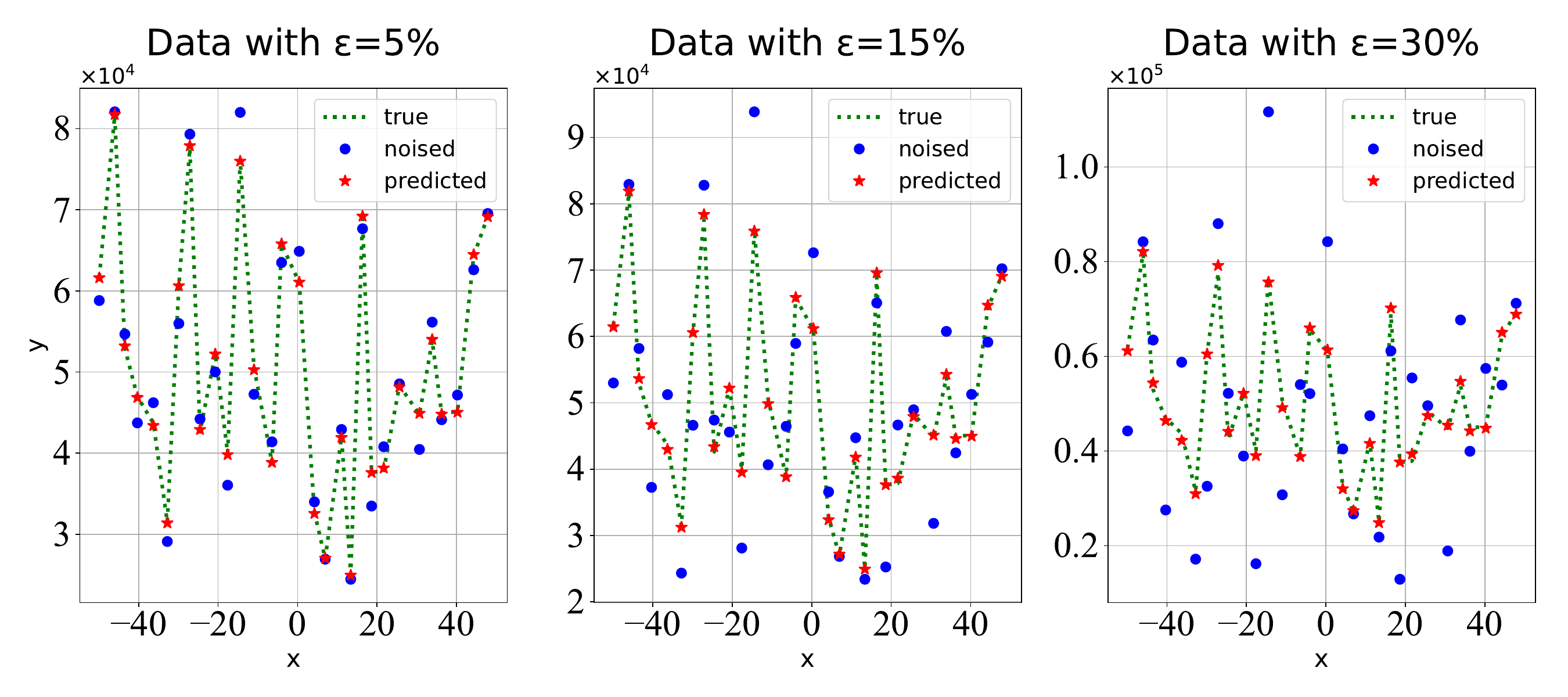} 
  \caption{The visualization of fitting performance by formulas learned by the proposed vectorized symbolic regression with respect to the polynomial $p_1(\x)$ under three different noise levels.}
  \label{fig:data_error}
}

\vspace{-0.2cm}   
    
\end{figure}

% \begin{figure}[htb]
% \vspace{-0.4cm}
%   \centering
% \subfigure[No.1 dataset with low noise]{\includegraphics[width=0.3\linewidth]{Figure/no1_dataset_with_low_noise_new.png}} 
% \subfigure[No.1 dataset with medium noise]{\includegraphics[width=0.3\linewidth]{Figure/no1_dataset_with_medium_noise_new.png}} 
%  \subfigure[No.1 dataset with high noise]{\includegraphics[width=0.3\linewidth]{Figure/no1_dataset_with_high_noise_new.png}}
%  \\

%   \caption{The visualization of fitting performance by formulas learned by the proposed vectorized symbolic regression.}
%     \label{fig:data_error}
%     \vspace{-0.1cm}
% \end{figure}

\begin{table*}[htb]
\caption{Comparing the linear and task-based neurons in five different network structures for synthetic data. The number $h_1$-$h_2$-$\cdots$-$h_k$-$1(z)$ means that this network has $k$ hidden layers, each with $h_k$ neurons, and $z$ is the number of parameters used in such a network. For example, 5-3-1 (145) means this network has two hidden layers with 5 and 3 neurons respectively in each layer, and 145 parameters.}
    \centering
    \scalebox{0.8}{\begin{tabular}{c|ll|ll|ll|ll|ll}
    \toprule    
     \multirow{2}{*}{No.} & \multicolumn{2}{c}{S1} & \multicolumn{2}{c}{S2} & \multicolumn{2}{c}{S3} & \multicolumn{2}{c}{S4}& \multicolumn{2}{c}{S5} \\ 
\cline{2-11}
& LN & TN & LN & TN & LN & TN & LN & TN & LN & TN\\ \hline
    $p_1(\bm{x})$ & 2-1(25) & 1(21) & 5-1(61) & 2-1(47) & 5-3-1(77) & 3-1(70) & 10-5-3-1(187) & 5-3-1(145) & 12-8-5-3-1(303) & 8-5-3-1(293) \\ \hline
    $p_2(\bm{x})$ & 3-1(37) & 1(31) & 6-1(73) & 2-1(69) & 7-5-1(123) & 3-1(103) & 10-8-5-1(249) & 5-3-1(213) & 12-10-6-3-1(353) & 6-4-2-1(295) \\ \hline
    $p_3(\bm{x})$ & 2-1(25) & 1(21) & 5-1(61) & 2-1(47) & 5-3-1(77) & 3-1(70) & 10-5-3-1(187) & 5-3-1(145) & 12-8-5-3-1(303) & 8-5-3-1(293) \\ \hline
    $p_4(\bm{x})$ & 3-1(37) & 1(31) & 6-1(73) & 2-1(69) & 7-5-1(123) & 3-1(103) & 10-8-5-1(249) & 5-3-1(213) & 12-10-6-3-1(353) & 6-4-2-1(295) \\ \hline
    $p_5(\bm{x})$ & 3-1(37) & 1(31) & 6-1(73) & 2-1(69) & 7-5-1(123) & 3-1(103) & 10-8-5-1(249) & 5-3-1(213) & 12-10-6-3-1(353) & 6-4-2-1(295) \\ \hline
    $p_6(\bm{x})$ & 4-1(49) & 1(41) & 8-1(97) & 2-1(91) & 8-5-1(139) & 3-1(136) & 12-8-5-1(287) & 5-3-1(281) & 12-10-8-5-1(401) & 6-4-2-1(389) \\ \hline
    $p_7(\bm{x})$ & 4-1(49) & 1(41) & 8-1(97) & 2-1(91) & 8-5-1(139) & 3-1(136) & 12-8-5-1(287) & 5-3-1(281) & 12-10-8-5-1(401) & 6-4-2-1(389) \\ \hline
    $p_8(\bm{x})$ & 4-1(49) & 1(41) & 8-1(97) & 2-1(91) & 8-5-1(139) & 3-1(136) & 12-8-5-1(287) & 5-3-1(281) & 12-10-8-5-1(401) & 6-4-2-1(389) \\ \hline
    $p_9(\bm{x})$ & 2-1(25) & 1(21) & 5-1(61) & 2-1(47) & 5-3-1(77) & 3-1(70) & 10-5-3-1(187) & 5-3-1(145) & 12-8-5-3-1(303) & 8-5-3-1(293) \\ \hline
    $p_{10}(\bm{x})$ & 5-1(61) & 1(51) & 10-1(121) & 2-1(113) & 10-5-1(171) & 3-1(169) & 12-10-8-1(359) & 5-3-1(349) & 13-10-8-5-1(422) & 5-4-2-1(412) \\   
    
    \bottomrule 
    \end{tabular}}
    \label{tab:syn_para}
\end{table*}

\begin{table*}[htb]
\caption{The mean squared errors of task-based networks and the linear networks of five structures.}    
    \centering
    \scalebox{0.75}{\begin{tabular}{c|cc|cc|cc|cc|cc}
    % \multicolumn{11}{r}{\small $\bullet$ denotes the best performance value for each data set.}
    \toprule  
\multirow{2}{*}{No.} & \multicolumn{2}{c}{S1} & \multicolumn{2}{c}{S2} & \multicolumn{2}{c}{S3} & \multicolumn{2}{c}{S4}& \multicolumn{2}{c}{S5} \\ 
\cline{2-11}
   & LN & TN & LN & TN & LN & TN & LN & TN & LN & TN\\
    \hline

$p_{1}(\bm{x})$ & 0.0215 (0.0013) & \textbf{0.0009 (0.0002)} & 0.0212 (0.0020) & \textbf{0.0008 (0.0002)} & 0.0217 (0.0023) & \textbf{0.0009 (0.0002)} & 0.0208 (0.0018) & \textbf{0.0009 (0.0002)} & 0.0207 (0.0023) & \textbf{0.0010 (0.0002)} \\ \hline
$p_{2}(\bm{x})$ & 0.0234 (0.0018) & \textbf{0.0040 (0.0003)} & 0.0238 (0.0012) & \textbf{0.0039 (0.0003)} & 0.0239 (0.0029) & \textbf{0.0038 (0.0002)} & 0.0219 (0.0014) & \textbf{0.0038 (0.0003)} & 0.0221 (0.0012) & \textbf{0.0040 (0.0002)} \\ \hline
$p_{3}(\bm{x})$ & 0.0253 (0.0015) & \textbf{0.0013 (0.0003)} & 0.0250 (0.0022) & \textbf{0.0010 (0.0002)} & 0.0255 (0.0022) & \textbf{0.0010 (0.0001)} & 0.0240 (0.0013) & \textbf{0.0012 (0.0003)} & 0.0228 (0.0032) & \textbf{0.0014 (0.0004)} \\ \hline
$p_{4}(\bm{x})$ & 0.0238 (0.0015) & \textbf{0.0038 (0.0003)} & 0.0224 (0.0016) & \textbf{0.0037 (0.0002)} & 0.0233 (0.0023) & \textbf{0.0036 (0.0002)} & 0.0224 (0.0028) & \textbf{0.0037 (0.0003)} & 0.0213 (0.0012) & \textbf{0.0037 (0.0003)} \\ \hline
$p_{5}(\bm{x})$ & 0.0218 (0.0017) & \textbf{0.0040 (0.0004)} & 0.0219 (0.0017) & \textbf{0.0038 (0.0003)} & 0.0223 (0.0028) & \textbf{0.0038 (0.0002)} & 0.0190 (0.0021) & \textbf{0.0037 (0.0002)} & 0.0199 (0.0022) & \textbf{0.0039 (0.0003)} \\ \hline
$p_{6}(\bm{x})$ & 0.0212 (0.0009) & \textbf{0.0037 (0.0003)} & 0.0212 (0.0019) & \textbf{0.0037 (0.0002)} & 0.0205 (0.0012) & \textbf{0.0035 (0.0003)} & 0.0180 (0.0027) & \textbf{0.0036 (0.0005)} & 0.0191 (0.0017) & \textbf{0.0041 (0.0008)} \\ \hline
$p_{7}(\bm{x})$ & 0.0236 (0.0024) & \textbf{0.0043 (0.0003)} & 0.0227 (0.0028) & \textbf{0.0041 (0.0004)} & 0.0237 (0.0018) & \textbf{0.0039 (0.0001)} & 0.0232 (0.0030) & \textbf{0.0042 (0.0007)} & 0.0236 (0.0019) & \textbf{0.0046 (0.0010)} \\ \hline
$p_{8}(\bm{x})$ & 0.0189 (0.0015) & \textbf{0.0035 (0.0003)} & 0.0188 (0.0015) & \textbf{0.0034 (0.0002)} & 0.0203 (0.0025) & \textbf{0.0033 (0.0003)} & 0.0175 (0.0028) & \textbf{0.0035 (0.0006)} & 0.0181 (0.0009) & \textbf{0.0039 (0.0007)} \\ \hline
$p_{9}(\bm{x})$ & 0.0136 (0.0105) & \textbf{0.0058 (0.0000)} & 0.0060 (0.0006) & \textbf{0.0056 (0.0002)} & 0.0084 (0.0072) & \textbf{0.0056 (0.0001)} & 0.0185 (0.0107) & \textbf{0.0056 (0.0002)} & 0.0199 (0.0114) & \textbf{0.0058 (0.0003)} \\ \hline
$p_{10}(\bm{x})$ & 0.0088 (0.0010) & \textbf{0.0024 (0.0003)} & 0.0095 (0.0008) & \textbf{0.0022 (0.0007)} & 0.0117 (0.0050) & \textbf{0.0017 (0.0004)} & 0.0103 (0.0020) & \textbf{0.0020 (0.0006)} & 0.0110 (0.0037) & \textbf{0.0023 (0.0009)} \\ 

\bottomrule 

    \end{tabular}}
    \label{tab:syn_res}
    \vspace{-0.2cm}
\end{table*}

\textbf{Experimental setups of the vectorized symbolic regression}. To ensure reproducibility, here we explicitly specify experimental setups. The random seed is set to 0. The depth of the tree that encodes formulas is set to no more than 6 to prevent the program from learning over-complicated formulas. The initial population size is set to 500, and the maximum number of generations is 80. The tournament size is 3\% of the population size, which represents the number of winning individuals selected to produce the next generation. The crossover probability is set at 30\%, the mutation probability at 60\%, and the reproduction probability at 10\%. The top 5\% of the dominant individuals in each generation are selected to generate the next generation. The range of the returned numeric symbols is set to $[-20, 20]$, following a uniform distribution. The mean squared error (MSE) serves as a fitness function.

\textbf{Results of the vectorized symbolic regression}. The learned mathematical expressions with respect to different noise levels are shown in Table \ref{tab:1}. 
The corresponding regression errors by the vectorized symbolic regression are shown in Table \ref{tab:reg_error}. The error rate is calculated as 
\begin{equation}
  \delta =  \frac{1}{600} \sum_{i=1}^{600} \frac{|\bm{y}_i-\tilde{\bm{y}}_i|}{|\bm{y}_i|},  
\end{equation}
where $\bm{y}$ is the truth, and $\tilde{\bm{y}}$ is the prediction. 
Per Table \ref{tab:reg_error}, the vectorized symbolic regression can learn the patterns behind data well, even when data are highly polluted, which suggests the effectiveness of the proposed vectorized symbolic regression.

% \begin{figure*}[!htb]
%   \centering

% \subfigure[No.1 dataset with low noise]{\includegraphics[width=0.3\linewidth]{Figure/no1_dataset_with_low_noise_.png}} 
% \subfigure[No.1 dataset with medium noise]{\includegraphics[width=0.3\linewidth]{Figure/no1_dataset_with_medium_noise_.png}} 
%  \subfigure[No.1 dataset with high noise]{\includegraphics[width=0.3\linewidth]{Figure/no1_dataset_with_high_noise_.png}}
%  \\

% \subfigure[No.6 dataset with low noise]{\includegraphics[width=0.3\linewidth]{Figure/no9_dataset_with_low_noise_.png}} 
% \subfigure[No.6 dataset with medium noise]{\includegraphics[width=0.3\linewidth]{Figure/no9_dataset_with_medium_noise_.png}} 
%  \subfigure[No.6 dataset with high noise]{\includegraphics[width=0.3\linewidth]{Figure/no9_dataset_with_high_noise_.png}}
%  \\

%  \subfigure[No.10 dataset with low noise]{\includegraphics[width=0.3\linewidth]{Figure/no14_dataset_with_low_noise_.png}} 
% \subfigure[No.10 dataset with medium noise]{\includegraphics[width=0.3\linewidth]{Figure/no14_dataset_with_medium_noise_.png}} 
%  \subfigure[No.10 dataset with high noise]{\includegraphics[width=0.3\linewidth]{Figure/no14_dataset_with_high_noise_.png}}

%   \caption{Datasets with different noise levels}
%     \label{fig:data_error}
%     \vspace{0.2in}
% \end{figure*}

% \begin{figure}[htb] 
% \centering
% \includegraphics[width=0.7\linewidth]{Figure/Figure_vectorized.pdf}
% \caption{Vectorized symbolic regression.} 
% \label{vector}
% \end{figure}

\textbf{Setups of task-based networks}. Now, we use the expression shown in Table \ref{tab:1} to build a neuron and the corresponding neural network to fit the synthetic data again. The activation function is ReLU. We compare the performance of the linear neuron and task-based neurons in five different structures from simple to complex. The network structure is the fully connected network. The network structure used and the corresponding number of network parameters are shown in Table \ref{tab:syn_para}. In every structure, a task-based network uses fewer parameters. Thus, the task-based network enjoys model efficiency. Each data set is divided into a training set and a test set by a ratio of $8:2$. Two types of networks use \textit{MSELoss} as the loss function and \textit{RMSProp} as the optimizer. We run ten times on each synthetic dataset to avoid the effects of randomness. The MSE is the evaluation metric, and the test results are presented in the form of $\mathrm{mean}~ (\mathrm{std})$.

\textbf{Performance of task-based networks}. Results are shown in Table \ref{tab:syn_res}. It can be seen that even if the number of parameters in \textit{TN} does not exceed \textit{LN} and the network depth is shallower than \textit{LN}, the mean and standard deviation of errors in the synthetic data of \textit{LN} are larger than \textit{TN}, which indicates that \textit{TN} is more powerful and efficient than \textit{LN}. Figure \ref{fig:data_error} visualizes the fitting performance by formulas learned by the proposed vectorized symbolic regression with respect to the polynomial $p_1(\x)$ under three different noise levels. It can be seen that the vectorized symbolic regression can capture the suitable formulas to fit the curve.

This experiment also confirms that the learned formula, and consequently the learned neuron, is indeed basic. If the learned neuron were non-basic, it would lead to a potential misalignment between the function space of task-based networks and the inherent characteristics of the data. In such a scenario, connecting task-based neurons into a deep network could potentially hinder performance. However, in practice, we observe that deep task-based networks perform well for all synthetic signals. This observation is crucial in evaluating the compatibility between task-based neurons and deep networks.

\vspace{-0.4cm}
\subsection{Necessity of Task-based Neurons}
\vspace{-0.1cm}

One may argue that instead of using the vectorized symbolic regression to construct an elementary neuron, why not directly use symbolic regression to fit data? The reason is that symbolic regression is not good at handling high-dimensional data.
Constructing a neuron is to leverage connectionism, which has been proven to be a powerful approach by numerous successes of deep learning.

% Linear regression has a better fitting effect on data with linear correlation, but if the data is relatively complex and the traditional symbolic regression cannot learn the inherent law of the data, the error of fitting the data with a straight line is usually smaller. Therefore, you can see from Table \ref{tab:compare_reg} that \textit{LR} has a smaller mean square error than \textit{SR}. \textit{VSR} tends to learn the patterns in the data, and does not seek to minimize the data fitting error, so the error of fitting the data directly with \textit{VSR} is the largest. However, using the pattern learned by \textit{VSR} to build a neural network can combine the advantages of both \textit{VSR} and neural network, so it can be seen that the fitting error of \textit{TN} is minimal.

To validate these points, we perform \textit{TN}, traditional symbolic regression (\textit{SR}), and vectorized symbolic regression (\textit{VSR}) on five publicly available tabular datasets and compare their performance. For \textit{SR} and \textit{VSR}, prediction on the test set is made directly using the expression obtained by \textit{SR} and \textit{VSR}, respectively. For \textit{TN}, we no longer perform the vectorized symbolic regression; instead, we use the mathematical expression obtained from VSR to build a neuron. The network structure is the fully connected network.

To eliminate the effect of large differences in the order of magnitude of features in the data, we do the normalization for features. The training and test sets are divided according to the ratio of $8:2$, and MSE is the evaluation index. We use machine learning library \textit{gplearn}\footnote{https://gplearn.readthedocs.io/en/stable/intro.html} to implement symbolic regression, \textit{gplearn} is a mature symbolic regression library based on Python, with good stability and superior performance. Traditional symbolic regression deals with each input variable individually. To implement the vectorized symbolic regression, we just need to feed the input vector into the symbolic regression in \textit{gplearn}. 

We set the fixed random seed to ensure that the symbolic regression can be repeated. For important hyperparameters, we finetune them to achieve the optimal performance for \textit{SR} and \textit{VSR}. Finally, in the traditional \textit{SR}, the population size is set to 5,000; the max generation is 20; the crossover probability is set at 55\%; the mutation probability at 40\%; the reproduction probability is set at 5\%. In \textit{VSR}, the random seed is set to 100; the population size is set to 5,000; the max generation is set to 30; the crossover probability is set at 30\%; the mutation probability at 60\%; the reproduction probability at 10\%; the tournament size is 3\% of the population size. The range of numeric symbols returned is set to $[-1,1]$, following a uniform distribution. 

\vspace{-0.3cm}
\begin{table}[htb!]
\caption{Test results of the symbolic regression (SR), the VSR, and task-based networks (TN).}
\vspace{-0.1cm}
    \centering
    \scalebox{0.8}{\begin{tabular}{lccccc}
    \toprule  
 Dataset & \#Instances & \#Features & SR &  VSR & TN (Strcture) \\ \hline
google stock price & 12,454 & 4 & 0.0362  &  0.5505  & \textbf{0.0319} (4-1) \\
concrete strength  & 10,308 & 8 & 0.4617  &  0.7914  & \textbf{0.1144} (4-1) \\ 
health insurance & 13,386 & 6 & 0.3072    & 0.7009  & \textbf{0.1849} (3-1) \\ 
white wine quality & 395,611 & 11 & 0.7430  &  0.9288  & \textbf{0.6174} (4-1) \\ 
song popularity  & 1,492,613 & 13 & 1.0137  &  1.0075  & \textbf{0.9366} (13-1) \\ 
  \bottomrule
    \end{tabular}}
    \label{tab:compare_reg}
    \vspace{-0.1cm}
\end{table}

The test results are shown in Table \ref{tab:compare_reg}, from which we can draw two highlights: First, overall, the performance of the \textit{VSR} is inferior to \textit{SR}. While \textit{VSR} slightly outcompetes \textit{SR} on the dataset of song popularity, \textit{SR} leads \textit{VSR} by a large margin on the rest of datasets. This is because the search space of \textit{VSR} is ablated. \textit{SR} is more likely to find a better formula. Second, \textit{TN} outperforms \textit{SR} on five datasets, which suggests the power of connectionism and the efficacy of \textit{VSR} to build a neuron. When a basic pattern regarding data is captured by a single neuron, the corresponding network can leverage these basic patterns to form an effective representation. 

It is true that VSR learns suboptimal formulas on low-dimensional datasets. Our investigation reveals that VSR is not merely a parallelizable approximation of SR; it can, in fact, surpass SR in fitting high-dimensional data.

To evaluate the pure formula-learning capabilities of VSR and conventional SR under controlled conditions, we tasked both methods with fitting synthetic data generated from a Gaussian Mixture Model (GMM)---a well-established model for real-world data distributions. We systematically varied the feature dimensionality from 3 to 400 to expose the scalability limits of traditional SR and to highlight the potential advantage of our vectorized formulation in high-dimensional regimes.

Each dataset is produced by sampling from $K=2$ Gaussian components in a $D$-dimensional space, with means drawn from $\mathcal{N}(0,5^2I)$ and diagonal covariances ensuring positive definiteness. 
The generated features $X \in \mathbb{R}^{N \times D}$ are then mapped to regression targets through an analytically defined nonlinear function:
\[
y = f(X) + \varepsilon, \quad 
f(X) = b_0 + b_1 \sum_j x_j + b_2 \sum_j x_j^2 + b_3 \sum_j x_j^3, 
\quad
\]
where $\varepsilon \sim \mathcal{N}\!\big(0, (\alpha \cdot \mathrm{std}(f(X)))^2\big)$, the coefficients $(b_0,b_1,b_2,b_3)=(0,1.0,0.3,0.05)$ and $\alpha=0.1$ control the signal-to-noise ratio.  
For each dimensionality $D\in\{3,10,30,50,80,110,150,200,250,300,400\}$, we generate 5,000 training and 1,000 validation samples, ensuring identical statistical properties across all datasets.

\begin{table}[!htbp]
\caption{Validation RMSE comparison between SR (gplearn) and VSR across dimensions.}
\label{tab:gmm-vsr-results}

\centering
\begin{tabular}{ccc}
\toprule
Dimension (D) & SR RMSE & VSR RMSE \\
\midrule
3   & \textbf{2.314} & 5.950 \\
10  & \textbf{9.736} & 14.442 \\
30  & \textbf{45.314} & 125.344 \\
50  & \textbf{29.940} & 30.146 \\
80  & \textbf{62.294} & 177.364 \\
110 & 314.336 & \textbf{72.758} \\
150 & 560.259 & \textbf{418.872} \\
200 & 798.784 & \textbf{310.205} \\
250 & 1482.034 & \textbf{419.371} \\
300 & 1911.685 & \textbf{372.673} \\
400 & 2775.011 & \textbf{1273.311} \\
\bottomrule
\end{tabular}
\end{table}

Both models—classical SR (using \textit{gplearn}) and our proposed VSR—are trained under identical evolutionary settings: population size 2,000, generation 20, crossover rate 0.3, and mutation rate 0.6. These hyperparameters are selected to balance convergence stability and computational feasibility, and they are the same across all tests to eliminate bias from search heuristics.  
The evaluation metric is the root mean squared error (RMSE) computed on the validation set. All experiments are repeated with a fixed random seed for reproducibility.  
We ensure consistent data loading, normalization (using z-score statistics from the training set), and performance reporting, so that the only difference between SR and VSR lies in their internal expression search mechanisms.

\begin{figure}[h]
\centering
\includegraphics[width=\linewidth]{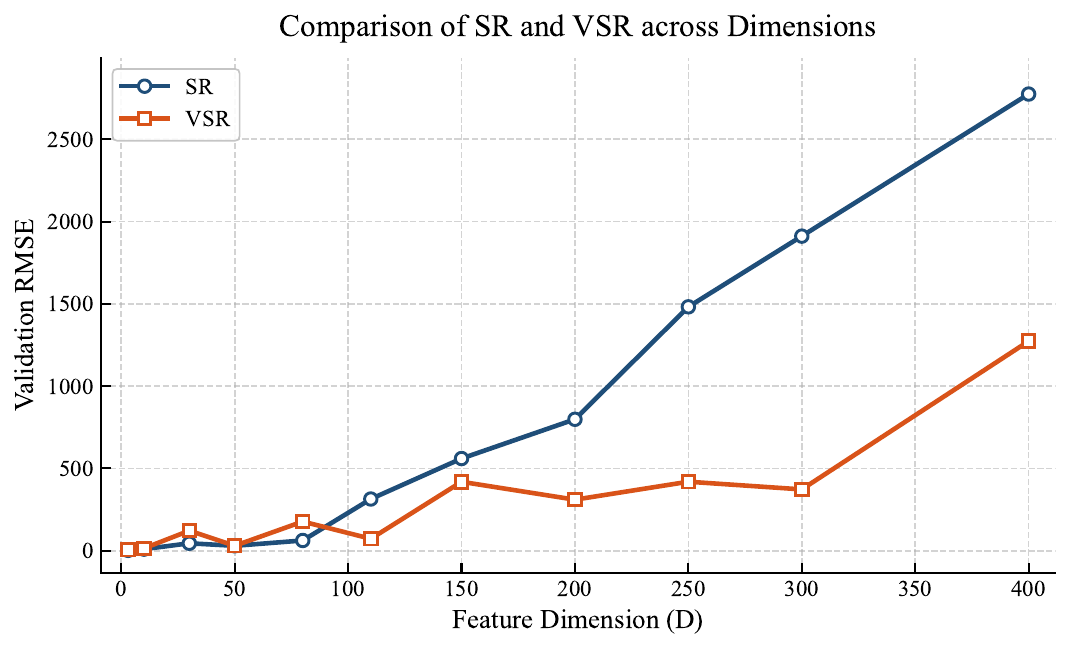}
\caption{Validation RMSE versus feature dimension. VSR exhibits superior scalability compared to traditional SR in high-dimensional regimes.}
\label{fig:gmm-vsr-rmse}
\end{figure}

As shown in Table~\ref{tab:gmm-vsr-results} and Figure~\ref{fig:gmm-vsr-rmse}, the performance of SR and VSR diverges with increasing dimensionality. While SR excels in low-dimensional space ($D \leq 80$) due to its unconstrained search, its RMSE escalates rapidly, revealing poor scalability. VSR, in contrast, shows a more robust performance profile, outperforming SR from $D \geq 80$ onward. The advantage is substantial at high dimensions; for example, at $D=400$, VSR's RMSE is less than half that of SR. This demonstrates that VSR's vectorized formulation, which shares representations across dimensions, effectively counters the curse of dimensionality. The rising errors for both methods at $D \geq 150$ underscore the intrinsic difficulty of the task, yet VSR's consistent superiority confirms its enhanced scalability and representational efficiency. Since real-world tasks are mostly high-dimensional, we think the proposal VSR is effective and useful.

\vspace{-0.36cm}
\subsection{Superiority of Task-based Neurons over Neurons Using Random Polynomials}

To further illustrate the necessity and effectiveness of using symbolic regression to generate neurons for different tasks, we perform experiments on 20 public datasets: 10 for classification and 10 for regression. MSE and classification accuracy are evaluation metrics for regression and classification, respectively. Datasets are collected from the scikit-learn package and the official website of OpenML.
We first normalize the original data to $[-1,1]$. Then we perform the vectorized symbolic regression on the normalized dataset.

The relevant information regarding these 20 datasets, and the regression results are shown in Table \ref{tab:symbolic_res}.
After learning formulas from different datasets, we use them to build a neuron and connect neurons into a network to conduct the training and test. We randomly generate a polynomial with the highest number of expressions as the highest number of expressions obtained by symbolic regression, as shown in Table \ref{tab:random_neurons}, and then use the randomly generated expressions to build a neuron and the associated network and compare it with the task-based network. In the expression construction stage, all coefficients and constants in randomly generated neurons are temporarily set to 1 as placeholders. These symbolic values are replaced with randomly initialized trainable parameters when the neurons are embedded into the network, ensuring consistent parameterization and learning across all neuron types. 

It should be noted that 1) on the dataset of \textit{electricity}, \textit{Airlines DepDelay}, and \textit{Oranges-vs.-Grapefruit}, because the expression obtained by symbolic regression is of degree 2, we can only set the randomly generated polynomial to a polynomial with only one term and degree 2 (which is $x^2$) to make it different from the expression obtained by the symbolic regression. 2) When using neurons to build the network, since the weight parameters and bias will be randomly initialized again, we set the coefficients and constant terms (if any) of the randomly generated neuron expression to 1. In fact, the coefficients and constants of the neuronal expressions do not affect the performance of the network, because they are randomly initialized.

To ensure a fair comparison, we constrain the total number of parameters in each task-based network (TN) to be fewer than its counterpart using randomly generated polynomials (RP). This design ensures that any performance gain from TN is not due to increased model capacity. In some cases, the RP networks are constructed with more hidden layers or neurons to exceed TN in parameter count, due to the difficulty in exactly aligning parameter numbers across different neuronal structures. This setting highlights the effectiveness and efficiency of task-based neurons under comparable or more constrained model sizes.

It can be seen that the fitting ability of the network built by randomly generated neurons is weaker than that of the neural network built by neurons obtained by symbolic regression. On both regression and classification, task-based networks scale better and use fewer parameters. These results indicate that symbolic regression plays an important role in identifying the appropriate polynomials.

\begin{table*}[htb!]
\caption{The formulas learned by the vectorized symbolic regression over 20 public data.}
\vspace{-0.3cm}
    \centering
   \scalebox{0.8}{ \begin{tabular}{lllll}
        \toprule  
Datasets & Instances & Features & Classes & Predicted Function  \\
        \hline 
       
 california housing & 20640 & 8 & continuous & $\bm{0.068}(\x\odot^3\x)^\top+\bm{0.15}\x^\top+0.76$  \\
 house sales & 21613 & 15 & continuous & $-\bm{0.062}(\x\odot^4\x)^\top+\bm{0.025}(\x\odot^3\x)^\top-\bm{0.010}(\x\odot\x)^\top+ \bm{0.067}\x^\top+0.74$   \\
 airfoil self noise & 1503 & 5 & continuous & $\bm{0.064}(\x\odot\x)^\top-\bm{0.038}\x^\top-0.087$   \\
  wine quality & 6497 & 11 & continuous & $\bm{0.0076}(\x\odot^4\x)^\top+\bm{0.055}(\x\odot^3\x)^\top+\bm{0.10}(\x\odot\x)^\top+\bm{0.055}\x^\top-0.00034$   \\
  fifa & 18063 & 5 & continuous & $\bm{0.30}(\x\odot^5\x)^\top-\bm{0.63}(\x\odot^4\x)^\top-\bm{0.10}(\x\odot^3\x)^\top+\bm{0.38}(\x\odot\x)^\top+\bm{0.13}\x^\top+0.010$   \\
  diamonds & 53940 & 9 & continuous & $-\bm{0.075}(\x\odot^7\x)^\top+\bm{0.16}(\x\odot^6\x)^\top+\bm{0.10}(\x\odot^5\x)^\top-\bm{0.27}(\x\odot^4\x)^\top+\bm{0.090}(\x\odot^3\x)^\top$   \\
  abalone & 4177 & 8 & continuous & $-\bm{0.088}(\x\odot^3\x)^\top-\bm{0.12}(\x\odot\x)^\top+\bm{0.046}\x^\top$   \\
  Bike Sharing Demand & 17379 & 12 & continuous & $-\bm{0.081}(\x\odot\x)^\top+\bm{0.054}\x^\top$   \\
  space ga & 3107 & 6 & continuous & $\bm{0.052}(\x\odot^4\x)^\top+\bm{0.12}(\x\odot^3\x)^\top+\bm{0.025}(\x\odot\x)^\top-\bm{0.073}\x^\top+0.54$   \\
  Airlines DepDelay & 8000 & 5 & continuous & $\bm{0.010}(\x\odot\x)^\top+\bm{0.042}\x^\top-0.27$  \\ \hline \hline
  credit & 16714 & 10 & 2 & $-\bm{0.43}(\x\odot^4\x)^\top+\bm{0.37}(\x\odot\x)^\top+0.21$  \\
  heloc & 10000 & 22 & 2 & $\bm{0.031}(\x\odot^6\x)^\top-\bm{0.026}(\x\odot^5\x)^\top+\bm{0.055}(\x\odot^3\x)^\top$   \\
  electricity & 38474 & 8 & 2 & $-\bm{0.21}(\x\odot\x)^\top+\bm{0.21}\x^\top+1.18$   \\
  phoneme & 3172 & 5 & 2 & $\bm{1.36}(\x\odot^5\x)^\top-\bm{2.91}(\x\odot^3\x)^\top+\bm{0.60}(\x\odot\x)^\top+\bm{1.22}\x^\top$   \\
  bank-marketing & 10578 & 7 & 2 & $-\bm{1.04}(\x\odot^4\x)^\top-\bm{0.14}(\x\odot^3\x)^\top+\bm{0.81}(\x\odot\x)^\top+\bm{0.043}\x^\top-0.068$  \\
 MagicTelescope & 13376 & 10 & 2 & $-\bm{0.30}(\x\odot^4\x)^\top+\bm{1.13}(\x\odot^3\x)^\top+\bm{0.46}(\x\odot\x)^\top-\bm{0.060}\x^\top+1.02$   \\
 vehicle & 846 & 18 & 4 & $-\bm{0.074}(\x\odot^4\x)^\top+\bm{0.068}(\x\odot^3\x)^\top+\bm{0.072}(\x\odot\x)^\top+\bm{0.0015}\x^\top$  \\
 Oranges-vs.-Grapefruit & 10000 & 5 & 2 & $-\bm{0.52}(\x\odot\x)^\top+\bm{0.70}\x^\top+0.89$   \\
 eye movements & 7608 & 20 & 2 & $-\bm{0.017}(\x\odot^3\x)^\top-\bm{0.011}(\x\odot\x)^\top$   \\
contaminant & 2400 & 30 & 2 & $-\bm{0.75}(\x\odot^5\x)^\top-\bm{0.67}(\x\odot^4\x)^\top+\bm{0.38}(\x\odot^3\x)^\top+\bm{0.24}(\x\odot\x)^\top+\bm{0.13}\x^\top$   \\
        \bottomrule 
    \end{tabular}   } 
    \label{tab:symbolic_res}
    \vspace{-0.1cm}
\end{table*}

\begin{table*}[ht]
\caption{Comparison of the network built by neurons using randomly generated polynomials and expressions generated by symbolic regression. The number $h_1$-$h_2$-$\cdots$-$h_k$-$y(z)$ means that this network has $k$ hidden layers, each with $h_k$ neurons, and $z$ is the number of parameters used in such a network. For example, 5-3-1 (145) means this network has two hidden layers with 5 and 3 neurons respectively in each layer, and 145 parameters. RP denotes random polynomials.
}
\vspace{-0.3cm}
    \centering
   \scalebox{0.8}{ \begin{tabular}{l|l|ll|ll}
        \toprule  
\multirow{2}{*}{Datasets} & \multirow{2}{*}{Random Polynomial (RP)} & \multicolumn{2}{c}{Structure} & \multicolumn{2}{c}{Test results}  \\
\cline{3-6}
       ~ & ~ & \makecell[c]{TN} & \makecell[c]{RP} & \makecell[c]{TN} & \makecell[c]{RP} \\  \hline

california housing & $(\x\odot^3\x)^\top + (\x\odot\x)^\top  + 1 $ & 6-3-1 (148) &	6-3-1 (148)&	\textbf{0.0720 (0.0024)} & 	0.0770 (0.0051) 
\\ 
house sales & $(\x\odot^4\x)^\top + (\x\odot\x)^\top  + \x $ & 6-4-1 (483) &	8-5-1 (509) & \textbf{0.0079 (0.0008)} &	0.0133 (0.0150) \\ 
airfoil self noise & $(\x\odot\x)^\top$ & 4-1 (53) &	8-1 (57) & \textbf{0.0438 (0.0065)} & 0.0840 (0.0102)  \\ 
wine quality & $(\x\odot^4\x)^\top + (\x\odot\x)^\top $ & 6-1 (295) &	9-5-1 (313) & \textbf{0.0545 (0.0026)} & 0.0567 (0.0034)  \\ 
fifa  & $(\x\odot^5\x)^\top + (\x\odot\x)^\top  + \x $ & 3-1 (94) &	5-1 (96) &	\textbf{0.0611 (0.0032)} & 0.0616 (0.0048) 
 \\ 
diamonds &  $(\x\odot^7\x)^\top + (\x\odot^3\x)^\top + (\x\odot\x)^\top $ & 4-1 (205) &	7-1 (218) &	\textbf{0.0109 (0.0036)} & 0.0123 (0.0052) \\ 
abalone  & $(\x\odot^3\x)^\top + \x  + 1 $ & 5-1 (141) &	8-1 (153) & \textbf{0.0239 (0.0024)} & 0.0240 (0.0030) 
 \\ 
Bike Sharing Demand  & $(\x\odot\x)^\top$ & 8-1 (217)&	10-8-1 (227)&	\textbf{0.0184 (0.0026)} & 0.0747 (0.0025) \\ 
space ga  & $(\x\odot^4\x)^\top + (\x\odot^3\x)^\top + (\x\odot\x)^\top $ 
 & 2-1 (59) & 3-1 (67) &	\textbf{0.0057 (0.0029)} & 0.0078 (0.0041) \\ 
Airlines DepDelay & $(\x\odot\x)^\top$ & 4-1 (53)&	8-1 (57)&	\textbf{0.1645 (0.0055)} & 0.1664 (0.0058) 
 \\ \hline \hline
credit & $(\x\odot^4\x)^\top + (\x\odot^3\x)^\top + \x + 1 $ & 6-2 (152) & 6-2 (224) & \textbf{0.7441 (0.0092)} &	0.7427 (0.0118)  \\ 
heloc  & $(\x\odot^6\x)^\top + (\x\odot\x)^\top + \x $ & 18-2 (1316)& 18-2 (1316)& \textbf{0.7077 (0.0145)}& 0.6944 (0.0157)  \\ 
electricity  & $(\x\odot\x)^\top$ & 5-2 (107) & 10-2 (112)& \textbf{0.7862 (0.0075)} & 0.7647 (0.0083) \\
phoneme  & $(\x\odot^5\x)^\top + (\x\odot^4\x)^\top + (\x\odot^3\x)^\top +\x + 1$ & 5-3 (89)& 5-3 (89) &	\textbf{0.8242 (0.0207)} & 0.8198 (0.0252) \\ 
bank-marketing  & $(\x\odot^4\x)^\top + (\x\odot\x)^\top + \x + 1 $  & 5-2 (187) & 7-2 (198) & \textbf{0.7938 (0.0099)} & 0.7919 (0.0099) \\ 
MagicTelescope  & $(\x\odot^4\x)^\top + (\x\odot\x)^\top + \x + 1 $ & 6-2 (296)& 8-2 (298)& \textbf{0.8449 (0.0092)} & 0.8417 (0.0092) \\ 
vehicle & $(\x\odot^4\x)^\top + 1 $ & 4-4 (360)& 13-7-4 (377)& \textbf{0.8176 (0.0362)} & 0.6894 (0.0401)  \\ 
Oranges-vs.-Grapefruit  & $(\x\odot\x)^\top$ & 3-2 (47) & 6-2 (50)& \textbf{0.9305 (0.0037)} & 0.8952 (0.0160) \\ 
eye movements  &$(\x\odot^3\x)^\top + 1$ & 10-2 (452) & 15-8-2 (461) & \textbf{0.5849 (0.0125)} & 0.5823 (0.0146) \\ 
Contaminant & $(\x\odot^5\x)^\top + (\x\odot^3\x)^\top + \x $ & 10-2 (1612) & 15-6-2 (1679) &	\textbf{0.9208 (0.0192)} & 0.9142 (0.0203)  \\  
        \bottomrule 
    \end{tabular}   } 
    \label{tab:random_neurons}
    \vspace{-0.5cm}
\end{table*}

\vspace{-0.4cm}
\subsection{Superiority of Task-based Neurons over Linear Neurons}

\label{Public_Data}

Here, we test the superiority of task-based neurons relative to linear ones. We use the same 20 datasets in the last subsection: 10 for regression and 10 for classification. We don't need to repeat the process of the vectorized symbolic regression. Instead, we directly use polynomials learned in Table \ref{tab:symbolic_res}. The training and test sets are divided according to the ratio of $8:2$. For \textit{TN} and \textit{LN}, the data division and the batch size are the same. We select 5 different network structures for each dataset for a comprehensive comparison. When designing the network structures of \textit{TN}, we ensure that the number of parameters of \textit{TN} is fewer than the \textit{LN} to show the superiority of task-based neurons in efficiency. The specific network structure and corresponding number of parameters are shown in Table \ref{tab:parameters}. Each dataset is tested 10 times for reliability of results. The MSE and classification accuracy are presented in the form of $\mathrm{mean}~(\mathrm{std})$ in Table \ref{tab:public data}.

\begin{table*}[htb]
\renewcommand\arraystretch{1.5}
\caption{Test results of linear networks and task-based networks of different structures on the public data. }
\vspace{-0.2cm}
    \centering
    \scalebox{0.7}{\begin{tabular}{l|cc|cc|cc|cc|cc}   
    \toprule
   Datasets & LN(S1) & TN(S1) & LN(S2) & TN(S2) & LN(S3) & TN(S3) & LN(S4) & TN(S4) & LN(S5) & TN(S5)\\
    \hline
     
california housing & 0.0760 (0.0057) & \textbf{0.0702 (0.0019)} & 0.0844 (0.0486)  & \textbf{0.0685 (0.0047)} & 0.0988 (0.0664) & \textbf{0.0667 (0.0057)} & 0.0861 (0.0452) & \textbf{0.0670 (0.0044)} & 0.1013 (0.0614) & \textbf{0.0540 (0.0036)} \\ 
        house sales & 0.0113 (0.0046) & \textbf{0.0109 (0.0041)} & 0.0111 (0.0037) & \textbf{0.0100 (0.0027)} & 0.0139 (0.0130) & \textbf{0.0098 (0.0017)} & 0.0139 (0.0105) & \textbf{0.0098 (0.0017)} & 0.0158 (0.0128) & \textbf{0.0112 (0.0047)} \\ 
        airfoil self noise & 0.0428 (0.0181) & \textbf{0.0402 (0.0050)} & 0.0329 (0.0286) & \textbf{0.0270 (0.0042)} & 0.0669 (0.0547) & \textbf{0.0287 (0.0091)} & 0.0250 (0.0070) & \textbf{0.0233 (0.0049)} & 0.0327 (0.0358) & \textbf{0.0134 (0.0024)} \\ 
        wine quality & 0.0636 (0.0084) & \textbf{0.0625 (0.0083)} & 0.0619 (0.0041) & \textbf{0.0602 (0.0021)} & 0.0634 (0.0098) & \textbf{0.0592 (0.0051)} & 0.0613 (0.0076) & \textbf{0.0592 (0.0051)} & 0.0604 (0.0094) & \textbf{0.0597 (0.0066)} \\ 
        fifa & 0.0895 (0.0283) & \textbf{0.0637 (0.0055)} & 0.0728 (0.0085) & \textbf{0.0608 (0.0023)} & 0.1138 (0.0477) & \textbf{0.0622 (0.0032)} & 0.1093 (0.0425) & \textbf{0.0634 (0.0051)} & 0.0946 (0.0420) & \textbf{0.0597 (0.0018)} \\ 
        diamonds & 0.0128 (0.0050) & \textbf{0.0097 (0.0023)} & 0.0108 (0.0066) & \textbf{0.0078 (0.0020)} & 0.0084 (0.0032) & \textbf{0.0081 (0.0027)} & 0.0103 (0.0073) & \textbf{0.0080 (0.0018)} & 0.0145 (0.0169) & \textbf{0.0084 (0.0024)} \\ 
        abalone & 0.0253 (0.0026) & \textbf{0.0250 (0.0025)} & 0.0289 (0.0109) & \textbf{0.0250 (0.0029)} & 0.0296 (0.0087) & \textbf{0.0269 (0.0074)} & 0.0282 (0.0098) & \textbf{0.0258 (0.0032)} & 0.0300 (0.0102) & \textbf{0.0268 (0.0037)} \\
        Bike Sharing Demand & 0.0144 (0.0040) & \textbf{0.0132 (0.0022)} & 0.0133 (0.0047) & \textbf{0.0110 (0.0022)} & 0.0103 (0.0015) & \textbf{0.0092 (0.0010)} & 0.0137 (0.0043) & \textbf{0.0101 (0.0021)} & 0.0104 (0.0018) & \textbf{0.0083 (0.0011)} \\ 
        space ga & 0.0077 (0.0021) & \textbf{0.0069 (0.0015)} & 0.0063 (0.0020) & \textbf{0.0054 (0.0007)} & 0.0120 (0.0050) & \textbf{0.0054 (0.0015)} & 0.0108 (0.0053) & \textbf{0.0061 (0.0036)} & 0.0103 (0.0046) & \textbf{0.0049 (0.0006)} \\ 
        Airlines DepDelay & 0.1631 (0.0055) & \textbf{0.1617 (0.0049)} & 0.1643 (0.0057) & \textbf{0.1616 (0.0046)} & 0.1662 (0.0062) & \textbf{0.1649 (0.0051)} & 0.1636 (0.0060) & \textbf{0.1634 (0.0047)} & 0.1651 (0.0063) & \textbf{0.1635 (0.0059)} \\ \hline \hline

credit & 0.7123 (0.0398) & \textbf{0.7433 (0.0077)} & 0.7302 (0.0108) & \textbf{0.7392 (0.0087)} & 0.7358 (0.0089) & \textbf{0.7447 (0.0055)} & 0.7267 (0.0260) & \textbf{0.7447 (0.0055)} & 0.7108 (0.0316) & \textbf{0.7372 (0.0083)} \\ 
        heloc & 0.7030 (0.0091) & \textbf{0.7080 (0.0077)} & 0.6980 (0.0112) & \textbf{0.7040 (0.0063)} & 0.6950 (0.0145) & \textbf{0.6950 (0.0098)} & 0.6890 (0.0073) & \textbf{0.6900 (0.0073)} & 0.6910 (0.0129) & \textbf{0.6930 (0.0100)} \\ 
        electricity & 0.7770 (0.0062) & \textbf{0.7860 (0.0045)} & 0.7880 (0.0068) & \textbf{0.7950 (0.0046)} & 0.7910 (0.0046) & \textbf{0.8000 (0.0047)} & 0.7900 (0.0055) & \textbf{0.7990 (0.0051)} & 0.7630 (0.0882) & \textbf{0.8100 (0.0035)} \\
        phoneme & 0.8170 (0.0321) & \textbf{0.8420 (0.0099)} & 0.8020 (0.0994) & \textbf{0.8510 (0.0116)} & 0.7790 (0.1460) & \textbf{0.8560 (0.0107)} & 0.8470 (0.0131) & \textbf{0.8550 (0.0079)} & 0.8120 (0.1080) & \textbf{0.8540 (0.0090)} \\ 
        bank-marketing & 0.7780 (0.0140) & \textbf{0.7840 (0.0078)} & 0.7870 (0.0102) & \textbf{0.7940 (0.0067)} & 0.7600 (0.0839) & \textbf{0.7930 (0.0066)} & 0.7560 (0.0893) & \textbf{0.7920 (0.0073)} & 0.7840 (0.0086) & \textbf{0.7920 (0.0083)} \\ 
        MagicTelescope & 0.8452 (0.0098) & \textbf{0.8531 (0.0071)} & 0.8430 (0.0078) & \textbf{0.8557 (0.0062)} & 0.8456 (0.0047) & \textbf{0.8580 (0.0040)} & 0.8462 (0.0073) & \textbf{0.8573 (0.0063)} & 0.8433 (0.0090) & \textbf{0.8580 (0.0066)} \\ 
         vehicle & 0.8090 (0.0335) & \textbf{0.8180 (0.0277)} & 0.8090 (0.0258) & \textbf{0.8240 (0.0237)} & 0.7980 (0.0312) & \textbf{0.8150 (0.0241)} & 0.8100 (0.0377) & \textbf{0.8110 (0.0264)} & 0.8090 (0.0349) & \textbf{0.8140 (0.0214)} \\ 
        Oranges-vs.-Grapefruit & 0.9288 (0.1381) & \textbf{0.9429 (0.0049)} & 0.9430 (0.1489) & \textbf{0.9740 (0.0020)} & 0.9429 (0.1490) & \textbf{0.9751 (0.0020)} & 0.9426 (0.1457) & \textbf{0.9751 (0.0020)} & 0.8927 (0.1944) & \textbf{0.9758 (0.0076)} \\ 
        eye movements & 0.5790 (0.0117) & \textbf{0.5840 (0.0116)} & 0.5800 (0.0188) & \textbf{0.5910 (0.0107)} & 0.5720 (0.0267) & \textbf{0.5880 (0.0179)} & 0.5680 (0.0293) & \textbf{0.5840 (0.0103)} & 0.5790 (0.0288) & \textbf{0.5800 (0.0113)} \\ 
        Contaminant & 0.9220 (0.0108) & \textbf{0.9300 (0.0095)} & 0.9290 (0.0134) & \textbf{0.9300 (0.0120)} & 0.9250 (0.0092) & \textbf{0.9310 (0.0083)} & 0.9260 (0.0126) & \textbf{0.9300 (0.0109)} & 0.9010 (0.0638) & \textbf{0.9340 (0.0096)} \\ 
                     
          \bottomrule
    \end{tabular}}
    \label{tab:public data}
    \vspace{-0.5cm}
\end{table*}

\textbf{Regression.} The designs of \textit{LN} and \textit{TN} all adopt a fully connected network. The activation function is \textit{ReLU} for \textit{LN} and \textit{Sigmoid} for \textit{TN}.  In Section 1.7 of SMs, we test the suitability of four different activation functions: ReLU, Leaky-ReLU, Sigmoid, and Tanh. We show that ReLU generally provides better or comparable performance for LN, while Sigmoid tends to yield more accurate results for TN in most cases. We think this is because LN utilizes piecewise-linear mappings where ReLU preserves linearity and gradient stability, while TN incorporates higher-order polynomial expressions that require bounded activations like Sigmoid to ensure numerical stability by preventing activation explosion.  Both networks use \textit{MSELoss} as the loss function and \textit{RMSProp} as the optimizer. The details of the specific network structures and the corresponding number of parameters are shown in Table \ref{tab:parameters} of SMs.  

In Table \ref{tab:public data}, in every dataset, the mean and standard deviation of fitting errors (MSE) of \textit{LN} are larger than those of \textit{TN}, indicating that \textit{TN} has a stronger fitting ability and better generalization than \textit{LN}. In some datasets such as \textit{fifa} and \textit{airfoil self noise}, \textit{TN} leads \textit{LN} by a large margin.

% \begin{figure}[htb] 
% \centering
% \includegraphics[width=\linewidth]{Figure/fifa_train.png}
% \caption{Fifa dataset training error.} 
% \label{fig:fifa_train}
% \end{figure}

% \begin{figure}[htb] 
% \centering
% \includegraphics[width=\linewidth]{Figure/fifa_test.png}
% \caption{Fifa dataset test error.} 
% \label{fig:fifa_test}
% \end{figure}

\textbf{Classification}. For this task, the network design uses a fully connected network both for \textit{LN} and \textit{TN}. The activation function uses \textit{ReLU} for \textit{LN} and \textit{Sigmoid} for \textit{TN}. Both networks use \textit{CrossEntropyLoss} as the loss function and \textit{Adam} as the optimizer. The network is trained and tested in the same way as the regression tasks.

The classification results are shown in Table \ref{tab:public data}. It is observed that \textit{TN} has higher training accuracy and test accuracy, indicating that \textit{TN} has stronger fitting ability. For every dataset, the accuracy of \textit{TN} over test sets is higher than \textit{LN}, and the standard deviation is smaller than \textit{LN}. For datasets like \textit{phoneme}, \textit{electricity}, and \textit{Orange-vs-Grapefruit}, the improvement by \textit{TN} is at least 3\%, which is significant. Moreover, \textit{TN} achieves better classification accuracy with fewer parameters.

% \begin{sidewaystable*}[htb]
\begin{table*}[htb]
\caption{Comparing the linear and task-based neurons in five different network structures for public data. The number $h_1$-$h_2$-$\cdots$-$h_k$-$y(z)$ means that this network has $k$ hidden layers, each with $h_k$ neurons, and $z$ is the number of parameters used in such a network.}
\vspace{-0.2cm}
    \centering
\renewcommand\arraystretch{1.2}
\resizebox{1.0\linewidth}{!}{   
\scalebox{0.8}{\begin{tabular}{l|ll|ll|ll|ll|ll}

    \hline
    \multirow{2}{*}{Datasets} & \multicolumn{2}{c}{S1} & \multicolumn{2}{c}{S2} & \multicolumn{2}{c}{S3} & \multicolumn{2}{c}{S4}& \multicolumn{2}{c}{S5} \\ 
\cline{2-11}

 & \makecell[c]{LN} & \makecell[c]{TN}  & \makecell[c]{LN} & \makecell[c]{TN}  & \makecell[c]{LN} & \makecell[c]{TN}  & \makecell[c]{LN} & \makecell[c]{TN} & \makecell[c]{LN} & \makecell[c]{TN}  \\ \hline

california housing & 7-7-1(127) & 5-1(96) & 12-12-8-1(377) & 8-1(153) & 18-14-10-6-1(651) & 23-1(438) & 14-12-10-8-1(533) & 14-1(267) & 12-16-24-16-8-1(1269) & 20-6-1(599) \\ 
        house sales & 15-15-1(496) & 7-1(456) & 22-22-15-1(1219) & 18-1(1171) & 32-25-22-15-1(2270) & 28-1(1821) & 30-25-20-15-1(2106) & 28-1(1821) & 23-30-45-30-15-1(4344) & 30-13-1(3456) \\ 
        airfoil self noise & 5-5-1(66) & 2-1(27) & 8-8-5-1(171) & 5-1(66) & 10-8-8-5-1(271) & 9-1(118) & 9-8-6-5-1(229) & 12-1(157) & 8-10-15-10-5-1(524) & 12-4-1(241) \\ 
        wine quality & 11-11-1(276) & 5-1(246) & 16-16-11-1(663) & 11-1(540) & 22-16-16-11-1(1103) & 19-1(932) & 19-16-14-11-1(963) & 19-1(932) & 16-22-33-22-11-1(2338) & 28-9-1(2314) \\ 
        fifa & 5-5-1(66) & 2-1(63) & 8-8-5-1(171) & 5-1(156) & 10-8-8-5-1-(271) & 8-1(249) & 9-8-6-5-1(229) & 7-1(218) & 8-10-15-10-5-1(524) & 10-4-1(485) \\ 
        diamonds & 8-8-1(161) & 3-1(154) & 14-14-9-1(495) & 9-1(460) & 18-14-14-10-1(817) & 15-1(766) & 16-14-11-9-1(681) & 12-1(613) & 15-18-25-15-10-1(1474) & 17-6-1(1329) \\ 
        abalone & 8-6-1(133) & 4-1(113) & 12-12-8-1(377) & 7-1(197) & 18-14-12-8-1(721) & 16-1(449) & 14-12-10-8-1(533) & 14-1(393) & 18-22-24-16-8-1(1677) & 20-15-1(1461) \\ 
        Bike Sharing Demand & 12-12-1(325) & 6-1(163) & 18-18-12-1(817) & 12-1(325) & 24-20-18-12-1(1431) & 30-1(811) & 24-18-15-12-1(1252) & 26-1(703) & 18-24-36-24-12-1(2791) & 30-10-1(1381) \\ 
        space ga & 3-1(25) & 1(25) & 6-3-1(67) & 2-1(59) & 6-6-3-1(109) & 3-1(88) & 6-12-6-3-1(229) & 6-1(175) & 12-18-12-6-3-1(649) & 6-12-1(499) \\ 
        Airlines DepDelay & 5-5-1(66) & 2-1(27) & 8-6-3-1(127) & 6-1(79) & 15-10-7-3-1(355) & 15-1(196) & 9-8-6-5-1(229) & 10-1(131) & 8-10-15-10-5-1(524) & 12-4-1(241) \\ \hline \hline
        credit & 10-10-2(242) & 5-2(127) & 15-15-10-2(587) & 10-2(252) & 20-15-15-10-2(957) & 18-2(452) & 18-15-12-10-2(827) & 18-2(452) & 15-20-30-20-10-2(1967) & 25-8-2(967) \\ 
        heloc & 22-22-2(1058) & 14-2(1024) & 33-33-22-2(2675) & 35-2(2557) & 40-30-30-20-2(3742) & 40-2(1864) & 38-33-28-22-2(3797) & 44-2(3214) & 30-40-60-40-20-2(7692) & 55-18-2(6783) \\ 
        electricity & 8-8-2(162) & 4-2(86) & 12-12-8-2(386) & 8-2(170) & 16-12-12-8-2(626) & 14-2(296) & 14-12-10-8-2(542) & 14-2(296) & 12-16-24-16-8-2(1278) & 20-6-8(690) \\
        phoneme & 5-5-2(72) & 2-2(60) & 8-8-5-2(177) & 5-2(147) & 10-8-8-5-2(277) & 9-2(263) & 9-8-6-5-2(235) & 8-2(234) & 8-10-15-10-5-2(530) & 12-4-2(482) \\ 
        bank-marketing & 7-7-2(128) & 2(58) & 10-10-7-2(283) & 7-2(261) & 14-10-10-7-2(465) & 12-2(446) & 12-10-9-7-2(411) & 10-2(372) & 10-15-20-15-6-2(990) & 15-6-2(851) \\ 
        MagicTelescope & 12-13-2(329) & 5-2(247) & 12-10-10-2(394) & 7-2(345) & 20-15-15-10-2(957) & 16-2(786) & 18-12-10-8-2(662) & 13-2(639) & 15-20-23-18-8-2(1570) & 15-10-2(1307) \\ 
        vehicle & 18-16-4(714) & 7-4(627) & 25-20-15-4(1374) & 10-4(894) & 36-27-27-18-4(3019) & 32-4(2852) & 32-27-22-18-4(2605) & 28-4(2496) & 27-36-54-36-18-4(6241) & 45-14-4(6047) \\ 
        Oranges-vs.-Grapefruit & 3-3-2(38) & 2(22) & 8-8-5-2(177) & 5-2(77) & 10-8-8-5-2(277) & 9-2(137) & 8-8-6-4-2(212) & 9-2(137) & 8-8-10-8-4-2(344) & 12-5-2(279) \\ 
        eyemovements & 18-18-2(758) & 15-2(677) & 25-15-10-2(1097) & 20-2(902) & 40-30-30-20-2(3662) & 35-2(1577) & 35-30-20-10-2(2667) & 40-2(1802) & 25-30-35-20-15-2(3457) & 40-20-2(3342) \\ 
        Contaminant & 30-18-2(1526) & 8-2(1290) & 45-45-30-2(4907) & 30-2(4832) & 65-45-45-30-2(8497) & 49-2(7891) & 52-45-38-30-2(6977) & 39-2(6281) & 45-60-90-60-30-2(16997) & 55-25-2(15457) \\ \hline
    
    \end{tabular}
    }
    }
    \label{tab:parameters}
    % \vspace{-0.4cm}
% \end{sidewaystable*}
\end{table*}

% Linear regression has a better fitting effect on data with linear correlation, but if the data is relatively complex and the traditional symbolic regression cannot learn the inherent law of the data, the error of fitting the data with a straight line is usually smaller. Therefore, you can see from Table \ref{tab:compare_reg} that \textit{LR} has a smaller mean square error than \textit{SR}. \textit{VSR} tends to learn the patterns in the data, and does not seek to minimize the data fitting error, so the error of fitting the data directly with \textit{VSR} is the largest. However, using the pattern learned by \textit{VSR} to build a neural network can combine the advantages of both \textit{VSR} and neural network, so it can be seen that the fitting error of \textit{TN} is minimal.

% \begin{sidewaystable*}[htb]

% \begin{figure}[htb] 
% \centering
% \includegraphics[width=\linewidth]{Figure/fifa_train.png}
% \caption{Fifa dataset training error.} 
% \label{fig:fifa_train}
% \end{figure}

% \begin{figure}[htb] 
% \centering
% \includegraphics[width=\linewidth]{Figure/fifa_test.png}
% \caption{Fifa dataset test error.} 
% \label{fig:fifa_test}
% \end{figure}

\begin{table}[!htbp]
\centering
\footnotesize % 设置字体大小为小号
\caption{Comparison of network training and inference time (seconds).}
\scalebox{0.9}{\begin{tabular}{lcccc}
\toprule
datasets  & TN\_train & LN\_train & TN\_infer & LN\_infer \\
\midrule
abalone  & 155.7479 & 91.4032 & 0.0617 & 0.0276 \\
airfoil self noise & 26.3141 & 36.6334 & 0.0111 & 0.0113 \\
Airlines DepDelay  & 64.9726 & 51.2741 & 0.0359 & 0.0224 \\
Bike Sharing Demand  & 72.1809 & 71.0346 & 0.0323 & 0.0199 \\
Oranges-vs.-Grapefruit  & 71.4266 & 56.2832 & 0.1364 & 0.0221 \\
\bottomrule
\end{tabular}}
\label{tab:time}
\end{table}

We also record the training and test time of task-based networks. The experiments are repeated five times, and the average value is reported as the final result. The network architecture used is from Table \ref{tab:parameters}. The number of epochs for training was set to 300. The experiments were conducted on a system with an AMD EPYC 7713 64-Core Processor (CPU) and an NVIDIA GeForce RTX 4090 24GB (GPU).  From Table \ref{tab:time}, it can be seen that, on average, task-based networks only increase the inference time mildly, which means that our method is not computationally prohibitive. The training time of task-based networks is also close to that of linear networks.

% \begin{figure}[htb] 
% \centering
% \includegraphics[width=\linewidth]{Figure/electricity_train.png}
% \caption{Electricity dataset training error.} 
% \label{fig:electricity_train}
% \end{figure}

% \begin{figure}[htb] 
% \centering
% \includegraphics[width=\linewidth]{Figure/electricity_test.png}
% \caption{Electricity dataset test error.} 
% \label{fig:electricity_test}
% \end{figure}

% Note that the IEEE does not put floats in the very first column
% - or typically anywhere on the first page for that matter. Also,
% in-text middle ("here") positioning is typically not used, but it
% is allowed and encouraged for Computer Society conferences (but
% not Computer Society journals). Most IEEE journals/conferences use
% top floats exclusively. 
% Note that, LaTeX2e, unlike IEEE journals/conferences, places
% footnotes above bottom floats. This can be corrected via the
% \fnbelowfloat command of the stfloats package.

\begin{table*}[htb]
\renewcommand\arraystretch{1.5}
\caption{Test results of task-based and quadratic networks on the 20 public datasets. The number $h_1$-$h_2$-$\cdots$-$h_k$-$1(z)$ means that this network has $k$ hidden layers, each with $h_k$ neurons, and $z$ is the number of parameters used in such a network.  Best results are highlighted in \textbf{bold}.}
    \centering
    \scalebox{0.7}{\begin{tabular}{c|cl|cl|cl|cl|cl}
    \toprule  
\multirow{2}{*}{Datasets} & \multicolumn{2}{c}{TN} & \multicolumn{2}{c}{Bu} & \multicolumn{2}{c}{Goyal} & \multicolumn{2}{c}{Fan} & \multicolumn{2}{c}{Xu} \\ \cline{2-11}
~ & Perf. & Structure &  Perf. & Structure &  Perf. & Structure &  Perf. & Structure &  Perf. & Structure \\ \hline

california housing & \textbf{0.0689 (0.0055)} & 5-3-1 (125) & 0.0737 (0.0061) & 5-3-1 (134) & 0.1083 (0.0128) & 7-7-1 (127) & 0.1237 (0.0677) & 3-3-1 (129) & 0.0926 (0.0423) & 3-3-1 (129) \\ 
    house sales & \textbf{0.0102 (0.0023)} & 4-1 (261) & 0.0170 (0.0093) & 8-3-1 (318) & 0.0245 (0.0039) & 11-8-1 (281) & 0.0488 (0.0317) & 5-3-1 (306) & 0.0220 (0.0164) & 5-3-1 (306) \\ 
    airfoil self noise & \textbf{0.0189 (0.0043)} & 6-4-1 (127) & 0.0204 (0.0054) & 7-4-1 (158) & 0.0479 (0.0165) & 11-8-1 (171) & 0.0492 (0.0445) & 5-3-1 (156) & 0.0382 (0.0331) & 5-3-1 (156) \\ 
    wine quality & \textbf{0.0586 (0.0045)} & 5-1 (246) & 0.0754 (0.0359) & 7-5-1 (260) & 0.0898 (0.0632) & 12-8-1 (257) & 0.0684 (0.0112) & 5-4-1 (267) & 0.0668 (0.0117) & 5-4-1 (267) \\ 
    fifa & \textbf{0.0613 (0.0021)} & 8-1 (249) & 0.0776 (0.0068) & 10-6-1 (266) & 0.0811 (0.0320) & 12-10-5-1 (263) & 0.0784 (0.0330) & 8-4-1 (267) & 0.0827 (0.0315) & 8-4-1 (267) \\ 
    diamonds &\textbf{ 0.0078 (0.0007)} & 4-1 (205) & 0.0411 (0.1029) & 9-5-1 (292) & 0.0352 (0.0493) & 12-8-5-1 (275) & 0.0457 (0.0687) & 6-4-1 (279) & 0.0292 (0.0518) & 6-4-1 (279) \\ 
    abalone & \textbf{0.0238 (0.0024)} & 5-3-1 (183) & 0.3189 (0.8751) & 6-5-1 (190) & 0.0289 (0.0028) & 12-6-1 (193) & 0.0369 (0.0163) & 5-3-1 (201) & 0.0319 (0.0121) & 5-3-1 (201) \\ 
    Bike sharing Demand & \textbf{0.0097 (0.0007)} & 7,4,1 (244) & 0.0129 (0.0026) & 7,4,1 (256) & 0.0814 (0.0093) & 12,7,1 (255) & 0.0519 (0.0436) & 5,3,1 (261) & 0.0167 (0.0033) & 5,3,1 (261) \\ 
    space ga & \textbf{0.0049 (0.0007)} & 5-1 (146) & 0.0064 (0.0026) & 6-5-1 (166) & 0.0109 (0.0035) & 10-7-1 (155) & 0.0100 (0.0050) & 5-2-1 (150) & 0.0086 (0.0051) & 5-2-1 (150) \\ 
    Airlines DepDelay & \textbf{0.1616 (0.0045)} & 6-5-1 (142) & 0.1642 (0.0047) & 6-5-1 (154) & 0.1637 (0.0052) & 10-8-1 (157) & 0.1646 (0.0061) & 5-3-1 (156) & 0.1630 (0.0056) & 5-3-1 (156) \\ \hline \hline
    credit & \textbf{0.7457 (0.0070)} & 6-4-2 (196) & 0.7165 (0.0276) & 6-5-2 (226) & 0.7370 (0.0082) & 10-8-2 (216) & 0.6943 (0.0679) & 5-3-2 (243) & 0.7235 (0.0107) & 5-3-2 (243) \\ 
    heloc & \textbf{0.7165 (0.0086)} & 5-4-2 (425) & 0.7118 (0.0098) & 8-4-2 (460) & 0.7127 (0.0123) & 14-6-2 (426) & 0.7117 (0.0106) & 5-4-2 (447) & 0.6915 (0.0680) & 5-4-2 (447) \\ 
    electricity & \textbf{0.8012 (0.0046)} & 10-5-2 (297) & 0.7865 (0.0097) & 10-5-2 (314) & 0.7798 (0.0079) & 14-8-6-2 (314) & 0.7910 (0.0078) & 7-4-2 (315) & 0.7280 (0.1134) & 7-4-2 (315) \\ 
    phoneme & \textbf{0.8561 (0.0102)} & 7-2 (205) & 0.8447 (0.0148) & 8-6-2 (232) & 0.7447 (0.0160) & 13-8-2 (208) & 0.8088 (0.1098) & 6-4-2 (222) & 0.8413 (0.0143) & 6-4-2 (222) \\ 
    bank-marketing & \textbf{0.7933 (0.0060)} & 7-2 (261) & 0.7835 (0.0071) & 10-6-2 (320) & 0.7777 (0.0068) & 12-10-6-2 (306) & 0.7895 (0.0065) & 8-4-2 (330) & 0.7867 (0.0100) & 8-4-2 (330) \\ 
    MagicTelescope & \textbf{0.8552 (0.0056)} & 8-2 (394) & 0.8474 (0.0075) & 12-5-2 (418) & 0.8105 (0.0068) & 14-10-8-2 (410) & 0.8487 (0.0077) & 8-5-2 (435) & 0.8471 (0.0079) & 8-5-2 (435) \\ 
    vehicle & \textbf{0.8312 (0.0190)} & 8-4 (716) & 0.7271 (0.0380) & 12-8-4 (736) & 0.6535 (0.0873) & 18-12-9-4 (727) & 0.7112 (0.0573) & 8-7-4 (741) & 0.7912 (0.0246) & 8-7-4 (741) \\ 
    Oranges-vs.-Grapefruit & \textbf{0.9906 (0.0035)} & 3-2 (47) & 0.9868 (0.0076) & 5-2 (84) & 0.8710 (0.1213) & 8-3-2 (83) & 0.9760 (0.0086) & 3-2 (78) & 0.9803 (0.0091) & 3-2 (78) \\ 
    eye movements & \textbf{0.5905 (0.0122)} & 10-2 (452) & 0.5823 (0.0140) & 10-2 (464) & 0.5673 (0.0079) & 15-8-2 (461) & 0.5687 (0.0270) & 20-8 (558) & 0.5803 (0.0140) & 20-8 (558) \\
    Contaminant & \textbf{0.9310 (0.0084)} & 8-2 (1290) & 0.8775 (0.0529) & 15-10-2 (1294) & 0.8435 (0.1135) & 28-12-5-2 (1293) & 0.8448 (0.1012) & 12-6-2 (1392) & 0.9229 (0.0140) & 12-6-2 (1392) \\ 
    
    \bottomrule
    \end{tabular}}
    \label{tab:quadratic}
\end{table*}

\subsection{Comparison of Linear and Task-based Networks over Four Activation Functions}

Activation functions should be chosen to match the nature of the underlying neurons. However, applying what kinds of activations for task-based neurons lacks a clear theoretical underpinning. We configure ReLU for LN and Sigmoid for TN based on fundamental architectural compatibility: LN utilizes piecewise-linear mappings where ReLU preserves linearity and gradient stability, while TN incorporates higher-order polynomial expressions derived via VSR that require bounded activations like Sigmoid to ensure numerical stability by preventing activation explosion.

\begin{table*}[!htbp]    
	\caption{Comparison of LN and TN under different activation functions across 4 regression and 4 classification datasets. The red background indicates the best result of TN, and the green background indicates the best result of LN.}
	\label{tab:activation_comparison}
	\scalebox{0.85}{
	\begin{tabular}{lcccccccc}         
		\toprule
		\multirow{2}{*}{\textbf{Datasets}} 
		& \multicolumn{2}{c}{\textbf{ReLU}} 
		& \multicolumn{2}{c}{\textbf{LeakyReLU}} 
		& \multicolumn{2}{c}{\textbf{Sigmoid}} 
		& \multicolumn{2}{c}{\textbf{Tanh}}  \\ 
		\cmidrule(lr){2-3} 
		\cmidrule(lr){4-5} 
		\cmidrule(lr){6-7} 
		\cmidrule(lr){8-9} 
		& LN & TN & LN & TN & LN & TN & LN & TN \\
		\midrule
					
		california housing & \cellcolor{green!15}0.0753 (0.0043) & 0.0733 (0.0035) & 0.0757 (0.0038) & 0.0735 (0.0034) & 0.0834 (0.0034) & \cellcolor{red!15}0.0719 (0.0032) & 0.0789 (0.0037) & 0.0735 (0.0036)  \\
		house sales & \cellcolor{green!15}0.0068 (0.0002) & 0.0071 (0.0003) & 0.0069 (0.0003) & 0.0071 (0.0005) & 0.0107 (0.0006) & \cellcolor{red!15}0.0066 (0.0003) & 0.0071 (0.0002) & 0.0069 (0.0003) \\		
		fifa & \cellcolor{green!15}0.0692 (0.0030) & 0.0648 (0.0024) & 0.0708 (0.0019) & 0.0640 (0.0019) & 0.1029 (0.0039) & \cellcolor{red!15}0.0625 (0.0017) & 0.0826 (0.0042) & 0.0634 (0.0019)  \\
        Bike Sharing Demand & \cellcolor{green!15}0.0152 (0.0037) & 0.0189 (0.0047) & 0.0163 (0.0030) & 0.0176 (0.0031) & 0.0496 (0.0029) & \cellcolor{red!15}0.0110 (0.0016) & 0.0351 (0.0093) & 0.0111 (0.0013) \\ \hline \hline

		heloc & \cellcolor{green!15}0.7244 (0.0087) & 0.7220 (0.0077) & 0.7227 (0.0085) & 0.7221 (0.0080) & 0.7198 (0.0112) & \cellcolor{red!15}0.7249 (0.0099) & 0.7230 (0.0104) & 0.7246 (0.0098) \\
		electricity & \cellcolor{green!15}0.7731 (0.0036) & 0.7713 (0.0054) & 0.7723 (0.0032) & 0.7710 (0.0055) & 0.7530 (0.0056) & \cellcolor{red!15}0.7738 (0.0036) & 0.7694 (0.0043) & 0.7729 (0.0044)  \\
		MagicTelescope & \cellcolor{green!15}0.8408 (0.0062) & 0.8487 (0.0052) & 0.8405 (0.0049) & 0.8478 (0.0047) & 0.8207 (0.0048) & \cellcolor{red!15}0.8489 (0.0053) & 0.8348 (0.0037) & 0.8482 (0.0057)  \\
		vehicle & \cellcolor{green!15}0.8494 (0.0231) & 0.8547 (0.0169) & 0.8412 (0.0186) & 0.8512 (0.0252) & 0.8088 (0.0304) & \cellcolor{red!15}0.8565 (0.0192) & 0.8359 (0.0191) & 0.8418 (0.0269)  \\	
		\bottomrule
	\end{tabular}}
\end{table*}

\begin{table}[!htbp]
    \centering
    \caption{Structural and Parametric Comparisons of LN and TN on 8 datasets.}
    \label{tab:activation_comparison_structure}

    \scalebox{1}{
    \begin{tabular}{lcc}
        \toprule
        \multirow{2}{*}{\textbf{Datasets}} 
        & \multicolumn{2}{c}{\textbf{Network structures (\# parameters)}} \\ 
        \cmidrule(lr){2-3} & LN & TN \\
        \midrule					
        california housing & 10-10-8-1 (297) & 11-1 (210) \\
        house sales & 22-20-15-1 (1143) & 15-1 (976) \\		
        fifa &  8-8-1 (129) & 4-1 (125) \\
        Bike Sharing Demand & 14-10-1 (343) & 8-6-1 (315) \\ \hline \hline

        heloc &  30-30-20-2 (2282) & 30-2 (2192)\\
        electricity & 15-2 (167) & 7-2 (149) \\
        MagicTelescope & 14-10-2 (326) & 6-2 (296) \\
        vehicle &  24-13-4 (837) & 9-4 (805) \\
        \bottomrule
    \end{tabular}}
    \label{parameters}
\end{table}

We do comprehensive comparison by exploring four common activation functions: ReLU, LeakyReLU, Sigmoid, and Tanh on linear neurons and task-based neurons, respectively, across multiple regression and classification datasets (see Table \ref{tab:activation_comparison}). The results indicate that ReLU generally provides better or comparable performance for LN, while Sigmoid tends to yield more accurate results for TN in most cases. Notably, Table \ref{parameters} further illustrates TN achieve competitive results with fewer parameters and simpler architectures, which verifies again the efficacy of task-based neurons.

Here, for task-based networks, we theoretically show that a bounded activation function (e.g., sigmoid, tanh) guarantees stability, whereas unbounded activations (e.g., ReLU) lead to uncontrollable growth.

\textbf{Network Definition}.
Without the loss of generality, let the input be a scalar \(x\). For layer \(\ell = 1, \dots, L\), define
\[
z_\ell = P_\ell(h_{\ell-1}), \qquad h_\ell = \sigma(z_\ell),
\]
with \(h_0 = x\). Here \(P_\ell\) is a polynomial of degree \(d_\ell \ge 1\) (for simplicity we may take \(P_\ell(z) = w_\ell z^{d_\ell}\) with constant weight \(w_\ell\)), and \(\sigma\) is the activation function. All coefficients of \(P_\ell\) are assumed fixed.

\textbf{Bounded Activation Prevents Explosion}.
Assume \(\sigma\) is bounded: there exists \(M > 0\) such that \(|\sigma(t)| \le M\) for all real \(t\).

\begin{itemize}
    \item \textit{Base case:} \(h_1 = \sigma(P_1(x))\) satisfies \(|h_1| \le M\).
    \item \textit{Inductive step:} Suppose \(|h_{\ell-1}| \le M\). Then \(z_\ell = P_\ell(h_{\ell-1})\) is some real number (its magnitude may be large, depending on the polynomial coefficients). Nevertheless, because \(\sigma\) is bounded, we have \(|h_\ell| = |\sigma(z_\ell)| \le M\).
\end{itemize}
Thus by induction, \(|h_\ell| \le M\) for all \(\ell = 1,\dots,L\). No matter how deep the network or how large the polynomial coefficients, the activations remain within \([-M,M]\). The numerical overflow is impossible.

\textbf{Unbounded Activation Leads to Explosion}.
Now suppose \(\sigma\) is unbounded. For a concrete illustration, take \(\sigma(t) = t\) (the identity) and let each polynomial be a monomial of the same degree \(d \ge 2\): \(P_\ell(z) = w_\ell z^d\). Then
\begin{equation}
\begin{cases}
& h_1 = w_1 x^d,\quad \\
& h_2 = w_2 (w_1 x^d)^d = w_2 w_1^d x^{d^2},\quad 
\dots,\quad \\
& h_L = \left(\prod_{\ell=1}^L w_\ell^{d^{L-\ell}}\right) x^{d^L}.
\end{cases}
\end{equation}
The magnitude is
\[
|h_L| = \left(\prod_{\ell=1}^L |w_\ell|^{d^{L-\ell}}\right) |x|^{d^L}.
\]
\begin{itemize}
    \item If any weight satisfies \(|w_\ell| > 1\), the product grows super-exponentially with \(L\) because the exponents increase geometrically.
    \item Even if all weights are exactly \(1\), we have \(|h_L| = |x|^{d^L}\). For inputs with \(|x| > 1\), this diverges as \(L\) increases.
    \item If some weights are less than \(1\) but others greater than \(1\), the product can still explode due to the dominant large weights.
\end{itemize}
Using ReLU instead of identity does not prevent explosion: for positive inputs, ReLU behaves exactly like identity, so the same recurrence holds.

\subsection{Comparison with Quadratic Neurons}

Here, we compare the fitting effects between task-based neurons and four different quadratic neurons on the 20 public datasets to demonstrate that task-based neurons are more flexible and have better generalization ability for various datasets than preset neurons. All the operations on 20 datasets and experimental setups are the same as those in the main body. We do not rerun the vectorized symbolic regression; instead, we use the formulas learned in the main body to prototype task-based neurons. The designs for \textit{TN} and the quadratic networks (\textit{QN}) all adopt the fully connected structure, and the activation function is \textit{ReLU} for \textit{QNN} and \textit{Sigmoid} for \textit{TN}. Each dataset was tested 10 times to avoid the accidental effects. The test results are presented in the form of $\mathrm{mean}~ (\mathrm{std})$, which are shown in Table \ref{tab:quadratic}. 

From Table \ref{tab:quadratic}, we can see that compared with four kinds of \textit{QNs}, \textit{TN} uses simpler network structures with fewer parameters, but it gets the lowest MSE and highest accuracy, which highlights the high efficiency and superior fitting ability of \textit{TN}.

Quadratic neurons are a special case of polynomial neurons with fixed neurons. The task-based neurons we design are also based on polynomials but more flexible and adaptive. Data usually have different features in the real world, and the quadratic neurons can fit the data suitable for quadratic functions. Compared to task-based neurons, the form of quadratic neurons is fixed and less flexible, naturally, and the scope of application (data with different characteristics) is not as wide as \textit{TN}. It is seen from our experiments that the depth and width of the network need to become large for quadratic networks to get a satisfactory result, which is inefficient. In contrast, if we can fit the data with a neuron that is flexibly designed for the data characteristics, the characteristics of the data will be easier to capture, which is easier to scale in this task.

\vspace{-0.3cm}
\subsection{Extension to a Broader Functional Class.}
In the earlier experiments, the proposed vectorized symbolic regression searches the space using polynomials as base functions. Actually, the framework we propose is quite flexible. Here, we explore to extend beyond the search for polynomial functions as artificial neurons for constructing neural networks. We observe that, in real-world problems, numerous physical phenomena or laws cannot be adequately described by polynomials. For example, in fields such as signal analysis and electronic circuits, polynomial functions are not the optimal mathematical tools. Instead, researchers commonly employ trigonometric functions to analyze related problems. Therefore, we straightforwardly extend the search space to include the sine and cosine functions, enabling a neural network to better learn from data with salient physical significance.

\begin{table}[htb]
\caption{Test results (trigonometric function as a neuron).}

    \centering
    \vspace{-0.3cm}
     % \scalebox{0.8}
   {\begin{tabular}{l|c}
    \toprule  

 Dataset & california housing \\ \hline
 Function space & \{sine function, cosine function \}  \\ \hline
Symbolic regression result  & \makecell[l]{$0.79 - \mathrm{cos}(\x - \mathrm{sin}(\x) + \mathrm{cos}(\x) $ 
\\ $- 0.67)\mathbf{1}^\top + \mathbf{2}\x^\top + 1$}  \\  \hline
Evaluation index & mean square error  \\ \hline 
Network structure & 15-15-10-5-1  \\  \hline
LN test result  & $0.0564 \pm0.0034$ \\  \hline
TN test result  & $\mathbf{0.0512 \pm0.0021}$ 
 \\  
 
  \bottomrule
    \end{tabular}}
    \label{tab:sine}
\end{table}

Along this direction, we perform the vectorized symbolic regression on the dataset \textit{california housing}, with trigonometric functions as the search space. Subsequently, the obtained regression formula is utilized to construct neurons and neural networks for data fitting. Here, in order to better capture the direct current (DC) component (linear component) in the data, we artificially add a linear component of $\mathbf{2}\mathbf{x}^{\top}+1$ to the regression formula. The test results of \textit{LN} are also compared. The \textit{LN} shares identical hyperparameters and network architecture settings with \textit{TN}. The only difference is the type of neurons used. Each model is tested three times with different random seeds. Detailed information is summarized in Table \ref{tab:sine}. As can be seen, \textit{TN} that uses trigonometric functions in its neurons can outperform the \textit{LN}. It is also better than \textit{TN} which uses polynomials in its neurons. The trigonometric functions as base functions validate that the proposed framework for task-based neurons is flexible to a broader class of functions. We argue that instead of manually selecting types of base functions, we should include more base functions and let the symbolic regression automatically select the optimal expression, as long as the computational resources are supported. 

Additionally, in the future, other functions like exponential functions can be utilized to fit data that exhibit exponential growth or decay trends. The exponential function is closely related to complex numbers and Euler's formula, finding extensive applications in signal processing, circuit analysis, and other fields. Therefore, by expanding the search space to include the exponential function, we can enhance the learning capacity of neural networks.

A novel angle is that we could use the single binary operator from \cite{odrzywolek2026all}: $\texttt{eml}(x,y)=\exp(x)-\log(y)$, which can constitute any elementary function by setting different constants and composition. Thus, symbolic regression with $\texttt{eml}(x,y)$ can be theoretically extended to any basis function automatically.

\subsection{Computational Properties of VSR}

\begin{figure*}[ht!]
	\centering
	\includegraphics[width=1.0\textwidth]{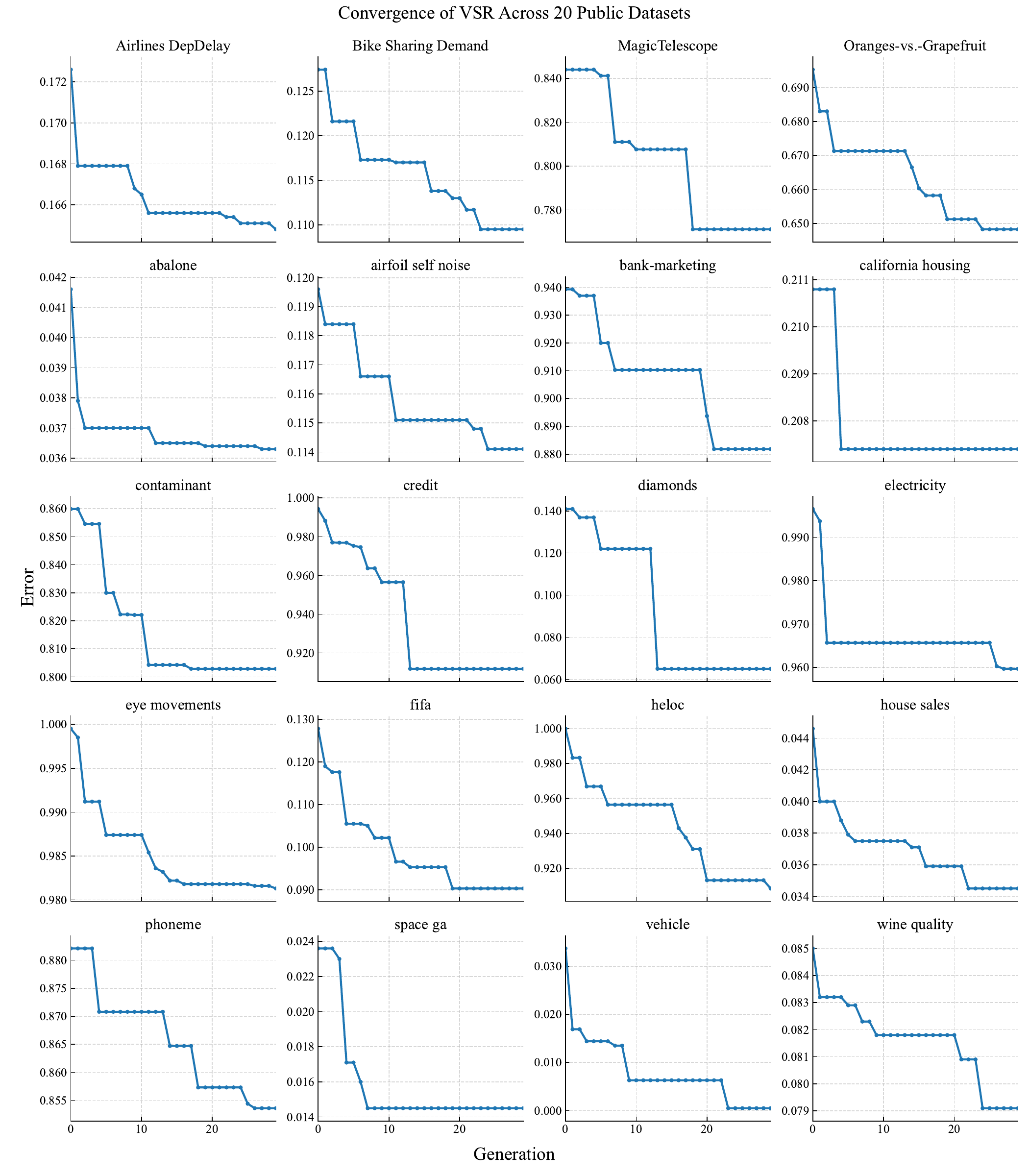}
	\caption{VSR convergence curves on 20 public datasets.}
\label{fig:sr_convergence}
\end{figure*}

To further clarify the convergence behavior of VSR, we visualize the optimization dynamics on 20 public datasets, as shown in Figure \ref{fig:sr_convergence}. Each subplot depicts the evolution of the error across generations during the symbolic search process. All VSR tests used the same hyperparameter settings: generations=30, mutation rate=0.6, population size=3000, crossover rate=0.3. From Figure \ref{fig:sr_convergence}, a consistent convergence pattern can be observed across all datasets. Specifically, the error decreases monotonically or in a step-wise manner as the number of generations increases, indicating that the evolutionary search continuously discovers improved symbolic expressions. Importantly, the optimization typically stabilizes after a limited number of generations, where the error reaches a plateau and no further significant improvement is observed. This behavior suggests that the search process reliably approaches a stable solution rather than oscillating or diverging.

As for the computational cost of VSR, the homogeneous computation in VSR can avoid the combinatorial explosion generated by the increased feature dimension. To analyze the computational behavior of VSR, we evaluate its runtime under different feature dimensions. Synthetic regression datasets are generated using a Gaussian Mixture Model (GMM), following the same data generation protocol as used in the previous experiments. 
The number of samples is 5,000, and the feature dimension is varied as $D \in \{50,100,150,200,250,300\}$.

\begin{figure}[h!]
	\centering
	\includegraphics[width=0.5\textwidth]{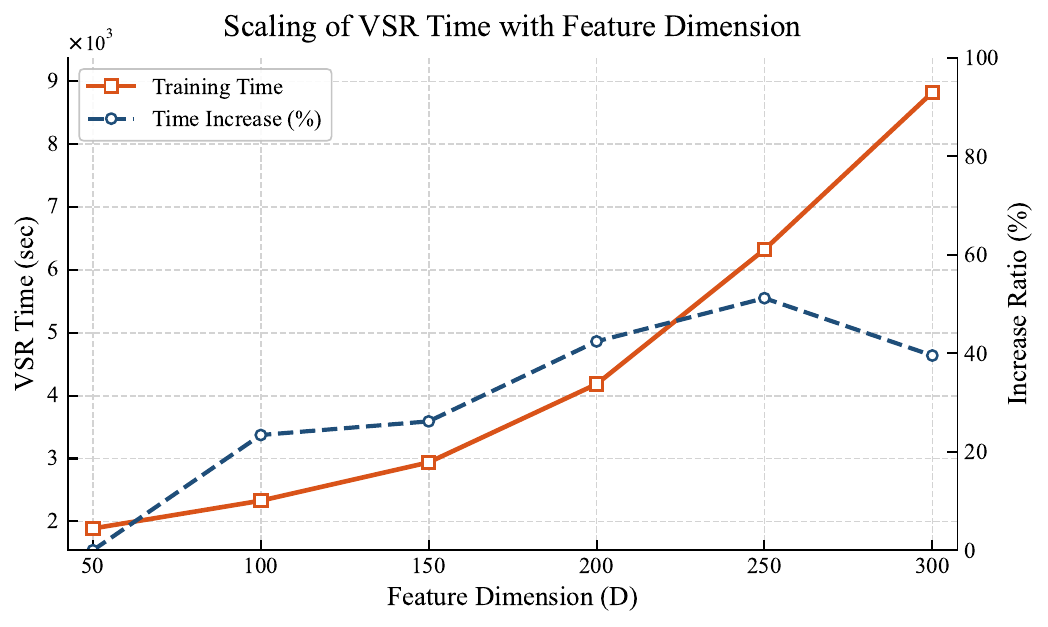}
	\caption{VSR time varies with feature dimensions.}
\label{fig:sr_time}
\end{figure}
For each dataset, VSR is executed five independent times with identical algorithmic settings (population size 2000 and 20 evolutionary generations), and the average runtime is reported as the final result. 
This repeated evaluation reduces the influence of runtime fluctuations and provides a more reliable estimate of the computational cost. To further characterize how the runtime evolves with increasing dimensionality, we compute the relative increase ratio between consecutive dimensions. Let $T(D_i)$ denote the average runtime of VSR when the feature dimension is $D_i$. 
The increase ratio is defined as
\begin{equation*}
\text{IncreaseRatio}(D_i) = \frac{T(D_i) - T(D_{i-1})}{T(D_{i-1})}.
\end{equation*}
This metric measures the relative change in runtime as the feature dimension increases, providing a clearer view of the scaling behavior compared with absolute runtime alone. 

Figure \ref{fig:sr_time} reports both the absolute VSR runtime (left axis) and the corresponding increase ratio (right axis). Although the runtime naturally increases with the feature dimension, the increase ratio remains within a moderate range and does not exhibit abrupt growth. 
This observation indicates that the computational cost of VSR scales in a controlled manner with dimensionality, suggesting that the proposed vectorized formulation maintains stable computational behavior as the feature dimension increases. Note that a recent Nature Computational Science paper \cite{ruan2026discovering} on improving symbolic regression only scales the number of variables to 50 aided with GPU acceleration. In contrast, our VSR is conducted on CPUs, and we have already been able to deal with 300 variables in a few hours.

\section{Sensitivity Experiments}

Here, we investigate the sensitivity of our VSR method to its most influential hyperparameters: i) hyperparameters concerning genetic programming including population size, number of generations, and crossover and mutation rate; ii) hyperparameters concerning the tree structure such as the maximum tree depth. The reason why we analyze VSR is because one of the key innovations in our framework is to use VSR to design task-based neurons.

i) To illustrate the influence of key evolutionary hyperparameters, we construct a controlled synthetic regression task, where the ground-truth expression is predefined as
\[
p(\bm{x}) = \bm{2} (\bm{x} \odot^{3} \bm{x})^\top + \bm{3} (\bm{x} \odot \bm{x})^\top + \bm{5} \bm{x}^\top,
\]
and each input $\bm{x} \in \mathbb{R}^{10}$ is independently sampled from a uniform distribution over $[-50, 50]$.

\noindent\textbf{Baseline configuration.}
All sensitivity curves are generated under a single-variable perturbation protocol: only one hyperparameter is varied at a time, while the remaining hyperparameters are fixed to a consistent baseline. 
Following the default configuration in our implementation, the baseline is defined as
\[
\text{Baseline} = 
\begin{aligned}
\{\;&\text{generations}=20,\\
   &\text{mutation rate}=0.60,\\
   &\text{population size}=2000,\\
   &\text{crossover rate}=0.30\;\},
\end{aligned}
\]
which serves as a fixed reference point for all parameter sweeps.

\noindent\textbf{Sensitivity protocol.}
For each hyperparameter, we sweep over a predefined interval while keeping the rest parameters fixed to the baseline above. Specifically, we consider generations $\{10,20,30,40,50,60\}$, mutation rate $\{0.20,0.40,0.60,0.70,0.80,0.90\}$, population size $\{200,500,1000,2000,4000,8000\}$, and crossover rate $\{0.10,0.20,0.35,0.50,0.60,0.70\}$. 
A fixed random seed is used for all runs to ensure strict reproducibility and to isolate the effect of the single varying hyperparameter.
To emphasize relative sensitivity rather than absolute magnitude, RMSE values are min--max normalized within each sweep, yielding normalized RMSE (NRMSE) curves for both training and validation sets.

\begin{figure*}[!h]
    \centering
    \begin{adjustbox}{width=\textwidth,center}
        \begin{tabular}{cc}
        \subfloat[\label{fig:sens_generations}]{\includegraphics[width=0.5\textwidth]{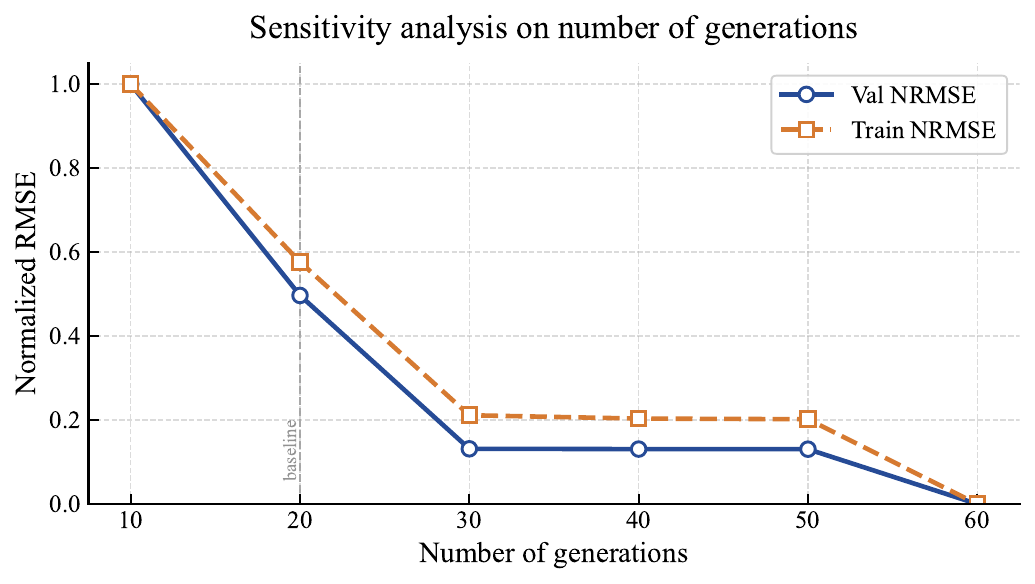}} &
            \subfloat[\label{fig:sens_mutate_pct}]{\includegraphics[width=0.5\textwidth]{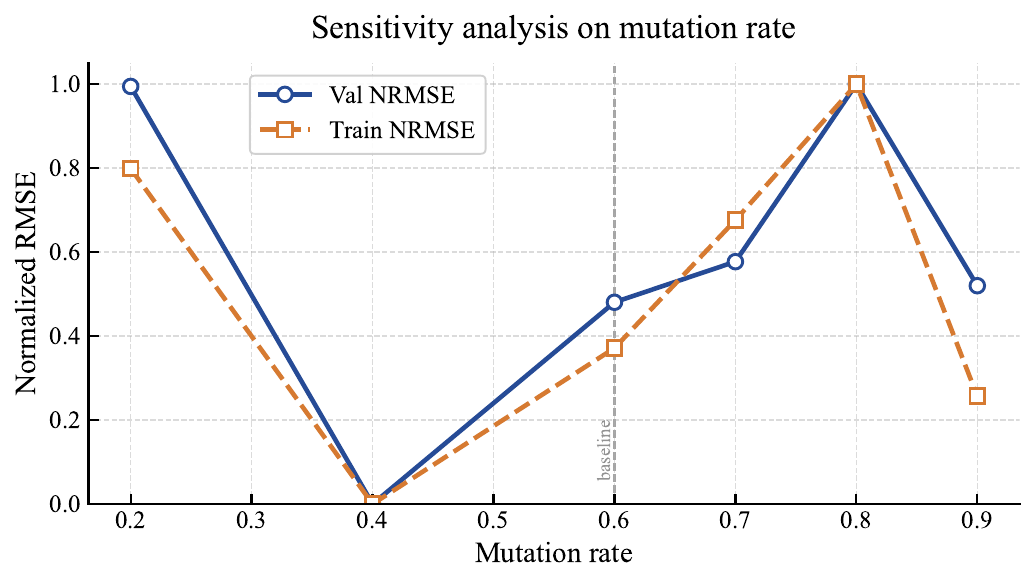}} \\
            \subfloat[\label{fig:sens_pop_size}]{\includegraphics[width=0.5\textwidth]{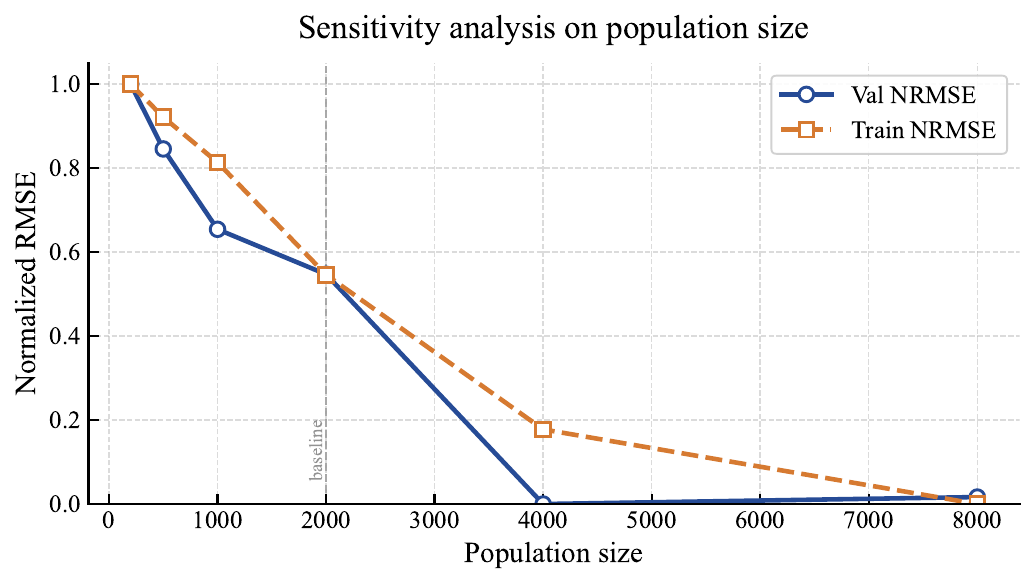}} &
            \subfloat[\label{fig:sens_xover_pct}]{\includegraphics[width=0.5\textwidth]{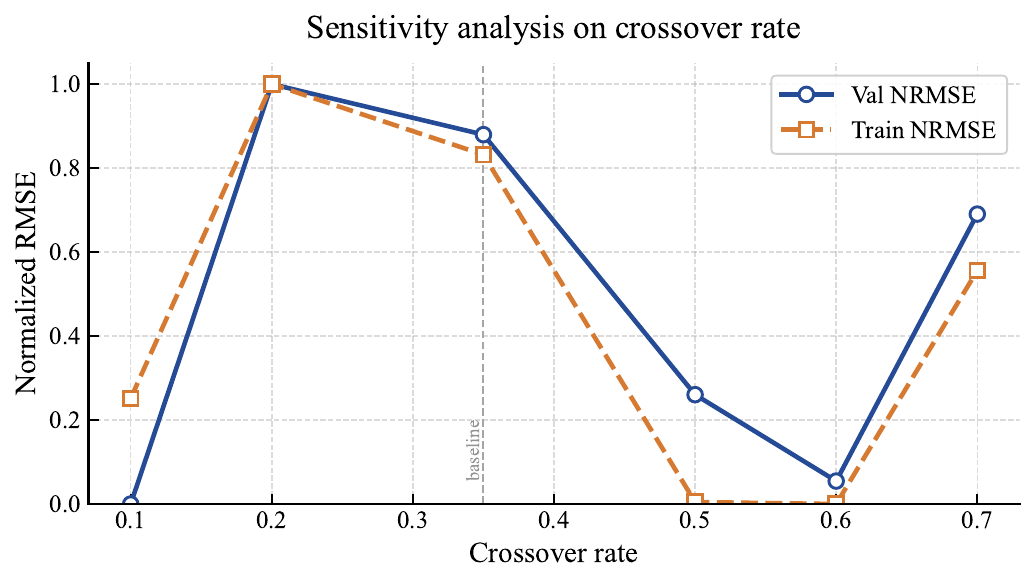}} \\
        \end{tabular}
    \end{adjustbox}
    \caption{
       Parameter sensitivity curves for different parameters.
    }
    \label{fig:curve_line}
\end{figure*}
Figure~\ref{fig:curve_line} summarizes the sensitivity result. All curves report min--max normalized RMSE on the training and validation sets, highlighting how VSR responds to controlled perturbations of a single evolutionary component. Several consistent trends can be observed from Figure \ref{fig:curve_line}:

\begin{itemize}
    \item Number of generations (Fig. \ref{fig:sens_generations}) — Performance improves almost monotonically as the number of generations increases, stabilizing after 40–50 generations. This suggests that VSR effectively converges within a moderate number of iterations, avoiding excessive computational cost.
    \item Mutation rate (Fig. \ref{fig:sens_mutate_pct}) — The model exhibits robustness in a broad range (0.4–0.7), with the best results at 0.4. Mutation thus acts as a crucial yet stable exploratory mechanism, preventing premature convergence without compromising stability.
    \item Population size (Fig. \ref{fig:sens_pop_size}) — Increasing the population leads to steady improvement in both training and validation performance until convergence around 4,000 individuals, beyond which further enlargement yields negligible gains. This indicates that VSR benefits from sufficient search diversity but quickly saturates, showing efficiency in population usage.
    \item Crossover rate (Fig. \ref{fig:sens_xover_pct}) — The optimal performance is achieved near the baseline (0.5–0.6). Extremely low or high crossover probabilities cause noticeable degradation, implying that excessive recombination can disrupt promising symbolic structures, while insufficient crossover limits diversity.
\end{itemize}

\begin{table*}[htbp]
\caption{Leaderboard (Top-$K$) with normalized RMSE.}
\centering
\scalebox{0.8}{
\centering
\begin{tabular}{ccccccc}
\toprule
Rank   & Population & Generations & Crossover & Mutation & Val NRMSE (min–max) & Train NRMSE (min–max) \\
\midrule
1  & 2000 & 60 & 0.3 & 0.6 & 0.000 & 0.000\\
2  & 2000 & 50 & 0.3 & 0.6 & 0.283 & 0.353\\
3  & 2000 & 40 & 0.3 & 0.6 & 0.284 & 0.355 \\
4  & 2000 & 30 & 0.3 & 0.6 & 0.285 & 0.369 \\
5  & 4000 & 20 & 0.3 & 0.6 & 0.304 & 0.496 \\
6  & 8000 & 20 & 0.3 & 0.6 & 0.327 & 0.249 \\
7  & 2000 & 20 & 0.3 & 0.4 & 0.405 & 0.545 \\
8  & 2000 & 20 & 0.1 & 0.6 & 0.708 & 1.000 \\
9  & 2000 & 20 & 0.6 & 0.6 & 0.769 & 0.830 \\
10 & 2000 & 20 & 0.5 & 0.6 & 1.000 & 0.833 \\
\bottomrule
\end{tabular}
}
\label{tab:leaderboard-nrmse}
\end{table*}

Table~\ref{tab:leaderboard-nrmse} summarizes the Top-$K$ configurations ranked by validation NRMSE after min--max normalization within the compared settings (i.e., $0$ denotes the best relative performance and $1$ the worst within that sweep, rather than zero absolute error). The leading entries exhibit consistently superior relative validation performance, with training NRMSE in close agreement, indicating no evident overfitting. In practice, a medium population ($\approx 2,000$--$4,000$) and a moderate evolutionary depth ($\approx 30$--$50$ generations), together with balanced crossover and mutation rates ($\approx 0.3$--$0.6$), offer a favorable accuracy--efficiency trade-off; increasing population well beyond $4,000$ or generations beyond $60$ yields diminishing returns in our controlled setup.

ii) Here, we explain why the maximum tree depth limit of 6 is selected.

First, we note that the depth limit of 6 is a default value in widely used symbolic regression libraries such as gplearn and DEAP, chosen as a pragmatic balance between expressive power and complexity. In our preliminary experiments, the depth limit of 6 consistently yielded good results, which is why we retained the default. In addition, it should be noted that "tree depth" refers to "maximum tree depth". The depth of each tree is not fixed but randomly sampled under the constraint of "maximum tree depth".

Second, the evolutionary process already incorporates implicit parsimony pressure. In genetic programming, the fitness function is the sole optimization objective (here, MSE). However, the tournament selection and the recombination operators naturally favor shorter, more robust trees because the search starts from shallow to deep. Unnecessarily deep structures are less likely to survive multiple generations unless they confer a substantial fitness advantage \cite{silva2003dynamic}. This phenomenon is well documented in the symbolic regression literature as a form of bloat control—the search dynamics themselves penalize complexity without an explicit multi-objective term \cite{silva2003dynamic}. Simultaneously maximizing fitness and minimizing tree depth is a more systematic approach in principle. However, such a formulation introduces additional hyperparameters (e.g., the trade-off coefficient) and increases search complexity. Our current implementation, relying on the natural parsimony of GP, has proven sufficient for the tasks at hand. We will note in the revised manuscript that multi-objective symbolic regression could be explored in future work to further automate complexity control.

\begin{figure}[h!]
	\centering
	\includegraphics[width=0.5\textwidth]{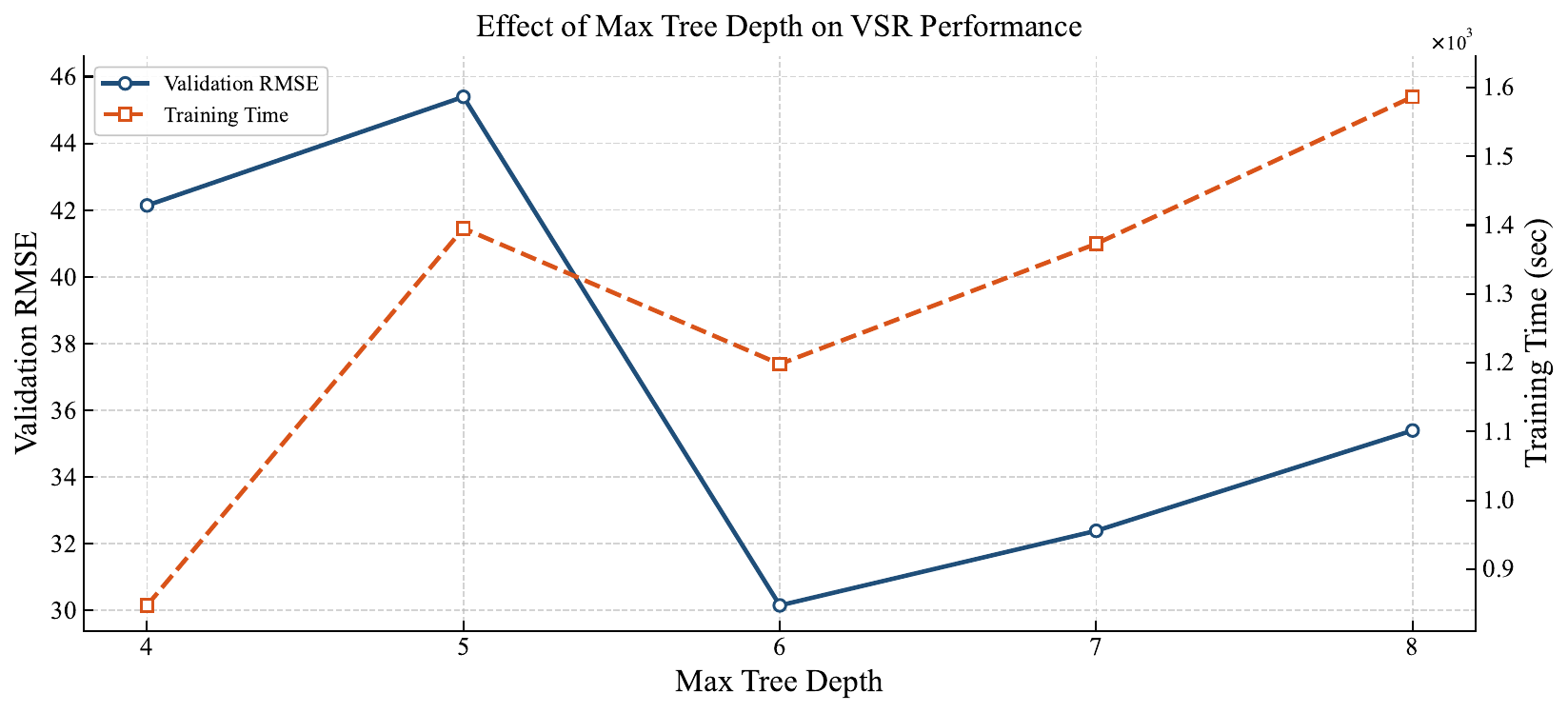}
	\caption{The effect of the max tree depth on VSR performance.}
\label{fig:sr_depth}
\end{figure}

Third, to further justify the default choice of the maximum tree depth, we conducted a sensitivity study on a fixed 50-dimensional GMM regression task by varying maximum tree depth from 4 to 8. As shown in Figure \ref{fig:sr_depth}, the effect of tree depth is non-monotonic. In particular, increasing the depth from 4 to 5 does not improve performance, whereas depth 6 yields the lowest validation RMSE among all tested settings. More importantly, this best-performing setting does not incur the highest computational cost: the training time at depth 6 is lower than that at depth 5 and substantially lower than that at depths 7 and 8. Beyond depth 6, increasing the tree depth leads to consistently higher training time, while the validation RMSE deteriorates rather than improves. This suggests that excessively deep expression trees enlarge the search space and increase the risk of over-fitting. Therefore, setting 
$\texttt{max\_tree\_depth}=6$ is empirically well justified. It provides the best validation performance in this experiment while avoiding the unnecessary computational overhead associated with deeper trees. We do not claim that depth 6 is universally optimal for all tasks, but the above results support it as a reasonable and effective default setting in our framework.

\section{Extending Task-based Neurons to CNNs} 
\label{extend_to_cnn}

While task-based neurons were initially developed for tabular learning, to make the proposed framework more impactful, we extend it to vision tasks. 
Without bells and whistles, we directly introduce an autoencoder-based dimensionality reduction to transform high-dimensional visual inputs into compact latent vectors. Latent features can be extracted from these latent vectors to represent image features, which is evidenced by numerous image-based autoencoders \cite{gondara2016medical, faysal2025denomae}. The process of extending task-based neurons from tabular data to images follows a three-step pipeline. 
This pipeline connects visual representation learning with symbolic neuron construction through VSR and reparameterization, and the inductive biases are transferred to convolutional layers.

\textbf{Step 1: Autoencoder-based latent feature extraction.}
Given an image dataset, each input image \( X \in \mathbb{R}^{C_{\mathrm{in}}\times H\times W} \) is first trained by an autoencoder. Then, the encoder is used to obtain a compact latent vector \( \boldsymbol{z} \in \mathbb{R}^{d} \) at the bottleneck layer. 
The encoder compresses the spatially structured input into a low-dimensional latent representation that captures the essential task-relevant semantics while removing redundant pixel-level variations. 
This step converts 2D visual signals into a vectorized form suitable for VSR. Otherwise, VSR has to be conducted in hundreds of thousands of dimensions.

\begin{figure*}[!htbp]
	\centering	
    
	% \begin{adjustbox}{width=1.1\textwidth,center}		
	% 	\subfloat[\label{fig:fcn_case}]{\includegraphics[width=0.5\textwidth]{Figure/fcn_case.pdf}}
	% 	\subfloat[\label{fig:conv_case}]{\includegraphics[width=0.2\textwidth]{Figure/conv_case.pdf}}
	% \end{adjustbox}		

    \begin{adjustbox}{width=1\textwidth,center}
        \begin{tabular}{cc}
            \subfloat[\label{fig:fcn_case}]{\includegraphics[width=0.5\textwidth]{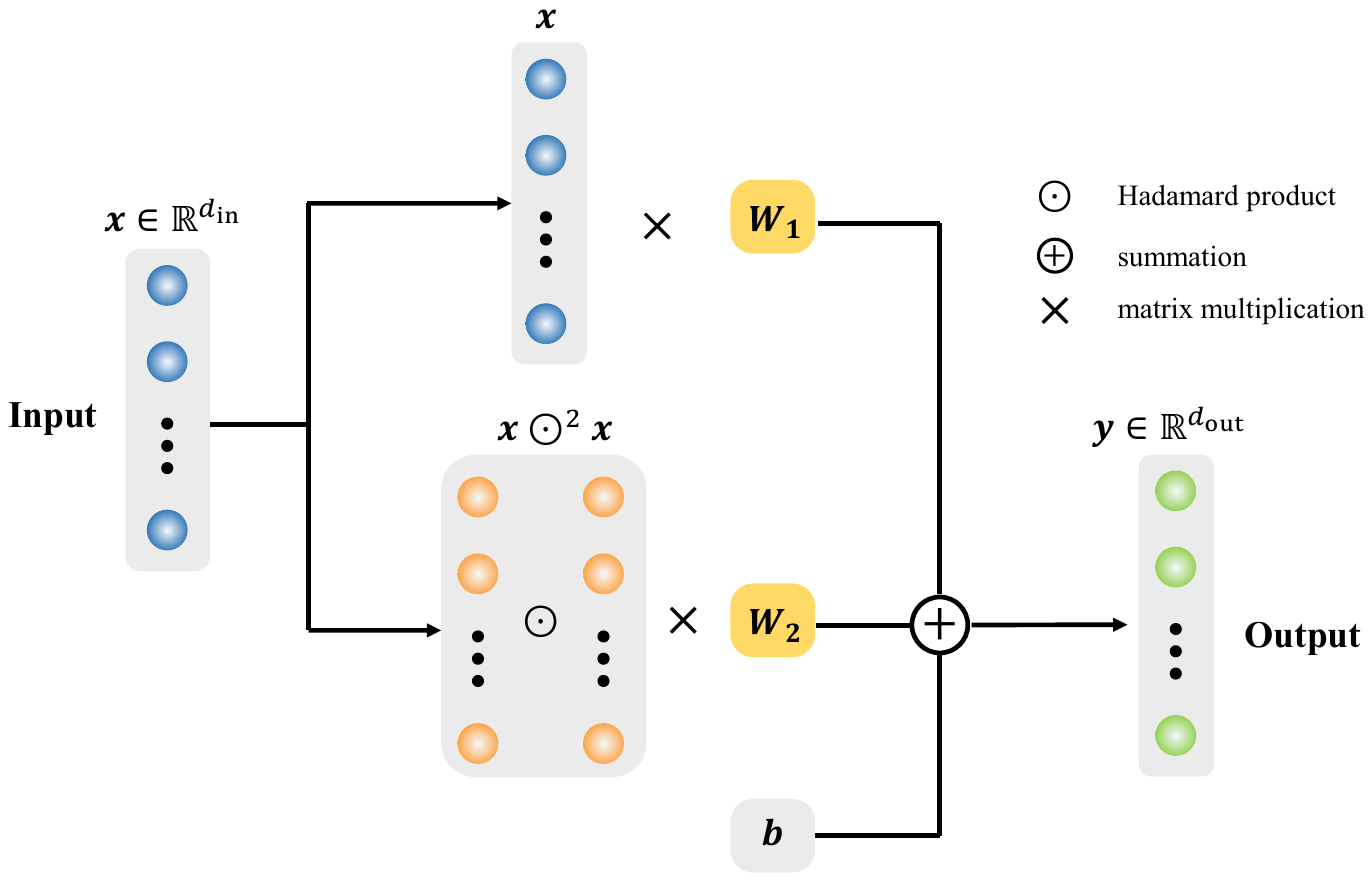}} &
            \subfloat[\label{fig:conv_case}]{\includegraphics[width=0.5\textwidth]{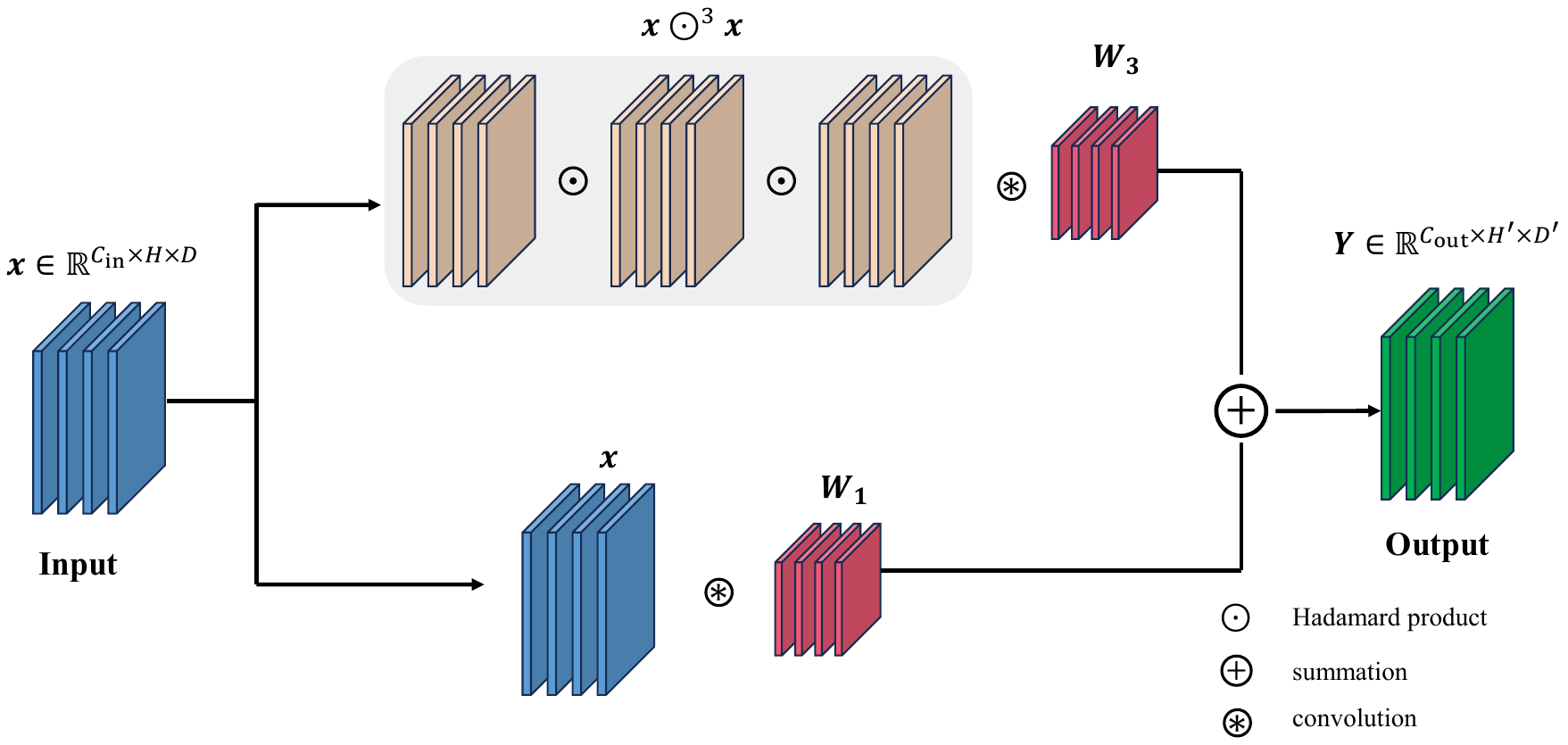}} 
        \end{tabular}
    \end{adjustbox}
	\caption{	
 Illustration of task-driven neurons in different architectures. (a) Fully connected case: the input vector $\x \in \mathbb{R}^{d_{\text{in}}}$ is expanded into polynomial terms (e.g., $\x$, $\x\odot\x$), each multiplied by a learnable weight matrix $\W_k$, and the results are summed to form the output $\y \in \mathbb{R}^{d_{\text{out}}}$. (b) Convolutional case: the input feature map $\x \in \mathbb{R}^{C_{\text{in}}\times H \times D}$ is expanded into higher-order terms (e.g., $\x$, $\x\odot^3\x$), each convolved with a learnable kernel $\W_k$, and the results are summated to produce the output $\bm{Y} \in \mathbb{R}^{C_{\text{out}}\times H' \times W'}$.}
	\label{fig:diff_arc}
\end{figure*}

\textbf{Step 2: VSR on latent representations.}
On the latent vectors \(\boldsymbol{z}\), we apply VSR to discover analytic expressions that characterize the nonlinear mapping between the latent features and the target labels. 
The outcome is a task-specific symbolic expression,
\[
f(\boldsymbol{z}) = \sum_{k} \langle \boldsymbol{a}_k, \boldsymbol{z}\odot^{p_k}\z \rangle,
\]
where the exponents \(\{p_k\}\) define the structural form of nonlinearity to be embedded in the network. 
This symbolic discovery phase isolates the structural essence of the task in an interpretable algebraic form.

\textbf{Step 3: Symbolic-to-convolutional reparameterization.}
Once the symbolic expression is obtained, as Figure \ref{fig:diff_arc} shows, we convert it into a trainable convolutional module by assigning a learnable kernel \(\W_k\) to each symbolic term \( \boldsymbol{z}\odot^{p_k}\z \). 
The exponents \(\{p_k\}\) act as a structural prior, guiding how input feature maps should be nonlinearly transformed within each convolutional layer. 
Formally, the symbolic expression is translated into the convolutional operation
\[
Y = \sum_{k} \mathtt{Conv}(\X\odot^{p_k}\X;\, \W_k),
\]
where the convolutional weights \(W_k\) will be optimized through gradient descent, while the symbolic powers \(\{p_k\}\) remain fixed. 
This reparameterization embeds the task-specific symbolic structure into a differentiable CNN block. It ensures that the learned symbolic structure acts as a reusable inductive bias, transforming convolutional kernels from purely linear filters into structurally informed nonlinear operators that better align with the semantics of the visual task.

Through this formulation, a symbolic neuron is reformulated as a set of nonlinear convolutional operators whose algebraic form is determined by the symbolic expression discovered by VSR. 
In contrast to conventional CNNs, where the convolution is inherently linear and nonlinearity is introduced only through subsequent activation functions, 
the proposed task-based convolution directly incorporates nonlinear mappings within the convolution itself. 
Each symbolic term defines a distinct polynomial transformation on the input feature maps, and their weighted combination constitutes a prior-informed nonlinear kernel.

\section{Details of All Models Used in Comparative Experiments and Two Datasets}

\textit{XGBoost:} We use XGBoost, a tree-based gradient boosting model, for regression. The model consists of up to 1,000 regression trees trained with a learning rate of 0.08 and a maximum depth of 7. Early stopping is applied with a patience of 40 rounds using a validation set. To improve generalization, we use the row subsampling (subsample=0.75) while retaining all features (colsample-bytree=1.0). The objective is reg:squarederror, and the standard L2 regularization is applied (reg-lambda=1.0).

\textit{LightGBM:} We utilize LightGBM for regression with the GOSS (Gradient-based One-Side Sampling) boosting strategy. The model is trained with a maximum of 2,000 iterations using the mean squared error (MSE) as the optimization metric. We allow unrestricted tree depth  and rely on GOSS for sample selection, which retains instances with large gradients while subsampling the rest. Default settings are used for feature and data sampling. 

\textit{CatBoost:} We use CatBoost for regression, with the objective metric set to Root Mean Squared Error (RMSE). The model is trained using the ordered boosting for 3,000 iterations with a maximum tree depth of 4 and a learning rate of 0.08. To control overfitting, the L2 regularization is applied. Early stopping with a patience of 1,000 rounds is enabled using a separate validation set. CatBoost automatically handles categorical features and performs internal feature border selection.

\textit{TabNet:} We use a TabNet model with 3 decision steps, a shared feature transformer, and sparse attention masking. The decision and attention embedding dimensions are both set to 10 ($n_d=n_a=10$). Each decision step includes both a shared and a step-specific GLU-based feature transformer. No categorical embeddings are used. The final output is generated via a linear mapping from the aggregated decision representation. BatchNorm is applied throughout, and Sparsemax is used to select the mask. The total trainable parameter count is approximately 10K.

\textit{TabTransformer:} We use a TabTransformer model consisting of 4 stacked Transformer blocks, each with 8-head multi-head self-attention, followed by a feed-forward network with 128 hidden units and GELU activation. All continuous features are embedded into 32-dimensional token vectors via a standard linear embedding layer. Each Transformer block includes dropout (0.2 for attention, 0.1 for FFN) and LayerNorm-based residual connections. The final pooled embedding is fed into a regression head to predict the target variable. The total trainable parameter count is approximately 51K.

\textit{FT-Transformer:} We employ the FT-Transformer model, which consists of 5 stacked transformer encoder blocks. Each block utilizes a lightweight LinearAttentionLinformer mechanism with a key-value compression factor of 0.5 to reduce the computational complexity of self-attention. All continuous features are embedded into 32-dimensional token vectors using a linear embedding layer. Each encoder block contains a feed-forward network with REGLU activation and a hidden size of 84, followed by a 42-unit projection layer. Residual connections with LayerNorm and dropout (0.2 for attention, 0.1 for FFN) are applied at each step. The total number of trainable parameters is approximately 45K.

\textit{DANETs:} We use DANet, a deep attentive network designed for structured regression tasks. The model starts with an initialization block followed by 2 BasicBlocks, each comprising two depth-wise 1D convolution layers with grouped connections (groups=5), Group Batch Normalization (GBN), and attention-based feature masking. The attention masks are learned via the LearnableLocality module with Entmax15 activation, enabling soft and sparse feature selection. The final representation is processed through a 3-layer MLP head to produce scalar regression outputs. Dropout with p=0.1 is applied after each block. The total number of trainable parameters is approximately 25K.

\textbf{High-energy Particle Collision Prediction}.
High-energy particle collision experiments are an important means to help people understand the fundamental composition and evolution of our universe. In the collision, the medium quickly becomes a soup of deconfined quarks, gluons, and partons within the first few microseconds. Quark-Gluon Plasma (QGP) is the name of this blazing and dense fireball. To investigate the distinctive qualities and evolution of the QGP, particles created at each phase of the medium developed in high-energy collision tests are employed as a probe. The important QGP phase probe is the $J / \psi $ meson, which is the bound state of the charm quark ($c$) and its antiparticle ($\bar{c}$). The invariant mass spectrum of the particle is very useful for selecting the distribution region of the $J / \psi $ signal. In high-energy physics, predicting the invariant mass spectrum of the particle is a critical research topic. 

CERN collected the collision data and disclosed a dataset on Kaggle \footnote{ https://www.kaggle.com/datasets/fedesoriano/cern-electron-collision-data}. This data set is used to predict the invariant mass of the two electrons in the collision experiment by the properties of electrons such as energy, charge, transverse momentum, and pseudorapidity. It consists of 99,915 observations, and each observation has 16 features.

\textbf{Asteroid Prediction}. Asteroids are petite celestial bodies composed of rocks that revolve around the Sun in elliptical orbits. They are remnants left over from the early stages of solar system formation. Two categories of asteroids, namely Near-Earth Asteroids (NEAs) and Potentially Hazardous Asteroids (PHAs), attract particular interest among researchers. NEAs are characterized by their proximity to Earth's orbit, indicating the potential risk of collision with our planet. PHAs, in particular, raise concerns due to their potential to approach Earth's orbit even closer, coupled with their substantial size, which could result in significant damage upon collision with Earth. The prediction of diameter holds crucial significance in identifying NEAs and PHAs.  

To predict the diameters of NEAs and PHAs, physicists from the Jet Propulsion Laboratory of CalTech collected a dataset that consists of 137,636 observations and each observation has 19 features, available on Kaggle \footnote{https://www.kaggle.com/datasets/basu369victor/prediction-of-asteroid-diameter}.

\section{Theorem \ref{superiority} and Its Proof}
\label{proof}

Without loss of generality, we use an MLP to establish our theorem. Let us first introduce the necessary notations. Specifically, we suppose that the input $\x$ has $d$ features, and for a vector  
\begin{equation}
\x:=\left[x_{1}, x_{2}, \ldots, x_{d}\right]^{\top} \in \mathbb{R}^{d}, 
\end{equation}
we have the vector-valued function by the activation function
\begin{equation}
\sigma(\x):=\left[\sigma\left(x_{1}\right), \ldots, \sigma\left(x_{d}\right)\right]^{\top} .
\end{equation}
A task-based network of $L$ layers has $L-1$  hidden layers and one output layer, where the $l$-th layer has $n_l$ neurons. Specifically, we have $n_1=d$ and $n_L=1$. Suppose that the $l$-th layer has $\mathtt{card}(K)$ weight matrices  $\mathbf{W}_{k}^{l} \in \mathbb{R}^{n_l\times n_{l-1}}, k\in K$ and the bias vector $\mathbf{b}^{l} \in \mathbb{R}^{n_l}$, where $K=\{k_1,k_2,\cdots,k_{\mathtt{card}(K)}\}$ is a set collecting the orders of monomial terms. Then, each layer conducts the following computation:
\begin{equation}
    f_l = \sum_{i=1}^{\mathtt{card}(K)} \mathbf{W}_{k_i}^{l}(\z^{l-1}\odot^{k_i}\z^{l-1})^\top + \b^l ,
\end{equation}
where $\z^{l-1}$ is the output of the $(l-1)$-th layer. Thus, an $l$-layer task-based network is denoted by
\begin{equation}
\mathcal{N}_l:=f_{l} \circ \sigma \circ f_{l-1} \circ \sigma \circ \cdots \circ f_{2} \circ \sigma \circ f_{1}.
\end{equation}

We further define the error function caused by $\mathcal{N}_L$ as
\begin{equation}
\mathbf{e}_{l}=\mathbf{f}(\x)- \mathcal{N}_l(\x), \x \in \mathbb{R}^{s},
\end{equation}
where the weight matrices and bias vectors are parameters to be learned from the optimization problem:
\begin{equation}
\min \left\{\left\|\mathbf{e}_{l}\left(\mathbf{W}_{k}^l, \mathbf{b}^{l} ; \cdot\right)\right\|^{2}: \mathbf{W}_{k}^l \in \mathbb{R}^{n_{l} \times n_{l-1}}, \mathbf{b}^{l} \in \mathbb{R}^{n_{l}} \right\}.
\end{equation}
We define the optimal  error as 
\begin{equation}
\mathbf{e}_{l}^{*}(\x):=\mathbf{f}(\x)-N_{l}(\x), \text { for } \x \in \mathbb{R}^{d},
\end{equation}
where $N_l^* \in \mathcal{N}_l$. Specially, we have $N_1^*=f_1^*$. Then, we have the following theorem: 

\begin{theorem}
For any $l>1$, we have
$\left\|\mathbf{e}_{l}^{*}\right\| \leq\left\|\mathbf{e}_{1}^{*}\right\|$.
\label{superiority}
\end{theorem}

Our idea of proving Theorem 1 is to establish a special construction $N_l^{\dagger}$ for $\mathcal{N}_l$ such that it can approximate $\f$ at least as good as $N_1^*$. Suppose $\e_l^\dagger = \f-N_l^{\dagger}$, then $\|\e_l^\dagger\|\leq\|\e_1^*\|$. While it is not certain if $N_l^{\dagger}$ is the optimal; therefore, the optimal error $\|\e_l^*\|\leq\|\e_l^\dagger\|$. As a result, we have 
\begin{equation}
\|\e_l^*\|\leq\|\e_l^\dagger\|\leq \|\e_1^*\|.
\end{equation}
Our proof is largely based on \cite{xu2025multi}. Therefore, we inherit notations of \cite{xu2025multi} in the following for convenience.

Let $\mathcal{H}$ be a Hilbert space and $\Omega \in \mathcal{H}$ be a closed set. $\Omega$ is not necessarily convex. We first characterize the properties of a best approximation from  $\Omega$  to an $\f$ in $\mathcal{H}$. \cite{xu2025multi} presents Lemma \ref{important_lemma}, a necessary condition and a sufficient condition, respectively, for an element $\g_{0}$ to be a best approximation. It first defines a star-shape set centered at a given  $\g_{0} \in \Omega$  by

\begin{equation}
\Omega\left(\g_{0}\right):=\left\{\g \in \Omega: \lambda \g+(1-\lambda) \g_{0} \in \Omega \text { for all } \lambda \in(0,1)\right\}. 
\end{equation}
Clearly, a point $\g$ of  $\Omega$  is in  $\Omega\left(\g_{0}\right)$  if and only if the line segment connecting the two points  $\g_{0}$ and $\g$ is completely located in  $\Omega$.

\begin{lemma}
Suppose that $\mathcal{H}$  is a Hilbert space and  $\Omega$  is a closed set in  $\mathcal{H}$. Let  $\f \in \mathcal{H} \backslash \Omega $. 
\begin{enumerate}
    \item If  $\g_{0} \in \Omega$  is a best approximation from  $\Omega$  to $\f$ , then for all  $\g \in \Omega\left(\g_{0}\right)$,
\begin{equation}
    \left\langle \f-\g_{0}, \g-\g_{0}\right\rangle \leq 0 .
    \label{eqn:inequality}
\end{equation}
\item If $\g_{0} \in \Omega $ and Eq. \eqref{eqn:inequality} holds for all $ \g \in \Omega $, then $\g_{0} $ is a best approximation from $ \Omega $ to $\f $.
\end{enumerate}
\label{important_lemma}
\end{lemma}

Now, let us introduce our construction. Suppose $N_1^*(\x)=f_1(\x)$ is the optimal approximation to $\f$, and $\e_1^*=N_1^*(\x)-\f$, then we use $\tilde{N}_l(\x)=f_l\circ f_{l-1}\circ \cdots \circ f_1^*$ to approximate the error $\e_1^*$, where only weights and biases of $f_l,f_{l-1},\cdots,f_2$ are optimized. Mathematically, 
\begin{equation}
    \min \{\|\e_1- \tilde{N}_l\|: \mathbf{W}_{k}^i \in \mathbb{R}^{n_{i} \times n_{i-1}}, \mathbf{b}^{i} \in \mathbb{R}^{n_{i}}, i=2,\cdots,l, k\in K\}.
\label{surrogate_network}
\end{equation}
Suppose that $\tilde{N}_l^*(\x)$ achieves the optimal error in the setting of Eq. \eqref{surrogate_network}, and the error is $\tilde{\e}_l^* = \e_1^*- \tilde{N}_l^*(\x)$, then we have the following theorem:   

\begin{theorem}
For all  $l \geq 2 $, 
\begin{equation}
\left\|\mathbf{e}_{1}^{*}\right\|^{2} \geq\left\|\tilde{N}_l^*\right\|^{2}+\left\|\tilde{\e}_l^*\right\|^{2}.
\end{equation}
Then, 
\begin{equation}
\left\|\mathbf{e}_{1}^{*}\right\| \geq\left\|\tilde{\e}_l^*\right\|.
\end{equation}
\end{theorem}

\begin{proof}

(ii) Note that for all  $l\geq 2$, we have $\mathbf{e}_{1}^{*}=\tilde{N}_l^*+\tilde{\mathbf{e}}_{l}^{*} $. Hence, we have that
\begin{equation}
\left\|\mathbf{e}_{1}^{*}\right\|^{2}=\left\langle\tilde{N}_l^*+\tilde{\mathbf{e}}_{l}^{*}, \tilde{N}_l^*+\tilde{\mathbf{e}}_{l}^{*}\right\rangle = \|\tilde{N}_l^*\|^2+ \|\tilde{\mathbf{e}}_{l}^{*}\|^2+2\left\langle\tilde{N}_l^*, \tilde{\mathbf{e}}_{l}^{*}\right\rangle.
\end{equation}

Apply Lemma \ref{important_lemma} with  $\f:=\e_{1}^{*}, \g_{0}:=\tilde{N}_l^*$ and $\g:=0 $. Since $\mathbf{0}$ belongs to   $\mathcal{N}_l$, we see that
\begin{equation}
    0\in \mathcal{N}_l \text { and } \lambda 0+(1-\lambda) \tilde{N}_l^* \in \mathcal{N}_l, \text { for all } \lambda \in(0,1) .
\end{equation}
Then, according to Eq. \eqref{eqn:inequality}, we have 
\begin{equation}
    \left\langle \e_{1}^{*}- \tilde{N}_l^*, 0-\tilde{N}_l^* \right \rangle \leq 0
\end{equation}
Thus,  
\begin{equation}
\left\langle\tilde{N}_l^*, \tilde{\mathbf{e}}_{l}^{*}\right\rangle=\left\langle\tilde{N}_l^*, \e_{1}^{*}-\tilde{N}_l^*\right\rangle \geq 0,
\end{equation}
which leads to
\begin{equation}
\left\|\mathbf{e}_{1}^{*}\right\|^{2} \geq\left\|\tilde{N}_l^*\right\|^{2}+\left\|\tilde{\e}_l^*\right\|^{2}.
\end{equation}
\end{proof}

Our construction $N_l^\dagger$ equals $\tilde{N}_l+N_1^*$, which is equivalent to augment an identity mapping onto $\tilde{N}_l$ to pass the result of $N_1^*$ to the final layer that sums $\tilde{N}_l$ and $N_1^*$ together. The error $\e_l^\dagger$ of $N_l^\dagger$ is $\e_l^\dagger=\tilde{\e}_l^*$, since $\tilde{\e}_l^* = \e_1^*- \tilde{N}_l^*(\x) = \f-\tilde{N}_1-\tilde{N}_l^*=\f-(\tilde{N}_1+\tilde{N}_l^*)$.

\begin{proof}[Proof of Theorem 1]
    With the construction $N_l^\dagger$ that can already achieve the error $\e_l^\dagger$ no greater than $\e_1^*$, the optimal error in $\mathcal{N}_l$ must be also no greater than $\e_1^*$. 
\end{proof}

Theorem \ref{superiority} indicates that a symbolic formula's low MSE in a standalone, shallow regression task serves as a reliable proxy for its performance within a deep neural network.

Here, we conduct an additional study on 10 public regression datasets to evaluate this point. For each dataset, we first obtain a symbolic expression via VSR and record its regression error as the proxy metric, which is the MSE. We then measure the performance gap of linear networks and task-based networks, referred to as $\texttt{error}(\mathrm{LN} - \mathrm{TN})$, in the same structure. Finally, we analyze the correlation of the VSR error and $\texttt{error}(\mathrm{LN} - \mathrm{TN})$. The absolute error $\texttt{error}(\mathrm{TN})$ is heavily influenced by dataset properties: noise level, intrinsic complexity, and scale of target values. A “good” TN on a hard dataset may still have high absolute error, while a “poor” TN on an easy dataset may have low absolute error. This would confound correlation. Therefore, we use the relative error to isolate the added benefit of task‑based neurons from dataset‑specific factors.

\begin{figure}[h!]
\centering
\includegraphics[width=\linewidth]{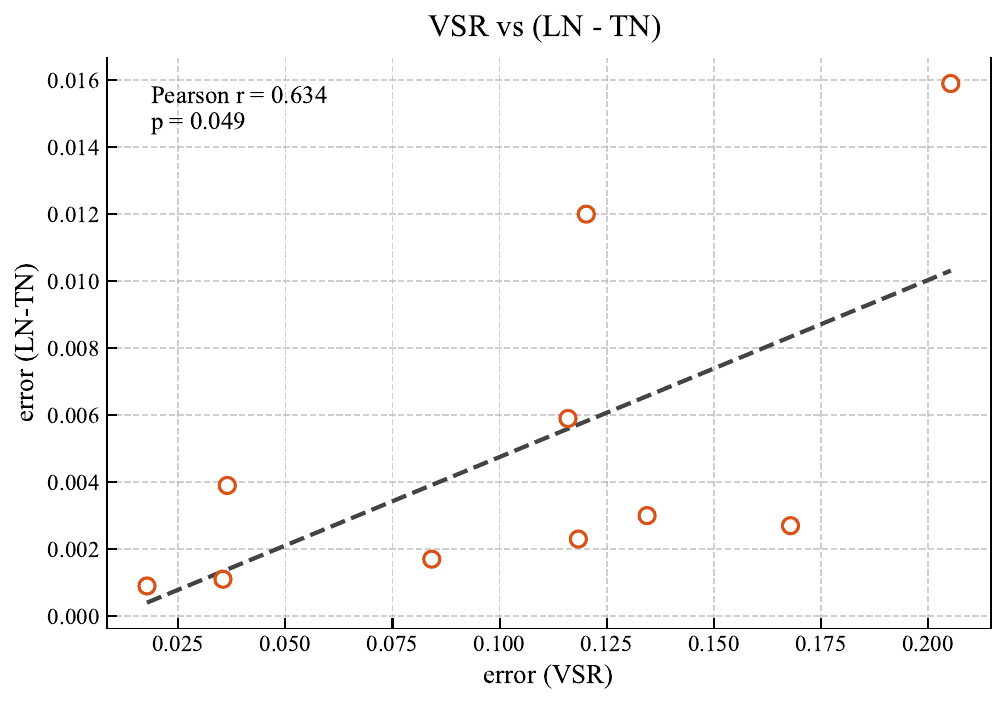}
\caption{Correlation analysis on public regression datasets in Table 9 of Supplementary Materials: symbolic expressions with lower regression error tend to yield larger performance improvements when deployed within deep models.}
\label{fig:vsr_tn_ln}
\end{figure}

As shown in Figure \ref{fig:vsr_tn_ln}, we observe a clear positive correlation between the VSR error and $\texttt{error}(\mathrm{LN} - \mathrm{TN})$, with a Pearson correlation coefficient of $r = 0.634$ and a statistically significant $p$-value of $0.049$. While the correlation is moderate, it is statistically significant and supports the claim that symbolic expressions with lower regression error tend to yield larger performance improvements when deployed within deep models. These results provide empirical evidence that the evolutionary search guided by regression error is not arbitrary, but aligned with the effectiveness of the resulting neuron in deep architectures.

\section{Network Structure in Image Experiments}

\begin{figure*}[h]
    \centering
    \begin{adjustbox}{width=0.8\textwidth,center}
        \subfloat[]{
            \includegraphics[width=0.35\textwidth]{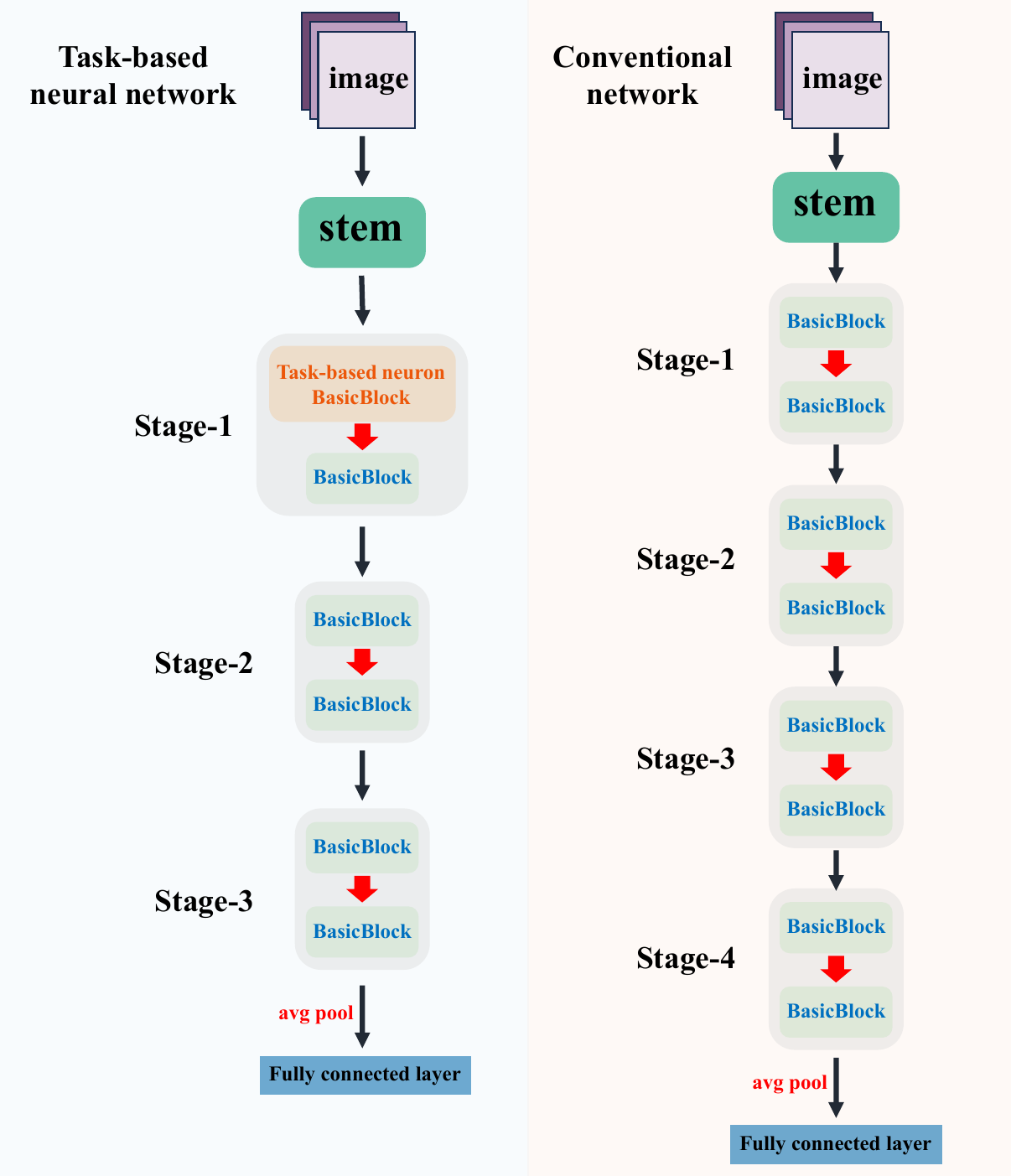}
            \label{fig:network_overview}
        }
        \hfill
        \subfloat[]{
            \includegraphics[width=0.5\textwidth]{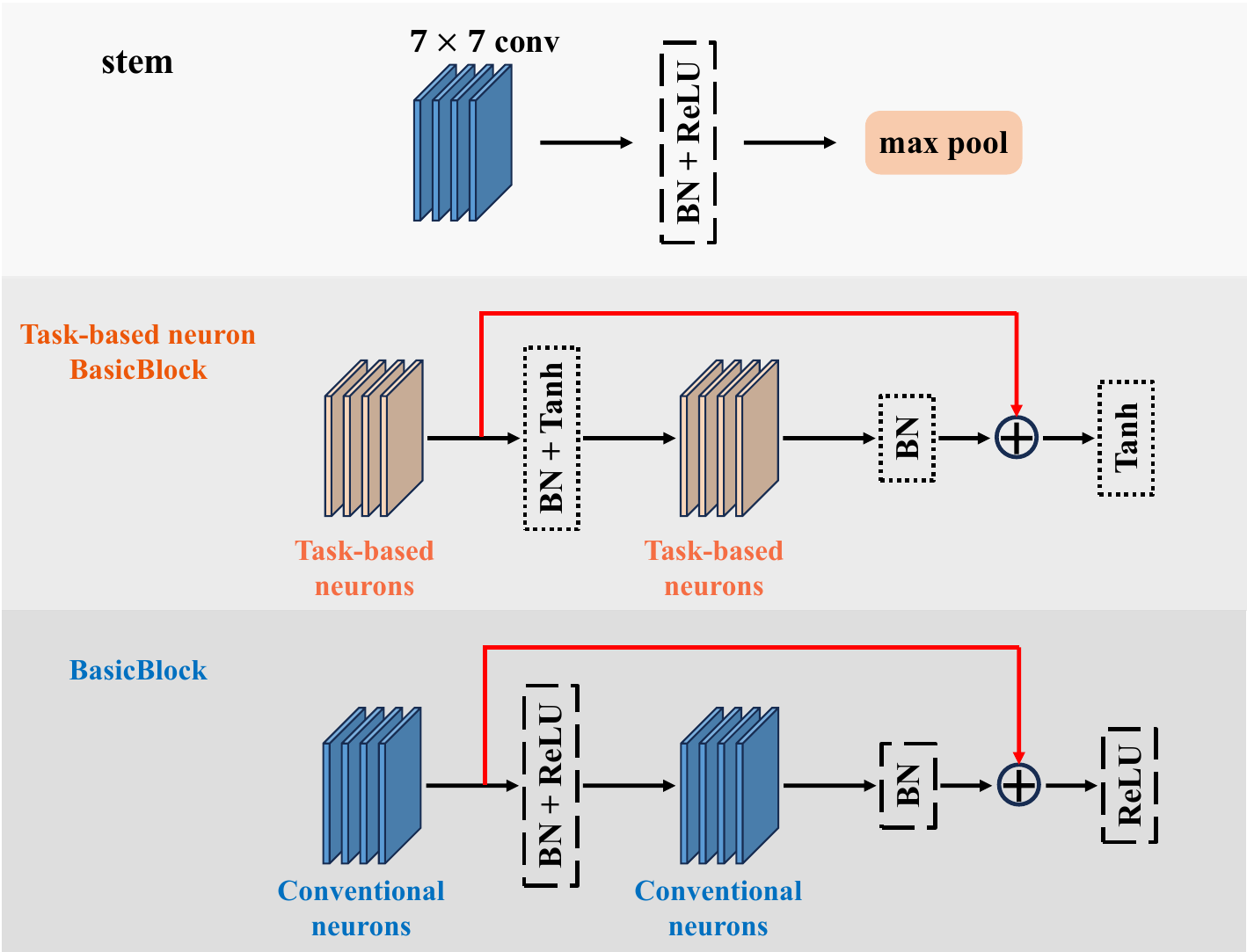}
            \label{fig:network_detail}
        }
    \end{adjustbox}
    \caption{Backbone networks structure and their components structure diagram.}
    \label{fig:backbone}
\end{figure*}

Please see Figure \ref{fig:backbone}. Each network begins with a stem consisting of a $7\times7$ convolution, batch normalization, ReLU activation, and a $3\times3$ max pooling layer with stride~2 (Figure \ref{fig:network_overview}). 
The backbone subsequently follows a staged residual design, where a \textit{stage} denotes a sequential group of residual blocks (Figure \ref{fig:network_detail}), typically sharing the same output dimensionality. 
% (Under this convention, Stage-1 refers to the first residual group immediately following the stem.) 
This is followed by three to four residual stages, a global average pooling layer, and a fully-connected output classifier. Conventional networks include an additional residual stage (Stage-4). We keep all training configurations identical.

We also test the performance of task-based neurons on the universal tasks. Here, the universal tasks mean the classification performance on the full-scale ImageNet. 

\begin{table*}[ht!]
	\centering
	\caption{VSR results under different class scales.}
	\label{tab:vsr_classes}
	\resizebox{0.85\textwidth}{!}{
	\begin{tabular}{ccc}
		\toprule
		Classes  & ResNet-18 & ViT-B-32 \\
		\midrule
		4    
		&  $\bm{0.00020}(\x \odot \x)^\top-\bm{0.028}\x^\top$         
		&    $\bm{-0.00024}(\x \odot^4 \x)^\top + \bm{0.0022}(\x \odot^3 \x)^\top+ \bm{0.0063}(\x \odot^2 \x)$ \\
        
		8          
		&     $\bm{0.001}(\x \odot^3 \x)^\top +\bm{0.0012}(\x \odot \x)^\top +\bm{0.00032}\x^\top$ 
		&  $\bm{0.00073}(\x \odot^3 \x)^\top +\bm{0.00073}(\x \odot \x)^\top +\bm{0.0016}\x^\top$  \\
		
		12         
		& $\bm{0.00053}(\x \odot \x)^\top -\bm{0.019}\x^\top+0.017$        
		&  $\bm{0.00013}(\x \odot^4 \x)^\top + \bm{0.00021}(\x \odot^3 \x)^\top - \bm{0.0049}(\x \odot^2 \x)^\top - \bm{0.030}\x^\top+0.0048$    \\
		
		16  		
        & $\bm{0.026}(\x \odot \x)^\top - 0.0026$ 
		&  $\bm{0.0022}(\x \odot \x)^\top -\bm{0.029}\x^\top-0.0014$    \\
		
		20         
		&   $\bm{-0.0048}(\x \odot \x)^\top + \bm{0.025}\x^\top-0.0076$        
		&  $\bm{ -0.0042}(\x \odot \x)^\top-\bm{0.026}\x^\top+0.024$         
        \\
		\bottomrule
	\end{tabular}
}
\end{table*}

Firstly, we apply the VSR on the embeddings of images from the ImageNet. Specifically, all images are first resized to $224 \times 224$ and normalized following standard ImageNet preprocessing. We then employ an autoencoder to learn compact representations of the images. The autoencoder is trained using a reconstruction objective with mean squared error loss, optimized by Adam with a learning rate of $10^{-3}$. The latent dimension is set to 16, and the model is trained with a mini-batch size of 128 for up to 50 epochs, with early stopping (patience = 10) applied to prevent overfitting. This dimensionality reduction step is essential to make the symbolic search tractable while preserving the most informative semantic structures of the data. After training, the encoder is used to extract low-dimensional embeddings for all images, resulting in a feature matrix $X \in \mathbb{R}^{N \times 16}$, where $N$ denotes the number of samples. The corresponding class labels are retained as targets. These embeddings serve as the input to VSR module. For VSR, we set population size=3,000, generations=30, mutation rate=0.6, crossover rate=0.3. VSR performs symbolic search directly on the embedding vectors. The entire pipeline is executed with fixed random seeds to ensure reproducibility.

Conducting this kind of regression over 20 times with different random seeds, we find that the obtained formulas are only two cases: linear and quadratic polynomials. The former corresponds to the linear neuron, while the latter corresponds to the quadratic neurons \cite{goyal2020improved,bu2021quadratic,xu2022quadralib,fan2018new}. Table~\ref{tab:vsr_classes} reports the VSR results obtained from two backbone models (ResNet18 and ViT-B-32) on subsets consisting of the lowest-accuracy classes. Specifically, for each model, we construct classification tasks using 4, 8, 12, 16, and 20 classes selected according to their classification difficulty, measured by the baseline model performance. Table~\ref{tab:vsr_classes} shows that VSR can learn high-order polynomials for a few classes.

We interpret this phenomenon as a degenerate behavior arising from a capacity mismatch between the function class and task complexity. Specifically, when modeling increasingly complex multi-class structures using a symbolic function defined on high-dimensional embedding vectors, the optimization tends to favor representations that minimize the global empirical risk rather than capturing fine-grained class-specific variations. As a result, VSR progressively focuses on the most common, low-frequency components of the data distribution, while higher-order terms are suppressed due to their limited contribution to overall error reduction. As we know, linear and quadratic polynomials are the most basic and of low frequency. According to the algebraic fundamental theorem \cite{fine1997fundamental}, with no introduction of complex numbers, any univariate polynomial can be decomposed into a combination of linear and quadratic terms.

Our earlier peer-reviewed work \cite{fan2025one} has conducted experiments on the ImageNet to show the effectiveness of quadratic neurons in the convolution and Transformer architecture.

%\section*{acknowledgement}

\renewcommand\refname{Reference}

\bibliographystyle{ieeetr}
\bibliography{reference.bib}

\ifCLASSOPTIONcaptionsoff
  \newpage
\fi